%% file: main.tex
\def\arxivversion{}

\documentclass[11pt]{article}

\ifdefined\arxivversion
  \usepackage[preprint]{acl}
\else
  \usepackage[review]{acl}
\fi

\usepackage{latexsym}
\usepackage{amsmath}
\usepackage{fontspec}
\newfontfamily\korean{NanumGothic.ttf}[Path=fonts/,BoldFont=NanumGothicBold.ttf]

\usepackage{microtype}

\usepackage{graphicx}
\usepackage{stfloats}
\usepackage{placeins}
\usepackage{capt-of}
\usepackage{tcolorbox}
\tcbuselibrary{breakable,skins}
\usetikzlibrary{arrows.meta}

\title{ReCoPII: A Relation-Grounded Framework for Building Contextual PII Masking Datasets}

\ifdefined\arxivversion
  \input{arxiv_authors}
\else
  \author{Anonymous ACL submission}
\fi

\begin{document}
\maketitle
\raggedbottom
\begin{abstract}
Training contextual masking systems for personally identifiable information (PII) requires examples distinguishing maskable from preservable mentions of one entity type, but privacy restrictions limit access to real data. Author-based augmentation (SPY) labels mentions unrelated to the author as non-PII, presuming a single data subject and overlooking others. We propose ReCoPII (Relation-grounded Contextual PII masking), a framework that builds such data without this premise: it plans ontology-constrained entity relations, writes insertion sentences with placeholders, reconstructs annotations with code, checks the result with a separate LLM, and derives MASK/KEEP labels from entity relations. On Korean documents from six domains, it yields ReCoPII-Ko, a corpus with relation, privacy, and MASK/KEEP annotations, which we validate. Across 300 pairs, four annotators agree with 88.42\% of its sampled KEEP labels, against 26.27\% for SPY-style augmentation of the same sources. Classifiers trained on it over-mask less than under that reference on a matched test. Contextual masking fine-tuned jointly with a PII detector removes most over-masking of detection alone within a fixed leakage budget; on human-verified entity records, the four detectors meeting it over-mask at most 5.04\% with at most 4.18\% leakage. The detector is comparable to XLM-R-large on KDPII with about a quarter of its parameters.
\end{abstract}

\section{Introduction}
\label{sec:introduction}

Privacy restrictions limit access to annotated personal data, motivating synthetic corpora for personally identifiable information (PII) detection \citep{savkin2025spy}. Contextual masking in Korean long documents is a problem of \emph{same type, different decisions}: mentions of one entity type need different MASK/KEEP labels, since a personal telephone number should be masked while an organization's service number can be preserved. Useful training data for this distinction require coherent documents whose text supports each masking label.

Existing resources address parts of this need. KDPII provides Korean PII span and type annotations \citep{fei2024kdpii}. CAPID and PII-Bench make contextual retention depend on relevance to a downstream question \citep{ponomarenko2026capid,shen2026piibench}. For documents without a query, SPY generates personal details about the text's author as PII and adds mentions unrelated to the author as non-PII \citep{savkin2025spy}. These approaches judge every mention against one reference point: the question in query-conditioned settings, and the author when no question is given, which is the query-free setting that we study.

That author-based criterion is insufficient when a document describes several people. In a contract, a counterparty's telephone number is unrelated to the author yet personal, and 703 of our 1,857 documents contain two or more maskable \texttt{NAME} entities. Its KEEP labels are therefore often not valid: in a human audit, annotators agree with only 26.27\% of sampled KEEP labels in SPY-style augmentation of our source documents (Table~\ref{tab:rq2a-human}), and classifiers trained on that augmentation in turn mask nearly half of the preservable mentions on a matched synthetic test (Section~\ref{sec:rq2b}). Detection by entity type alone masks nearly all human-verified preservable mentions (Section~\ref{sec:rq4}).

The author-based rule assigns privacy outcomes from a mention's association with one person. A labeling framework for documents with several people and organizations must instead establish whose information a mention conveys and in what role. We represent the domain, relation labels, and endpoint entity types as a \emph{domain-conditioned entity-relation configuration}; the document supplies the evidence for each edge-level privacy label, so the relation structure alone does not fix the privacy outcome.

We hypothesize that deriving MASK/KEEP labels from these configurations and the document's evidence yields valid contextual masking data without presuming a single data subject.

Contextual Integrity motivates this hypothesis by assessing information flows through the actors and context in which they occur, a view that Privacy Checklist operationalizes for privacy judgment \citep{li2025privacychecklist}. GoldCoin further shows that structured contextual features can guide the synthesis of privacy-law cases \citep{fan2024goldcoin}.

We propose ReCoPII (Relation-grounded Contextual PII masking), a framework for constructing contextual masking data from PII-annotated documents and a domain ontology (Figure~\ref{fig:construction-overview}). The technical challenge is to add preservable mentions while keeping their textual roles, values, and annotations consistent. We plan ontology-constrained entity relations and let an LLM write insertion sentences containing placeholders; code fills the values and reconstructs spans, relations, and edge-level privacy labels, and a separate LLM reviewer checks the completed text without seeing the planned labels. Equation~\ref{eq:entity-masking} aggregates the privacy labels of incident edges into each entity's MASK/KEEP decision. Applying it to Korean documents from six domains yields the ReCoPII-Ko corpus, on which we evaluate the framework.

We then check whether the resulting data are valid and usable. On validity, four annotators agree with 88.42\% of the sampled KEEP labels of ReCoPII-Ko, more than three times the rate under the author-based criterion on the same source documents. On use, the same classifiers over-mask less when trained on ReCoPII-Ko than under that SPY-style reference (Section~\ref{sec:rq2b}). A complete system can also be trained on the data: under a fixed validation leakage budget, contextual masking fine-tuned jointly with a detector removes most of the over-masking of detection alone (Table~\ref{tab:rq3-main}), and for the four detectors that meet the budget it over-masks at most 5.04\% of human-verified preservable mentions with at most 4.18\% leakage. The detector is comparable to XLM-R-large on KDPII with about a quarter of its parameters.

Our contributions are threefold:
\begin{itemize}
    \item \textbf{Method:} ReCoPII, a relation-grounded framework for building contextual masking datasets that plans ontology-constrained augmentation, reconstructs annotations with code, checks their textual support, and derives MASK/KEEP labels from relation graphs.
    \item \textbf{Resource:} ReCoPII-Ko, a Korean long-document corpus from six domains with maskable and preservable mentions within the same entity types, annotated with relations and edge-level privacy labels.
    \item \textbf{Evaluation:} Validation of these data: label and document validity in human and LLM audits, and classifiers and an integrated masking system trained on them, scored on generated labels and on human-verified entity records.
\end{itemize}

\section{Related Work}
\label{sec:related-work}

Rules, gazetteers, and encoder-based sequence labelers detect PII spans and entity types \citep{savkin2025spy}; our candidate detector follows this line. Query-conditioned approaches decide retention relative to a downstream question: CAPID fine-tunes a local small language model to detect and type PII spans and estimate their relevance to the question \citep{ponomarenko2026capid}, and PII-Bench tests whether a system masks query-unrelated PII \citep{shen2026piibench}. SPY builds query-free supervision around the text's author, and its authors note that personal information about individuals closely related to the author is not covered \citep{savkin2025spy}. SPIA identifies the same single-subject assumption in anonymization evaluation and measures protection per individual \citep{oh2026spia}. We address it in data construction for contextual masking: ReCoPII keeps the query-free setting but replaces the single-author criterion with the domain-conditioned entity-relation configuration, which can represent several data subjects and organizations in one document and follows the contextual view of privacy cited in Section~\ref{sec:introduction}.

\section{Method}
\label{sec:method}

\begin{figure*}[t]
\centering
\includegraphics[width=0.86\textwidth,keepaspectratio]{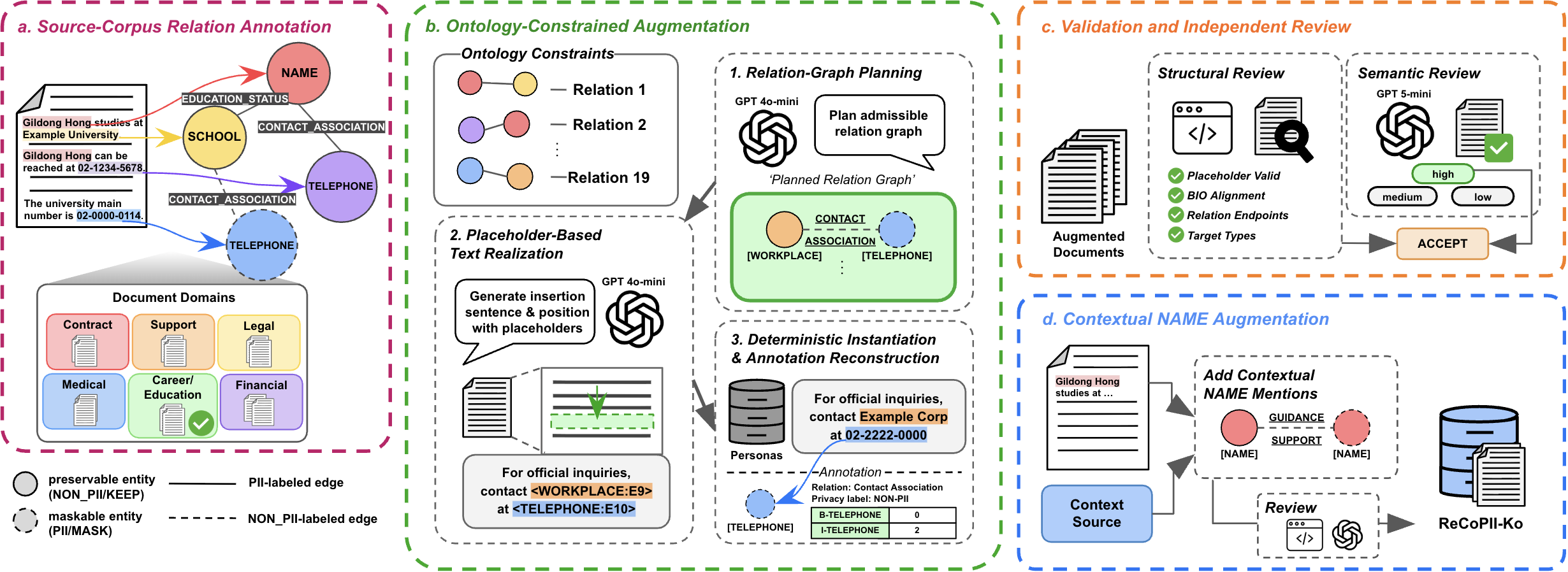}
\caption{Overview of relation-grounded data construction. (a) Source documents receive entity-relation annotations. (b) Ontology-constrained relation-graph planning guides placeholder-based text realization, followed by deterministic instantiation and annotation reconstruction. (c) Structural validation and independent semantic review determine acceptance. (d) Contextual \texttt{NAME} augmentation adds preservable name mentions and supporting relations, followed by review.}
\label{fig:construction-overview}
\end{figure*}

\subsection{Task Definition}

Given a Korean long document $D$ with domain $d$, our task is to identify PII-type mentions and determine whether each entity should be masked or preserved from its role in the document. A preservable mention is a PII-type mention that does not require masking in the given context, not a value that is inherently non-PII. A domain-conditioned entity-relation configuration specifies the relation structure; the document supplies the evidence for its privacy label.

Figure~\ref{fig:method-context-contrast} illustrates this distinction in a synthetic financial document. Both examples link the same company to a bank account through \texttt{FINANCIAL\_ASSET\_ASSOCIATION}, but an employee's payroll account is masked, whereas the company's settlement account is preserved. Because the endpoint types and relation label match while the roles differ, we separate PII candidate detection, which locates mentions and their types, from the contextual masking decision, which uses the document context and domain.

\input{figures/method/context_contrast}

A mention $m=(s,b,e)$ identifies one occurrence in sentence $s$ at character span $[b,e)$. An entity $v_i$ groups its mentions $\mathcal{M}_i$ and has type $\tau(v_i)$. We represent document relations as a graph $G=(V,E)$, where $V$ contains entities and each directed edge $a=(v_i,r,v_j)\in E$ has relation label $r$ and a separate privacy label $y(a)\in\{\texttt{PII},\texttt{NON\_PII}\}$, assessed in the context of $D$ and $d$ rather than fixed per triplet. An edge label does not directly determine both endpoints' masking labels: an official contact edge can be \texttt{NON\_PII}, while its organization endpoint remains maskable through another relation to a person.

We therefore aggregate all edges incident to $v_i$, irrespective of direction, into a set $I_i$. Let $n_i=|I_i|$ and $p_i=|\{a\in I_i:y(a)=\texttt{PII}\}|$. The final entity masking decision is
\begin{equation}
\mu(v_i)=\left\{\begin{array}{ll}
0, & n_i>0\ \land\ p_i=0,\\
1, & \mathrm{otherwise}.
\end{array}\right.
\label{eq:entity-masking}
\end{equation}
Here, $1$ denotes a maskable entity (\texttt{PII}/\texttt{MASK}) and $0$ a preservable entity (\texttt{NON\_PII}/\texttt{KEEP}). Preservation thus requires relation evidence and no incident \texttt{PII} edge; an entity with no relations defaults to masking.

\subsection{Controlled Data Construction}

Figure~\ref{fig:construction-overview} summarizes the construction. We start from Basic, a source corpus of 2,355 Korean long documents across six domains and 18 entity types with sentence-level PII spans and character-level masking BIO tags (Figure~\ref{fig:source-corpus-construction} in Appendix~\ref{sec:synthetic-data-statistics}). We organize its mentions into document-level entity registries and annotate relations and edge-level privacy labels in context, using domain-specific relation schemas defined with GPT-5.6-Sol; Equation~\ref{eq:entity-masking} then gives each entity's masking label. To obtain maskable and preservable outcomes within one entity type, we augment these documents with preservable entities and supporting relations, whose values come from the persona pool that also supplied the source values (Appendix~\ref{app:construction-instantiation}).

\paragraph{Ontology-constrained planning.}
\label{sec:graph-planning}
An LLM that writes text and labels together can invent relations that its text does not support, so we fix the relation structure before any wording is chosen. GPT-4o-mini receives the document, its graph, and the domain's relation meanings and proposes organization-managed entities with provisional \texttt{NON\_PII} edges. Let $\Omega_d$ be the set of endpoint-type and relation combinations that the domain ontology permits and the persona pool can instantiate in organizational contexts. Every planned edge $(v_i,r,v_j)$ must satisfy
\begin{equation}
(\tau(v_i),r,\tau(v_j))\in\Omega_d.
\label{eq:ontology-constraint}
\end{equation}
Code rejects plans with incompatible endpoint types, unconnected new entities, or unfillable values and requests regeneration (Appendices~\ref{app:construction-planning} and~\ref{app:prompt-planner}).

\paragraph{Placeholder realization and deterministic annotation.}
\label{sec:text-realization}
\label{sec:instantiation}
An LLM that also writes values and annotations can produce incompatible values, such as an organization with a department it lacks, or spans that do not match the text. GPT-4o-mini therefore writes only insertion sentences expressing each planned role, with placeholders such as \texttt{<TELEPHONE:E10>} for planned entities (Appendix~\ref{app:prompt-realizer}). Code checks the placeholders against the plan and rejects sentences attributing a planned target to a competing source of the same type (Appendix~\ref{app:construction-realization}). It then fills values through label-independent queries to the persona pool, where related entities needing consistent attributes share a persona row, and reconstructs spans, relations, and edge-level privacy labels from the replacement trace. Only mentions of maskable entities enter \texttt{PII\_set} and masking BIO tags; candidate detection targets all registry mentions (Appendix~\ref{app:construction-annotation}).

\paragraph{Independent review and acceptance.}
\label{sec:independent-review}
Structural validity does not establish the privacy label, so a separate GPT-5-mini reviewer judges each relation from the completed document without the intended labels or the construction plan (Appendix~\ref{app:prompt-reviewer}). The reviewer labels a relation \texttt{PII} when it attributes information to a specific individual or helps identify that individual, \texttt{NON\_PII} when a supported relation does not, and \texttt{AMBIGUOUS} when the document gives no evidence. A new edge judged \texttt{PII} with a verified supporting quote, or a reported missing \texttt{PII} relation involving a new entity, rejects the document, since retaining it could leave a maskable mention unmasked; Appendix~\ref{app:construction-review} gives the remaining acceptance conditions and the review's partial blinding. A final filter removes accepted documents that contain a new entity without an incident relation, which cannot be preserved under Equation~\ref{eq:entity-masking} (Table~\ref{tab:generation-attrition}).

\paragraph{NAME augmentation.}
\label{sec:name-augmentation}
Organization-managed entities provide a limited range of preservable roles for names. We therefore add \texttt{NAME} mentions used as public references supported by source facts or as fictional participants in hypothetical examples, each linked by one text-supported \texttt{NON\_PII} relation to an existing \texttt{NAME} entity and checked by a dedicated review (Appendix~\ref{app:construction-name}).

\subsection{PII Candidate Detection and Contextual Masking}
\label{sec:two-stage}

Candidate detection locates PII-type mentions; contextual masking decides which of those mentions require protection. We use both to test whether the data can train a complete masking system, and they share one encoder (Appendix~\ref{app:detector-training} gives the architecture and its equations).

\paragraph{PII candidate detection.}
Our detector, A.X+\allowbreak{}Gazetteer+\allowbreak{}CRF+\allowbreak{}MLP, combines A.X, a training-only Gazetteer, a linear-chain CRF that decodes candidate types and token spans, and a character boundary MLP that refines their offsets, because a Korean entity boundary may fall inside a token. We train it on KDPII; used alone, it masks every predicted candidate.

\paragraph{Contextual masking.}
Contextual masking groups predicted mentions into entities by type and normalized surface form, a heuristic rather than coreference resolution, and learns from the labels of Equation~\ref{eq:entity-masking} in one of two variants. The \emph{direct head} predicts each entity's masking label $\mu(v_i)$ from the mean of its mention representations, without relations. \emph{Relation-conditioned heads} instead predict, for each unordered entity pair, its relation labels and the probability that each relation has a PII label, and aggregate the thresholded edges with Equation~\ref{eq:entity-masking}, so that every decision can be traced to the edges that caused it. \emph{Joint Fine-tuning} updates the shared encoder and detector together with the heads, adding the detection objective to the contextual loss, and is run with both variants; \emph{Frozen Detector} trains only relation-conditioned heads. Both train the heads from gold mentions, including KEEP mentions, and use predicted candidates at inference; thresholds and checkpoints are selected on validation data (Section~\ref{sec:rq4}).

\section{Dataset Statistics}
\label{sec:dataset}

ReCoPII-Ko contains 1,857 documents derived from a cleaned Basic corpus of 2,355 documents, with 18 entity types and 25,656 annotated edges across 19 relation labels (Table~\ref{tab:dataset-overview}). Of the 2,355 source documents, 1,996 pass augmentation acceptance and 1,857 remain after the isolated-entity filter (Appendix Table~\ref{tab:generation-attrition}). Appendix~\ref{sec:synthetic-data-statistics} defines the entity types and reports domain, length, type, and relation composition (Tables~\ref{tab:pii-type-definitions}, \ref{tab:appendix-domains}, \ref{tab:appendix-corpus}, \ref{tab:appendix-type-mentions}, and~\ref{tab:appendix-relations}).

\paragraph{Official split.}
We designate a source-group split of the current archive as the official partition, with 1,245 training, 208 validation, and 404 test documents; the ReCoPII-Ko and SPY versions of a document share its assignment. Documents that share a Basic source identifier or identical text form one source group, and each group stays in one partition, so no source document contributes to both training and evaluation.

\begin{table}[t]
\centering
\small
\setlength{\tabcolsep}{3pt}
\renewcommand{\arraystretch}{1.05}
\begin{tabular}{@{}lcc@{}}
\hline
Measure & ReCoPII-Ko & SPY \\
\hline
Mean chars / doc & 2,157.0 & 2,063.5 \\
PII mentions & 50,620 & 49,307 \\
MASK entities & 17,104 & 16,895 \\
KEEP entities & 6,780 & 11,603 \\
KEEP share (\%) & 28.4 & 40.7 \\
Relation edges & 25,656 & --- \\
\hline
\end{tabular}
\caption{Corpus-level statistics of ReCoPII-Ko and the SPY-style version of the same 1,857 source documents. PII mentions count sentence-level occurrences including repeats, whereas MASK/KEEP count document-level entity records; SPY has no relation annotations.}
\label{tab:dataset-overview}
\end{table}

\paragraph{Outcomes within entity types.}
MASK/KEEP composition is not fixed by entity type: \texttt{EMAIL} contains 1,233 MASK and 1,653 KEEP entities, \texttt{TELEPHONE} 1,035 and 1,851, and \texttt{NAME} 2,667 and 58 (all 18 types in Appendix Table~\ref{tab:dataset-type-decisions}; examples in Appendix Figure~\ref{fig:pii-mask-keep-example}).

\section{Experiments}
\label{sec:experiments}

\subsection{Overall Setup}

The experiments check whether the framework's data are valid and usable. RQ1 checks the detector used by the masking system: how accurately KDPII-trained detectors identify PII and transfer to synthetic documents. RQ2 asks whether the data are (A) valid in document quality and labels and (B) usable as supervision for contextual MASK/KEEP decisions; SPY-style augmentation of the same source documents \citep{savkin2025spy} is the reference point. RQ3 asks whether a complete masking system can be trained on the data, relative to masking every detected candidate and to LLMs given the same candidates, and compares its relation-conditioned and direct variants as a sub-question.

RQ1 and RQ3 use the test partition of the official split (Section~\ref{sec:dataset}). RQ2-B predates that split and uses an earlier snapshot of ReCoPII-Ko with a split by document ID, in which no document belongs to two partitions and no test document repeats the text of a training or validation document (Appendix Table~\ref{tab:appendix-data-versions}); we therefore do not compare scores across RQ1/RQ3 and RQ2-B. All model evaluations score labels generated by the construction pipeline, except the human-verified test of RQ3.

\subsection{RQ1: Detection and Cross-Dataset Transfer}
\label{sec:rq1}

We train each detector on KDPII \citep{fei2024kdpii}, using 19 of its 33 PII labels and a leakage-aware 3,586/\allowbreak{}399/\allowbreak{}996-document split, and evaluate its validation-selected checkpoints from seeds 42--44 on the KDPII test set and, without synthetic training or threshold tuning, on the 404-document official test, using exact character-span-and-type micro Detection F1 (Appendix~\ref{sec:rq1-complete}).

On KDPII, A.X+\allowbreak{}Gazetteer+\allowbreak{}CRF+\allowbreak{}MLP is comparable to XLM-R-large, the strongest language-model baseline of the original KDPII study \citep{fei2024kdpii}, with about a quarter of its total parameters: a pre-specified five-seed test gives $86.71\pm1.47$ versus $85.32\pm1.24$, and the confidence interval of the 1.40-point difference, $[-0.37, 3.04]$, includes zero. Among nine KDPII-trained encoders it also has the highest three-seed mean on both test sets, on the synthetic test by a negligible margin (Appendix Table~\ref{tab:rq1-transfer-all}). The boundary MLP was informed by analysis of earlier test errors, and both seed extensions were decided after the three-seed results.

\subsection{RQ2: Data Validity and Usability}

\subsubsection{RQ2-A: Document and Label Validity}
\label{sec:rq2a}

\paragraph{Setup.}
We randomly sampled 50 source documents from each of the six domains and paired the ReCoPII-Ko and SPY versions of each, giving 300 pairs that both LLM judges and human annotators evaluate. Three LLM judges (Claude Haiku 4.5, GPT-5.6 Luna, and Solar Pro 4) rate sentence naturalness, coherence with the source, and whether the inserted preservable mentions are contextually valid, and select the better document, producing 900 valid judgments (Appendix~\ref{app:prompt-judge}); paired sign-flip tests use Holm correction across the four measures. Four human annotators each assess all 300 pairs without seeing the Basic source or generated privacy labels (Appendix~\ref{sec:additional-evaluation}).

\begin{table}[t]
\centering
\small
\setlength{\tabcolsep}{5pt}
\begin{tabular}{@{}lrrrr@{}}
\hline
 & \multicolumn{2}{c}{LLM judges} & \multicolumn{2}{c}{Human} \\
Measure & R-Ko & SPY & R-Ko & SPY \\
\hline
Naturalness (1--5) & $\mathbf{2.77}$ & 2.37 & 3.24 & 2.65 \\
Coherence (1--5) & $\mathbf{2.14}$ & 1.85 & 3.15 & 2.24 \\
Preference (\%) & $\mathbf{60.89}$ & 30.33 & 76.92 & 7.92 \\
KEEP validity (\%) & 35.06 & 29.19 & 88.42 & 26.27 \\
MASK agreement (\%) & --- & --- & 96.29 & 95.75 \\
\hline
\end{tabular}
\caption{RQ2-A results on 300 matched document pairs of ReCoPII-Ko (R-Ko) and SPY (900 LLM judgments, 1,200 human reviews); bold marks Holm-corrected $p<0.001$. LLM judges rate coherence with the source and whether each item presented as NON\_PII is valid in context. Humans rate document coherence and, without seeing labels, judge up to two stored KEEP and two stored MASK entities per document as MASK, KEEP, or UNCERTAIN; agreement counts UNCERTAIN as disagreement with the generated label and is not adjudicated gold. Preference ties: 8.78\% (LLM) and 15.17\% (human). Appendix~\ref{sec:additional-evaluation} gives 95\% confidence intervals.}
\label{tab:rq2a-quality}
\label{tab:rq2a-human}
\end{table}

\paragraph{Results.}
Annotators accept most labels of ReCoPII-Ko: they agree with 88.42\% of its sampled KEEP labels, whereas under the SPY reference KEEP agreement is 26.27\%, with about two fifths of KEEP judgments uncertain; MASK agreement is similarly high for both (Table~\ref{tab:rq2a-quality}). On document quality, annotators prefer the ReCoPII-Ko document in about three quarters of reviews, and the LLM judges rate it higher in naturalness and source coherence and prefer it about twice as often as the SPY document; these differences remain significant after Holm correction, whereas the NON\_PII-validity difference is not ($p=0.236$). On whether preserved items are valid, the LLM judgments thus do not separate the two data sources, whereas the human judgments differ widely.

\subsubsection{RQ2-B: Contextual Masking Supervision}
\label{sec:rq2b}

\paragraph{Setup.}
We train six classifier configurations on ReCoPII-Ko documents or, as the reference, on matched SPY-style documents: four A.X configurations read the full document, and KLUE-R-base and XLM-R-large read 512-token windows. Each input contains a supplied mention span and its entity type, and the target is MASK or KEEP without relation labels. The training sets share 1,392 source documents with 4,140 KEEP mentions each. Every model is scored on both versions of the 279 test documents, with 15,604 MASK mentions (7,891 ReCoPII-Ko and 7,713 SPY) and 2,422 KEEP mentions (1,211 per version), over seeds 42--44 with document-paired bootstrap intervals. A preregistered validity check requires PII leakage below 5\% on both original and inserted sentences of each model's own generation method.

\begin{table}[t]
\centering
\small
\setlength{\tabcolsep}{2.5pt}
\begin{tabular}{@{}lcccccc@{}}
\hline
 & \multicolumn{2}{c}{Macro-F1 $\uparrow$} & \multicolumn{2}{c}{Over-mask $\downarrow$} & \multicolumn{2}{c}{Leakage $\downarrow$} \\
Model & SPY & R-Ko & SPY & R-Ko & SPY & R-Ko \\
\hline
A.X & 82.68 & \textbf{86.50} & 47.01 & \textbf{38.18} & 0.12 & 0.14 \\
A.X+Gaz & 82.73 & \textbf{85.26} & 46.86 & \textbf{41.18} & 0.13 & 0.13 \\
A.X+CRF & 82.23 & \textbf{85.13} & 47.91 & \textbf{41.45} & 0.13 & 0.13 \\
A.X+Gaz+CRF & 82.35 & \textbf{85.69} & 47.66 & \textbf{39.94} & 0.13 & 0.19 \\
KLUE-R-base$^\dagger$ & 81.17 & \textbf{82.97} & 49.75 & \textbf{45.79} & 0.24 & 0.27 \\
XLM-R-large$^\dagger$ & 80.26 & \textbf{83.54} & 52.16 & \textbf{44.49} & 0.09 & 0.28 \\
\hline
\end{tabular}
\caption{RQ2-B mention-level results (\%) on the matched common test: means over three seeds for classifiers trained on SPY-style documents (SPY) or ReCoPII-Ko (R-Ko). Gaz abbreviates Gazetteer, so the A.X rows are the encoder alone, A.X+Gazetteer, A.X+CRF, and A.X+Gazetteer+CRF. Bold marks the better value within each model. $^\dagger$ marks configurations that failed the preregistered validity check and are descriptive only. Appendix Table~\ref{tab:rq2b-masking-full} adds standard deviations.}
\label{tab:rq2b-masking}
\end{table}

\paragraph{Results.}
Among the four valid A.X configurations in Table~\ref{tab:rq2b-masking}, classifiers trained on ReCoPII-Ko reach 2.54--3.83 points higher Macro-F1 and 5.68--8.84 points less over-masking than under the SPY reference; every document-paired confidence interval for these differences excludes zero, whereas all four leakage intervals include zero. KLUE-R-base and XLM-R-large show the same Macro-F1 direction but fail the validity check, with 8.83\% and 9.78\% maximum leakage on inserted sentences of their own training method, and training XLM-R-large on ReCoPII-Ko raises its common-test leakage. RQ2-B therefore indicates that the data are usable as contextual MASK/KEEP supervision for the valid A.X family; the two windowed encoders remain descriptive evidence. A context analysis shows what these decisions use: when a classifier reads the full document instead of only the entity's value and type, entity-level Macro-F1 rises far more for ReCoPII-Ko than for SPY (Appendix Table~\ref{tab:rq3-context}). Value and type thus determine most SPY decisions, whereas ReCoPII-Ko decisions depend on textual context (Appendix~\ref{sec:rq2b-appendix}).

\input{analysis/human_gold_test/combined_main_table}

\subsection{RQ3: Integrated Masking System}
\label{sec:rq4}

\paragraph{Setup.}
RQ3 trains the system of Section~\ref{sec:two-stage} on the data and evaluates it end to end on the official 404-document test, where each condition masks the candidates of its own detector. \emph{Detection only} masks every candidate of a KDPII-trained detector. Frozen Detector and Joint Fine-tuning train on the 1,245 training documents from one KDPII detector checkpoint (seed 42), with seeds 42--44 varying the contextual heads and joint training; Joint Fine-tuning is run with both variants for five detectors, the three in Table~\ref{tab:rq3-main}, A.X+Span, and KLUE-R-base. The candidate-assisted LLM baseline gives GPT-4o-mini, K-EXAONE, or Solar Pro 3 the document and the A.X+Gazetteer+CRF detection-only candidates; the LLM predicts relations and edge-level privacy labels, which Equation~\ref{eq:entity-masking} turns into MASK/KEEP decisions (Appendix~\ref{sec:rq4-appendix}).

\paragraph{Metrics and threshold selection.}
Over-masking is the share of the 1,750 KEEP mentions with any masked character, and leakage is the share of the 11,045 MASK mentions with any unmasked character. Joint Fine-tuning selects thresholds and checkpoints by the lowest validation over-masking among settings whose total validation leakage is at most 1.5\%. We fixed this rule after the registered runs, from which Detection only, Frozen Detector, and the LLM baseline come, but before observing any of its results, so the variant comparison is post hoc (Appendix~\ref{sec:rq4-appendix}).

\paragraph{Results.}
A complete masking system can be trained on ReCoPII-Ko: joint fine-tuning removes almost all over-masking, and leakage falls as well (Table~\ref{tab:rq3-main}). Masking every candidate (Detection only) or training only the heads (Frozen Detector) leaves most preservable mentions masked, whereas Joint Fine-tuning over-masks at most 2.80\% at no more than 1.44\% leakage across the four detectors that meet the budget (Appendix Table~\ref{tab:rq4-all}); the fine-tuned detectors alone leak 0.62--0.90\%, so the contextual heads add 0.19--0.68 points. XLM-R-large does not meet the budget in five of its six runs, in which masking every candidate of the fine-tuned detector already exceeds it on validation. The candidate-assisted LLMs release preservable mentions, but all six runs add leakage beyond that of their candidates.

\paragraph{Human-verified test.}
The reduction also appears when annotators verify the labels. Because the scores above use generated labels, we also score the same systems on a human-verified test: 100 ReCoPII-Ko/SPY document pairs from the official test partition, in which four annotators judged sampled entity records without seeing the stored labels, and a record's reference label is the MASK or KEEP decision of at least three of them (Appendix~\ref{sec:human-gold-test}). On the 238 reference-KEEP and 670 reference-MASK mentions of the ReCoPII-Ko documents (192 and 199 entities), Joint Fine-tuning over-masks 2.52--5.04\% at 1.14--4.18\% leakage for the four detectors that meet the budget, whereas Detection only masks nearly all preservable mentions for each of nine KDPII-trained detectors (three-seed means, Appendix Table~\ref{tab:human-gold-all}). Against the generated labels of the same records, Joint Fine-tuning over-masks less (Appendix Table~\ref{tab:human-gold-main}); a decomposition that was not pre-specified traces the difference to two entities that the data labels MASK and the annotators judge preservable. With 238 preservable mentions the intervals are wide, and because entity records were sampled per document, leakage here is not comparable with leakage on the full test.

\paragraph{Insertion-free probe.}
Almost all preservable mentions sit in inserted sentences (1,734 of the test's 1,750), so these scores could partly reflect recognition of inserted sentences; a pre-specified test finds that sentence origin affects the release of MASK mentions (Appendix~\ref{sec:sentence-origin}). We therefore apply the same checkpoints and thresholds to 402 documents a separate pipeline writes whole from a planned relation graph (generated labels, not human-verified). For the four detectors within budget, Joint Fine-tuning still leaves most preservable mentions unmasked, over-masking 11.33--34.21\% against at least 88.59\% for Detection only and leaking at most 4.95\%, both above the official test's 2.80\% and 1.44\%. The decision does not rest on inserted sentences alone, but part of the ReCoPII-Ko accuracy does not carry over (Appendix~\ref{sec:probe}).

\paragraph{Relation-conditioned heads versus direct head.}
On the full test, the direct head over-masks about one to one and a half points less than relation-conditioned heads, and the paired interval excludes zero for all four detectors; leakage differs only for A.X+Gazetteer+CRF, where relation-conditioned heads leak more (Appendix Table~\ref{tab:rq4-all}). On the human-verified records the direction is the same, but the paired interval excludes zero only for KLUE-R-base, so that test does not establish the difference on its own (Appendix Table~\ref{tab:human-gold-sensitivity}). This comparison shows only that a direct head suffices at inference; separately, the supervision itself is built from relations, since Equation~\ref{eq:entity-masking} derives each entity's label from its annotated edges. Because every relation-conditioned decision follows from retained edges, the gap can be traced: in the 12 relation-conditioned runs of the four detectors, 73.2--98.2\% of over-masked KEEP mentions are masked through an edge predicted as PII, and 84.8--95.7\% of those edges have an endpoint pair and relation label absent from the annotated graph. Their over-masking therefore comes mainly from predicted relations that the annotation does not contain.

\section{Conclusion}
\label{sec:conclusion}

We presented a relation-grounded framework that builds contextual masking data without presuming a single data subject, deriving MASK/KEEP labels for PII-type mentions from planned and verified entity relations. On Korean documents from six domains, annotators accept its labels: they agree with 88.42\% of its sampled KEEP labels, where author-based labels on the same source documents reach 26.27\%. The data are also usable: classifiers trained on them over-mask less on a matched test, and contextual masking fine-tuned jointly with a compact detector removes most over-masking within a fixed leakage budget; the reduction holds on human-verified entity records, where masking every candidate masks nearly all preservable mentions, and, to a lesser degree, on insertion-free documents. A direct head suffices at inference, while relation-conditioned heads keep each decision traceable to edges but over-mask more. Verifying every mention with independent adjudication is the next step. We will release the dataset and code.

\section*{Limitations}

Contextual MASK/KEEP decisions are evaluated only on synthetic Korean documents; performance on naturally occurring documents and other languages is untested. Except for the human-verified test, model evaluations score generated labels, so they measure agreement with the construction pipeline rather than the correctness of each privacy decision. The human-verified test covers at most four sampled entity records per document, and its 238 preservable mentions give wide intervals. Preservable mentions sit almost only in inserted sentences, so scores on the official test may partly reflect sentence origin.

\bibliography{custom}

\appendix
\makeatletter
\setlength{\@fptop}{0pt}
\setlength{\@fpsep}{20pt}
\setlength{\@dblfptop}{0pt}
\setlength{\@dblfpsep}{20pt}
\makeatother

\clearpage
\input{appendices/synthetic_data_statistics}

\input{appendices/synthetic_data_evaluation_templates}
\input{appendices/synthetic_data_analysis}

\input{appendices/augmentation_prompts}

\input{appendices/construction_details}

\end{document}

%% file: arxiv_authors.tex
\author{Taehwan Lee \\
  Department of Data Science, Hanyang University \\
  Seoul, Republic of Korea \\
  \texttt{peterson0620@hanyang.ac.kr}}

%% file: figures/method/context_contrast.tex
\begin{figure}[t]
\centering
\begingroup
\definecolor{ctxOrg}{HTML}{F7E4B5}
\definecolor{ctxAccount}{HTML}{D9EAF7}
\definecolor{ctxCue}{HTML}{FFF0AC}
\definecolor{ctxMask}{HTML}{A8333A}
\definecolor{ctxKeep}{HTML}{176451}
\newcommand{\ctxspan}[2]{\begingroup\setlength{\fboxsep}{1pt}\colorbox{#1}{\color{black}#2}\endgroup}
\newcommand{\ctxgraph}[5]{%
\begin{tikzpicture}[scale=0.9,transform shape,>=Stealth,every node/.style={font=\footnotesize},ctxentity/.style={circle,draw=black!45,minimum size=2cm,inner sep=0pt}]
\path[use as bounding box] (0,-1.15) rectangle (7.5,1.15);
\node[ctxentity,fill=ctxOrg] (person) at (1.05,0) {};
\node[ctxentity,fill=ctxAccount] (phone) at (6.45,0) {};
\node[inner sep=0pt,align=center] at (person.center) {{\korean #1}\\[-1pt]{\scriptsize WORKPLACE}};
\node[inner sep=0pt,align=center] at (phone.center) {{\scriptsize #2}\\[-1pt]{\fontsize{6}{7}\selectfont BANK\_ACCOUNT\_}\\[-2pt]{\fontsize{6}{7}\selectfont NUMBER}};
\draw[<->,semithick,#5] (person) -- node[above,font=\scriptsize,align=center]{FINANCIAL\_ASSET\_\\ASSOCIATION} node[below,font=\scriptsize\bfseries,text=#4]{#3} (phone);
\end{tikzpicture}\par}
\ctxgraph{한국철강}{7934-232-707420}{PII}{ctxMask}{solid}
\begin{tcolorbox}[colback=white,colframe=black!25,boxrule=.4pt,arc=1mm,left=1.5mm,right=1.5mm,top=1mm,bottom=1mm,before skip=3pt,after skip=3pt]
\footnotesize\raggedright
\textbf{(a) Personal payroll account}\hfill\textcolor{ctxMask}{\textbf{PII $\rightarrow$ MASK}}\par\smallskip
{\korean\fontsize{8.5}{11}\selectfont\color{black!48}
본 계약은 \ctxspan{ctxOrg}{한국철강}의 제관부 부서에서 근무 중인 \ctxspan{ctxCue}{최경환의 급여 자동이체 계좌}(\ctxspan{ctxAccount}{7934-232-707420})를 통해 매월 자동이체 방식으로 보험료를 납입합니다.
\ [\ldots]\par}
\smallskip
\textit{Insurance premiums are paid monthly from the payroll account (7934-232-707420) of Choe Gyeong-hwan, who works in the can-making department of Hanguk Cheolgang. [\ldots]}
\end{tcolorbox}
\ctxgraph{한국철강}{055-907337-96139}{NON\_PII}{ctxKeep}{dashed}
\begin{tcolorbox}[colback=white,colframe=black!25,boxrule=.4pt,arc=1mm,left=1.5mm,right=1.5mm,top=1mm,bottom=1mm,before skip=3pt,after skip=3pt]
\footnotesize\raggedright
\textbf{(b) Company settlement account}\hfill\textcolor{ctxKeep}{\textbf{NON\_PII $\rightarrow$ KEEP}}\par\smallskip
{\korean\fontsize{8.5}{11}\selectfont\color{black!48}
계약자는 모든 보험료를 \ctxspan{ctxOrg}{한국철강}의 \ctxspan{ctxCue}{정산 계좌}인 \ctxspan{ctxAccount}{055-907337-96139}를 통해 납부해야 합니다.
\par}
\smallskip
\textit{The policyholder must pay all premiums through Hanguk Cheolgang\textquotesingle s settlement account, 055-907337-96139.}
\end{tcolorbox}
\endgroup
\caption{Same company, endpoint types, and relation; different account roles in one synthetic document. The employee's payroll account is MASK, while the company's settlement account is KEEP. Account values differ. The company remains MASK through other PII edges (not shown). Graphs use stored generated labels, not human gold. Double-headed arrows indicate that masking aggregates the relation at both endpoints. Yellow highlights contextual evidence; English translations follow. [\ldots] marks omitted text.}
\label{fig:method-context-contrast}
\end{figure}
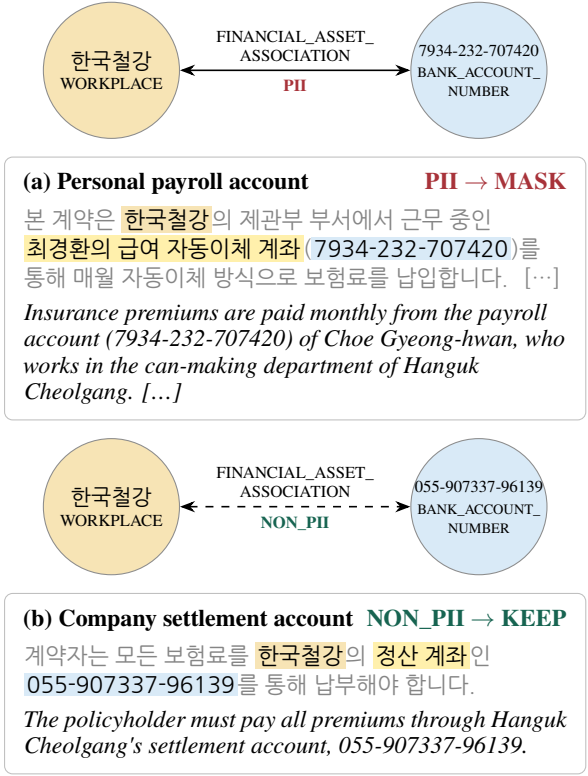

%% file: analysis/human_gold_test/combined_main_table.tex
\begin{table*}[t]
\centering
\small
\setlength{\tabcolsep}{4pt}
\begin{tabular}{llcccc}
\hline
 & & \multicolumn{2}{c}{Full test, generated labels} & \multicolumn{2}{c}{Human-verified records} \\
Detector & Condition & Over-mask $\downarrow$ & Leakage $\downarrow$ & Over-mask $\downarrow$ & Leakage $\downarrow$ \\
\hline
A.X+Gazetteer+CRF+MLP & Detection only & 95.66 & 13.02 & 99.16 & 19.25 \\
 & Frozen Detector & $95.66\pm0.00$ & $13.02\pm0.00$ & $99.16\pm0.00$ & $19.25\pm0.00$ \\
 & Joint FT, relation-conditioned & $2.36\pm0.43$ & $1.20\pm0.34$ & $5.04\pm1.11$ & $1.14\pm0.60$ \\
 & Joint FT, direct & $1.10\pm0.03$ & $1.13\pm0.22$ & $2.52\pm0.00$ & $1.74\pm1.12$ \\
\hline
A.X+Gazetteer+CRF & Detection only & 98.11 & 12.70 & 99.16 & 19.55 \\
 & Frozen Detector & $98.11\pm0.00$ & $12.70\pm0.00$ & $99.16\pm0.00$ & $19.55\pm0.00$ \\
 & Joint FT, relation-conditioned & $2.57\pm0.57$ & $1.32\pm0.24$ & $4.48\pm0.64$ & $1.94\pm0.54$ \\
 & Joint FT, direct & $1.41\pm0.09$ & $0.97\pm0.17$ & $3.78\pm0.00$ & $1.99\pm0.74$ \\
\hline
XLM-R-large & Detection only & 98.86 & 14.46 & 99.58 & 17.31 \\
 & Frozen Detector & $87.70\pm1.71$ & $14.57\pm0.08$ & $91.46\pm1.94$ & $17.31\pm0.00$ \\
 & Joint FT, relation-conditioned & 5.94$^\dagger$ & 0.91$^\dagger$ & 9.66$^\dagger$ & 1.49$^\dagger$ \\
 & Joint FT, direct & \multicolumn{4}{c}{budget not met$^\ddagger$} \\
\hline
GPT-4o-mini & LLM, 0-shot & 62.40 & 16.88 & 60.92 & 23.88 \\
 & LLM, 3-shot & 28.34 & 24.28 & 30.25 & 36.27 \\
K-EXAONE & LLM, 0-shot & 95.37 & 13.17 & 97.06 & 19.55 \\
 & LLM, 3-shot & 61.94 & 15.30 & 64.71 & 20.90 \\
Solar Pro 3 & LLM, 0-shot & 80.11 & 15.01 & 86.13 & 20.90 \\
 & LLM, 3-shot & 42.97 & 15.59 & 40.76 & 21.34 \\
\hline
\end{tabular}
\caption{RQ3 masking results (\%). Left: the official 404-document test with generated labels (1,750 KEEP and 11,045 MASK mentions). Right: human-verified entity records of 100 test documents (192 reference-KEEP entities, 238 mentions; 199 reference-MASK entities, 670 mentions). Each row is the same system on both sides: all conditions start from one KDPII-trained detector (seed 42), and values with $\pm$ are mean and sample SD over contextual seeds 42--44. Joint Fine-tuning (FT) uses the 1.5\% validation leakage budget; Detection only, Frozen Detector, and the single-run LLMs, which receive the A.X+Gazetteer+CRF detection-only candidates, come from the registered runs. Nothing was tuned on the human-verified records. $^\dagger$ The only XLM-R-large run that met the budget (seed 43); $^\ddagger$ no direct-head run met it, so 5 of 6 XLM-R-large runs had no epoch within the budget. Appendix Tables~\ref{tab:rq4-main} and~\ref{tab:human-gold-main} add detector leakage and the same records scored against generated labels, and Table~\ref{tab:human-gold-all} averages Detection only over detector seeds 42--44.}
\label{tab:rq3-main}
\end{table*}

%% file: appendices/synthetic_data_statistics.tex
\begin{table*}[p]
\begin{minipage}{\textwidth}
\section{Dataset Statistics and Definitions}
\label{sec:synthetic-data-statistics}
\label{sec:appendix}
The archive census covers 2,355 cleaned Basic documents and 1,857 shared Basic/ReCoPII-Ko/SPY IDs, distinct from the experimental cohorts. PII counts use sentence-level \texttt{PII\_set} mentions; MASK/KEEP counts use ReCoPII-Ko/SPY entity records, which Basic does not provide. The 1,857 ReCoPII-Ko documents comprise 302 career/education, 318 contract, 334 financial, 246 legal, 299 medical, and 358 support documents. The official split (Section~\ref{sec:dataset}) places every domain, relation label, and entity-type/MASK--KEEP combination that occurs in at least ten source groups in all three partitions.
\par\smallskip
\input{analysis/synthetic_data/domain_overview_table}
\par\smallskip
\input{analysis/synthetic_data/opening_statistics_tables}
\par\medskip
\input{analysis/synthetic_data/opening_type_table}
\end{minipage}
\end{table*}

\input{analysis/data_versions/data_versions_table}

\input{analysis/synthetic_data/type_decisions_table}

\begin{figure*}[p]
\centering
\includegraphics[width=.84\textwidth,keepaspectratio]{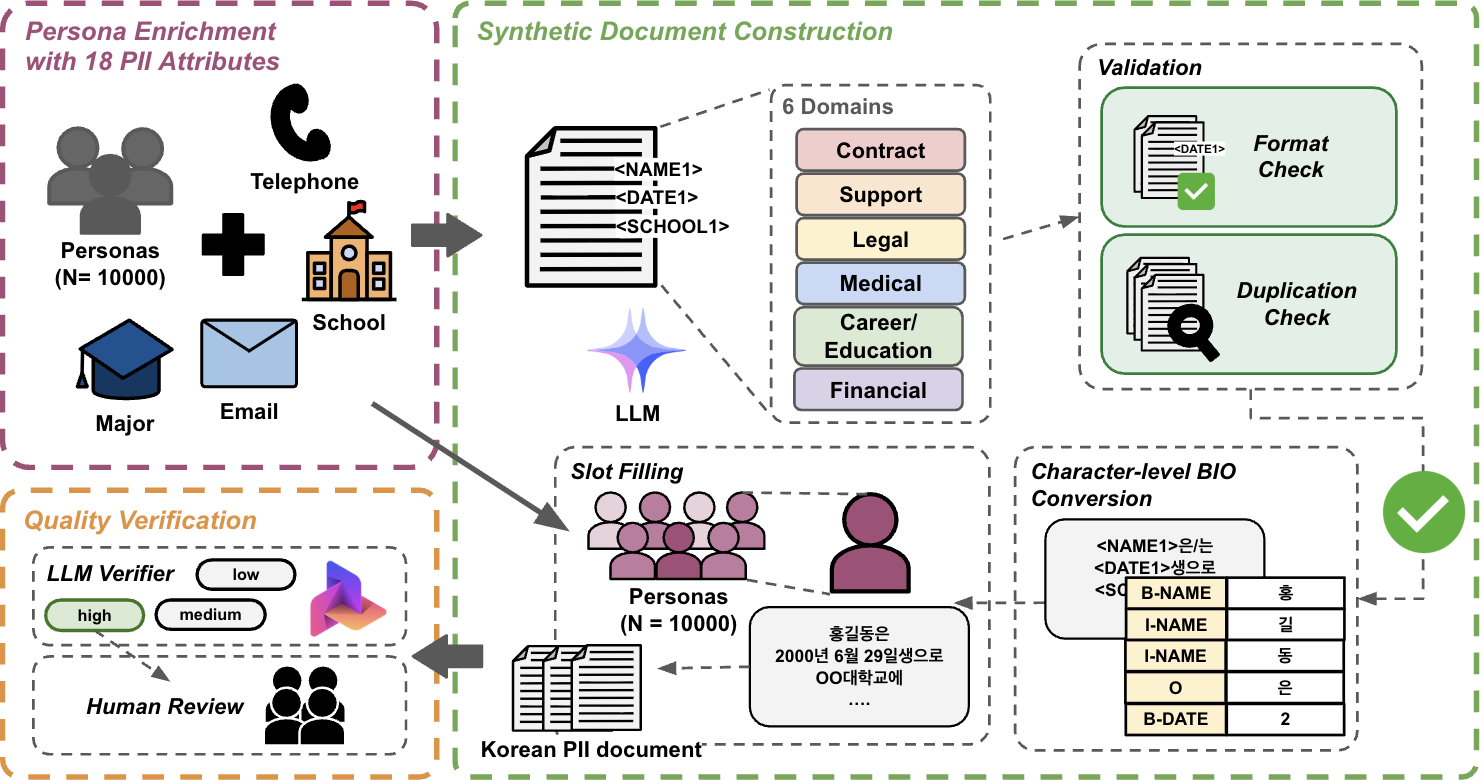}
\caption{Source-corpus construction workflow used before the relation annotation and contextual augmentation described in Section~\ref{sec:method}. Persona enrichment supports placeholder-based document generation across six domains, followed by automated validation, deterministic slot filling, character-level BIO conversion, and quality verification. The later paired-document human audit is reported in Appendix~\ref{sec:additional-evaluation}.}
\label{fig:source-corpus-construction}
\end{figure*}

\input{analysis/synthetic_data/relations_table}
\clearpage

\newcommand{\maskspan}[1]{\colorbox{red!17}{#1}}
\newcommand{\keepspan}[1]{\colorbox{blue!15}{#1}}
\newcommand{\syntheticexamplepanel}[2]{%
  \begingroup
  \setlength{\fboxsep}{5pt}%
  \fcolorbox{black!35}{black!2}{%
    \begin{minipage}{\dimexpr\textwidth-2\fboxsep-2\fboxrule-2pt\relax}
    \footnotesize
    \begin{minipage}[t]{.48\linewidth}
    \raggedright\textbf{Korean source}\par\smallskip
    {\korean\setlength{\fboxsep}{1.5pt}#1\par}
    \end{minipage}\hfill
    \begin{minipage}[t]{.48\linewidth}
    \raggedright\textbf{English translation}\par\smallskip
    #2\par
    \end{minipage}
    \end{minipage}%
  }
  \endgroup
}

\begin{figure*}[p]
\centering
{\small\textbf{(a) Customer and department email}\hfill\texttt{SUPPORT\_200397}}\par\smallskip
\syntheticexamplepanel{%
④ \maskspan{김현석}님의 추가 문의에 따라, 현재 진행 중인 계정 보안 점검 결과를 2025년 7월 15일까지 이메일로 안내드릴 예정입니다. 점검 완료 후에는 \maskspan{kimhyeonseok19650226@naver.com}를 통해 보안 강화 조치에 대한 상세 보고서를 발송하며, 필요 시 추가 인증 절차를 안내해 드리겠습니다.
\par\smallskip
또한, 고객님께서는 \keepspan{eunseong19911009@naver.com}으로 관련 내용을 전달하실 수 있으며, 이 이메일 주소는 해당 부서의 문의 창구입니다.
}{%
(4) In response to additional questions from customer Kim Hyeon-seok, the results of the ongoing account-security review will be emailed by July 15, 2025. Once the review is complete, a detailed report on security improvements will be sent to kimhyeonseok19650226@naver.com, with instructions for further verification if needed.
\par\smallskip
The customer may also send related information to eunseong19911009@naver.com; that address is the department's inquiry channel.
}
\par\medskip
{\small\textbf{(b) Customer and corporate accounts}\hfill\texttt{SUPPORT\_200124}}\par\smallskip

\syntheticexamplepanel{%
[결과 및 메모] 부분 환불 처리가 결정되어 두 번째 주문에 대한 결제 금액을 전액 환불하였으며, 환불 계좌는 기존 결제 시 사용된 계좌인 \maskspan{339-9736-5545-72}로 지정되었다. 또한 앱 오류 재발 방지를 위해 개발팀에 버그 리포트를 제출하고 시스템 개선 요청을 진행하였다. 향후 유사 사례 발생 시 동일한 처리 절차를 적용하도록 내부 지침을 수정하였다.
\par\smallskip
환불 처리에 사용된 계좌는 \keepspan{272-324-012263}이며, 해당 계좌는 \keepspan{STX그린로지스}의 공식 정산 계좌다.
}{%
[Outcome and notes] A partial refund was approved, and the payment for the second order was refunded in full. The refund account was set to 339-9736-5545-72, the account used for the original payment. A bug report was sent to the development team to prevent the app error from recurring, and a system improvement was requested. Internal guidance was updated to apply the same procedure in similar future cases.
\par\smallskip
The account used to process the refund is 272-324-012263, the official settlement account of STX Green Logistics.
}
\par\medskip
{\small\textbf{(c) Service-office and delivery addresses}\hfill\texttt{SUPPORT\_200505}}\par\smallskip

\syntheticexamplepanel{%
고객 지원 부서의 사무실 주소는 \keepspan{서울특별시 강서구 화곡로 296-108}이며, 방문 상담이 가능합니다.
\par\smallskip
고객 \maskspan{장미정}은 반품 접수 외에도 오배송된 상품에 대한 배송비 환급을 요청하였습니다. 물류 비용 정산 기준에 따라 배송비 전액 또는 일부를 환급하는 방안을 검토하고 있습니다.
\par\smallskip
고객 \maskspan{장미정}의 반품 상품은 현재 물류 센터에 입고되어 검수 중입니다. 검수 완료 후 고객의 주소 \maskspan{경기도 평택시 평택로 293 104동 523호}로 재배송이 진행될 예정이며, 배송 추적 번호를 제공할 계획입니다.
}{%
The customer support department's office is at 296-108 Hwagok-ro, Gangseo-gu, Seoul, where in-person consultation is available.
\par\smallskip
Customer Jang Mi-jeong requested reimbursement of shipping costs for a misdelivered item in addition to filing a return. The company is reviewing whether to reimburse the full cost or part of it under its logistics settlement rules.
\par\smallskip
The item returned by Jang Mi-jeong is being inspected at the logistics center. After inspection, it will be shipped again to the customer's address at 293 Pyeongtaek-ro, Building 104, Unit 523, Pyeongtaek-si, Gyeonggi-do, and a tracking number will be provided.
}
\par\medskip
{\small\textbf{(d) Recommendation subject and cited author}\hfill\texttt{CAREER\_EDUCATION\_300930}}\par\smallskip

\syntheticexamplepanel{%
또한, 피추천인은 2023년 하반기에 '제로 트러스트 아키텍처 도입' 프로젝트에 참여하며 기존 네트워크 구조를 전면 개편하는 데 기여했습니다. 이 과정에서 그는 복잡한 정책 설계와 구현 작업을 주도하여 프로젝트 일정을 예정보다 2주 앞당겼습니다.
\par\smallskip
\maskspan{김순근}은 \keepspan{존 킨더바그}가 제안한 제로 트러스트 보안 모델을 설계 근거로 삼아 해당 아키텍처의 정책 구조를 구체화했습니다.
}{%
The recommended candidate also joined a project to adopt a zero-trust architecture in the second half of 2023 and helped overhaul the existing network. He led the design and implementation of complex policies and completed the project two weeks ahead of schedule.
\par\smallskip
Kim Sun-geun used the zero-trust security model proposed by John Kindervag as a design basis and specified the architecture's policy structure.
}
\caption{Four ReCoPII-Ko excerpts showing contextual MASK/KEEP decisions for PII-type mentions. Korean passages reproduce consecutive source records; English passages are translations. Red marks MASK spans in \texttt{PII\_set}, and blue marks KEEP spans in the entity registry. The same type can receive both decisions: \texttt{EMAIL} in (a), \texttt{BANK\_ACCOUNT\_NUMBER} in (b), \texttt{ADDRESS} in (c), and \texttt{NAME} in (d). These are generated labels, not human judgments.}
\label{fig:pii-mask-keep-example}
\end{figure*}
\clearpage

\input{analysis/synthetic_data/pii_type_definitions}

%% file: analysis/synthetic_data/domain_overview_table.tex
\begingroup
\centering
\footnotesize
\setlength{\tabcolsep}{3pt}
\renewcommand{\arraystretch}{1.08}
\begin{tabular*}{\linewidth}{@{}@{\extracolsep{\fill}}lrrrrrr@{}}
\hline
Domain & Basic docs (\%) & Basic PII & ReCoPII-Ko M/K & ReCoPII-Ko KEEP & SPY M/K & SPY KEEP \\
\hline
Career / education & 400 (17.0\%) & 9,504 & 2,472 / 1,065 & 30.1\% & 2,432 / 1,818 & 42.8\% \\
Contract & 386 (16.4\%) & 12,047 & 2,837 / 1,135 & 28.6\% & 2,766 / 2,074 & 42.9\% \\
Financial & 409 (17.4\%) & 12,841 & 4,049 / 1,142 & 22.0\% & 3,989 / 2,286 & 36.4\% \\
Legal & 326 (13.8\%) & 8,355 & 2,716 / 968 & 26.3\% & 2,706 / 1,318 & 32.8\% \\
Medical & 414 (17.6\%) & 10,365 & 2,319 / 1,144 & 33.0\% & 2,310 / 2,108 & 47.7\% \\
Support & 420 (17.8\%) & 10,007 & 2,711 / 1,326 & 32.8\% & 2,692 / 1,999 & 42.6\% \\
\hline
\end{tabular*}
\captionof{table}{Six-domain composition. Basic counts and PII mentions use the full cleaned corpus; ReCoPII-Ko and SPY entity counts use the 1,857 common documents. Each MASK/KEEP share is calculated within its domain, not across domains.}
\label{tab:appendix-domains}
\par\endgroup

%% file: analysis/synthetic_data/opening_statistics_tables.tex
\noindent\begin{minipage}{\textwidth}
\centering\footnotesize
\setlength{\tabcolsep}{2pt}
\renewcommand{\arraystretch}{1.08}
\begin{tabular*}{\linewidth}{@{}@{\extracolsep{\fill}}lrrrrr@{}}
\hline
Measure & Basic all & Basic common & ReCoPII-Ko & SPY & KDPII (raw) \\
\hline
Documents & 2,355 & 1,857 & 1,857 & 1,857 & 4,981 \\
PII mentions & 63,119 & 48,561 & 50,620 & 49,307 & --- \\
Mean (median) PII/doc & 26.8 (25) & 26.2 (24) & 27.3 (25) & 26.6 (24) & --- \\
MASK entities & --- & --- & 17,104 & 16,895 & --- \\
KEEP entities & --- & --- & 6,780 & 11,603 & --- \\
Mean characters & 1930.0 & 1931.5 & 2157.0 & 2063.5 & 282.8 \\
Min / P25 characters & 1,463 / 1,705 & 1,496 / 1,702 & 1,657 / 1,925 & 989 / 1,836 & 107 / 222 \\
Median characters & 1,853 & 1,856 & 2,083 & 1,999 & 270 \\
P75 / P95 characters & 2,052 / 2,539 & 2,051 / 2,552 & 2,290 / 2,820 & 2,188 / 2,674 & 328 / 440 \\
Max characters & 4,869 & 4,329 & 4,642 & 4,505 & 749 \\
\hline
\end{tabular*}
\captionof{table}{Corpus overview and document lengths. Lengths count Unicode characters with stored sentences joined by single newlines and whitespace retained; quantiles use linear interpolation and are rounded to integers. PII counts include repeated sentence-level mentions. MASK/KEEP are entity records and unavailable for Basic. KDPII is an independent length reference from the full raw export before RQ1 duplicate removal.}
\label{tab:appendix-corpus}
\label{tab:appendix-characters}
\end{minipage}

%% file: analysis/synthetic_data/opening_type_table.tex
\begingroup
\centering
\footnotesize
\setlength{\tabcolsep}{3pt}
\renewcommand{\arraystretch}{1.08}
\begin{tabular*}{\linewidth}{@{}@{\extracolsep{\fill}}p{4.3cm}rrrr@{}}
\hline
Type & Basic all & Basic common & ReCoPII-Ko & SPY \\
\hline
\texttt{ADDRESS} & 3,555 (5.6\%) & 2,706 (5.6\%) & 2,693 (5.3\%) & 2,678 (5.4\%) \\
\texttt{AGE} & 839 (1.3\%) & 587 (1.2\%) & 587 (1.2\%) & 609 (1.2\%) \\
\texttt{BANK\_\allowbreak{}ACCOUNT\_\allowbreak{}NUMBER} & 2,172 (3.4\%) & 1,677 (3.5\%) & 1,678 (3.3\%) & 1,680 (3.4\%) \\
\texttt{CARD\_\allowbreak{}NUMBER} & 764 (1.2\%) & 587 (1.2\%) & 586 (1.2\%) & 579 (1.2\%) \\
\texttt{DATE\_\allowbreak{}OF\_\allowbreak{}BIRTH} & 2,913 (4.6\%) & 2,189 (4.5\%) & 2,190 (4.3\%) & 2,167 (4.4\%) \\
\texttt{DEPARTMENT} & 1,633 (2.6\%) & 1,305 (2.7\%) & 1,384 (2.7\%) & 1,475 (3.0\%) \\
\texttt{DRIVER\_\allowbreak{}LICENSE\_\allowbreak{}NUMBER} & 234 (0.4\%) & 166 (0.3\%) & 166 (0.3\%) & 166 (0.3\%) \\
\texttt{EMAIL} & 3,619 (5.7\%) & 2,929 (6.0\%) & 2,930 (5.8\%) & 2,893 (5.9\%) \\
\texttt{MAJOR} & 886 (1.4\%) & 624 (1.3\%) & 624 (1.2\%) & 678 (1.4\%) \\
\texttt{MOBILE\_\allowbreak{}PHONE} & 4,616 (7.3\%) & 3,555 (7.3\%) & 3,554 (7.0\%) & 3,431 (7.0\%) \\
\texttt{NAME} & 30,533 (48.4\%) & 23,672 (48.7\%) & 23,761 (46.9\%) & 24,178 (49.0\%) \\
\texttt{PASSPORT\_\allowbreak{}NUMBER} & 87 (0.1\%) & 79 (0.2\%) & 79 (0.2\%) & 79 (0.2\%) \\
\texttt{POSITION} & 1,169 (1.9\%) & 928 (1.9\%) & 926 (1.8\%) & 1,012 (2.1\%) \\
\texttt{RRN} & 1,608 (2.5\%) & 1,176 (2.4\%) & 1,176 (2.3\%) & 1,175 (2.4\%) \\
\texttt{SCHOOL} & 1,501 (2.4\%) & 848 (1.7\%) & 927 (1.8\%) & 928 (1.9\%) \\
\texttt{TELEPHONE} & 2,324 (3.7\%) & 1,860 (3.8\%) & 1,822 (3.6\%) & 1,787 (3.6\%) \\
\texttt{VEHICLE\_\allowbreak{}NUMBER} & 349 (0.6\%) & 256 (0.5\%) & 256 (0.5\%) & 257 (0.5\%) \\
\texttt{WORKPLACE} & 4,317 (6.8\%) & 3,417 (7.0\%) & 5,281 (10.4\%) & 3,535 (7.2\%) \\
\hline
\end{tabular*}
\captionof{table}{PII mention composition by stored entity type. Each cell gives the count and its share of all PII\_set mentions in that corpus. Basic (all) is shown separately because it is not the paired comparison cohort.}
\label{tab:appendix-type-mentions}
\par\endgroup

%% file: analysis/data_versions/data_versions_table.tex
\begin{table*}[t]
\centering\footnotesize
\setlength{\tabcolsep}{3pt}
\renewcommand{\arraystretch}{1.12}
\begin{tabular}{@{}p{0.2\linewidth}p{0.25\linewidth}p{0.1\linewidth}p{0.15\linewidth}p{0.24\linewidth}@{}}
\hline
\raggedright Experiment & \raggedright Data version & \raggedright Train / valid / test & \raggedright Split unit & \raggedright Evaluation set \tabularnewline
\hline
\raggedright Dataset census (Section~\ref{sec:dataset}, App.~\ref{sec:synthetic-data-statistics}) & \raggedright Basic \texttt{4ea8b3045d57}; ReCoPII-Ko \texttt{6181e9b1fbb0}; SPY \texttt{8fe4d13913c4} & \raggedright --- & \raggedright --- & \raggedright 2,355 Basic; 1,857 common IDs \tabularnewline
\raggedright RQ1: KDPII detection & \raggedright KDPII (19 labels) & \raggedright 3,586 / 399 / 996 & \raggedright Document; no exact sentence in two splits & \raggedright 996 documents \tabularnewline
\raggedright RQ1 synthetic transfer and RQ3: integrated masking & \raggedright ReCoPII-Ko \texttt{6181e9b1fbb0}; split \texttt{d0d8539fe839} & \raggedright 1,245 / 208 / 404 & \raggedright Source group (1,744 groups); official split & \raggedright 404 documents; 11,045 MASK, 1,750 KEEP mentions \tabularnewline
\raggedright RQ2-A: LLM and human quality & \raggedright ReCoPII-Ko \texttt{6181e9b1fbb0}; SPY \texttt{8fe4d13913c4} & \raggedright --- & \raggedright --- & \raggedright 300 ReCoPII-Ko/SPY pairs \tabularnewline
\raggedright RQ2-B and its context analysis: MASK/KEEP classification & \raggedright Earlier ReCoPII-Ko snapshot (\texttt{212f4d8f1c3d}); SPY \texttt{8fe4d13913c4}; split \texttt{81defafc6aba} & \raggedright 1,392 / 186 / 279 & \raggedright Document ID, stratified by domain & \raggedright 279 source documents in both versions; 7,891 + 7,713 MASK and 1,211 + 1,211 KEEP mentions (ReCoPII-Ko + SPY, RQ2-B) \tabularnewline
\hline
\end{tabular}
\caption{Data version and split used by each experiment. Hashes are SHA-256 prefixes of the archive or split file. The source-group split links documents that share a Basic source identifier or identical text and assigns each linked group to one split. The RQ2-B split assigns each document ID to one of 1,392 training, 186 validation, and 279 test documents, so no document belongs to two partitions. It predates the source-group split and does not group documents by source: 27 of its test documents have the same Basic source identifier value as a training (24) or validation (3) document, which gives 28 document pairs. The documents in these pairs have different texts: none of the pairs is identical, and their character 5-gram Jaccard similarity (median 0.045, maximum 0.086) is close to that of random same-domain pairs (median 0.038, 95th percentile 0.071); 2 of the pairs have entity values in common (8 at most). The earlier ReCoPII-Ko snapshot (stored as \texttt{ours\_fixed} in our files) applies a local label fix to the pre-correction ReCoPII-Ko archive, whose hash its row gives; relative to the current archive, all 1,857 document texts and relation annotations are identical, and \texttt{PII\_set} differs in 2 documents (50,622 versus 50,620 mentions).}
\label{tab:appendix-data-versions}
\end{table*}

%% file: analysis/synthetic_data/type_decisions_table.tex
\begin{table*}[t]
\centering
\small
\setlength{\tabcolsep}{4pt}
\renewcommand{\arraystretch}{1.05}
\begin{tabular}{lrrrrrrrr}
\hline
& \multicolumn{5}{c}{ReCoPII-Ko} & \multicolumn{3}{c}{SPY} \\
\cline{2-6}\cline{7-9}
& \multicolumn{2}{c}{Mentions} & \multicolumn{3}{c}{Entities} & \multicolumn{3}{c}{Entities} \\
\cline{2-3}\cline{4-6}\cline{7-9}
Entity type & MASK & KEEP & MASK & KEEP & KEEP (\%) & MASK & KEEP & KEEP (\%) \\
\hline
\texttt{ADDRESS} & 2,693 & 709 & 2,162 & 697 & 24.4 & 2,149 & 1,080 & 33.4 \\
\texttt{AGE} & 587 & 0 & 525 & 0 & 0.0 & 524 & 1 & 0.2 \\
\texttt{BANK\_ACCOUNT\_NUMBER} & 1,678 & 748 & 1,041 & 738 & 41.5 & 1,024 & 1,441 & 58.5 \\
\texttt{CARD\_NUMBER} & 586 & 2 & 361 & 1 & 0.3 & 357 & 1 & 0.3 \\
\texttt{DATE\_OF\_BIRTH} & 2,190 & 0 & 1,576 & 0 & 0.0 & 1,573 & 0 & 0.0 \\
\texttt{DEPARTMENT} & 1,384 & 1,200 & 773 & 888 & 53.5 & 734 & 1,419 & 65.9 \\
\texttt{DRIVER\_LICENSE\_NUMBER} & 166 & 0 & 138 & 0 & 0.0 & 138 & 0 & 0.0 \\
\texttt{EMAIL} & 2,930 & 1,777 & 1,233 & 1,653 & 57.3 & 1,220 & 1,241 & 50.4 \\
\texttt{MAJOR} & 624 & 0 & 309 & 0 & 0.0 & 307 & 631 & 67.3 \\
\texttt{MOBILE\_PHONE} & 3,554 & 3 & 1,987 & 3 & 0.2 & 1,922 & 1,166 & 37.8 \\
\texttt{NAME} & 23,761 & 58 & 2,667 & 58 & 2.1 & 2,667 & 1 & 0.0 \\
\texttt{PASSPORT\_NUMBER} & 79 & 0 & 72 & 0 & 0.0 & 72 & 0 & 0.0 \\
\texttt{POSITION} & 926 & 30 & 728 & 27 & 3.6 & 702 & 1,177 & 62.6 \\
\texttt{RRN} & 1,176 & 0 & 1,044 & 0 & 0.0 & 1,044 & 0 & 0.0 \\
\texttt{SCHOOL} & 927 & 1 & 304 & 1 & 0.3 & 302 & 1,036 & 77.4 \\
\texttt{TELEPHONE} & 1,822 & 2,040 & 1,035 & 1,851 & 64.1 & 1,017 & 803 & 44.1 \\
\texttt{VEHICLE\_NUMBER} & 256 & 0 & 159 & 0 & 0.0 & 156 & 838 & 84.3 \\
\texttt{WORKPLACE} & 5,281 & 1,678 & 990 & 863 & 46.6 & 987 & 768 & 43.8 \\
\hline
\end{tabular}
\caption{MASK/KEEP composition by entity type on the 1,857 common documents. Mentions count sentence-level occurrences in the ReCoPII-Ko entity registry, including repeats; the ReCoPII-Ko MASK mentions are exactly the \texttt{PII\_set} entries, which contain no KEEP mention. Entities count document-level registry records, and KEEP (\%) uses MASK+KEEP entities within each type. These are dataset labels rather than model predictions.}
\label{tab:dataset-type-decisions}
\label{tab:appendix-type-entities}
\end{table*}

%% file: analysis/synthetic_data/relations_table.tex
\begin{table*}[p]
\centering
\footnotesize
\setlength{\tabcolsep}{3pt}
\renewcommand{\arraystretch}{1.08}
\begin{tabular}{p{7.1cm}rrrr}
\hline
Relation label & Edges & PII & NON\_PII & NON\_PII \% \\
\hline
\texttt{ASSET\_\allowbreak{}RELATION} & 298 & 297 & 1 & 0.3\% \\
\texttt{BUSINESS\_\allowbreak{}SERVICE\_\allowbreak{}RELATION} & 410 & 402 & 8 & 2.0\% \\
\texttt{CASE\_\allowbreak{}AND\_\allowbreak{}INCIDENT\_\allowbreak{}RELATION} & 399 & 394 & 5 & 1.3\% \\
\texttt{COLLABORATION\_\allowbreak{}PARTNERSHIP} & 153 & 119 & 34 & 22.2\% \\
\texttt{CONTACT\_\allowbreak{}ASSOCIATION} & 8,100 & 4,432 & 3,668 & 45.3\% \\
\texttt{CREDIT\_\allowbreak{}AND\_\allowbreak{}GUARANTEE\_\allowbreak{}RELATION} & 64 & 64 & 0 & 0.0\% \\
\texttt{EDUCATION\_\allowbreak{}STATUS} & 951 & 656 & 295 & 31.0\% \\
\texttt{EVENT\_\allowbreak{}PARTICIPATION} & 3 & 3 & 0 & 0.0\% \\
\texttt{FINANCIAL\_\allowbreak{}ASSET\_\allowbreak{}ASSOCIATION} & 2,200 & 1,457 & 743 & 33.8\% \\
\texttt{FINANCIAL\_\allowbreak{}TRANSACTION} & 1,013 & 1,007 & 6 & 0.6\% \\
\texttt{GUIDANCE\_\allowbreak{}SUPPORT} & 134 & 125 & 9 & 6.7\% \\
\texttt{IDENTITY\_\allowbreak{}ASSOCIATION} & 3,391 & 3,391 & 0 & 0.0\% \\
\texttt{INFORMATION\_\allowbreak{}GOVERNANCE} & 470 & 432 & 38 & 8.1\% \\
\texttt{INSURANCE\_\allowbreak{}RELATION} & 61 & 61 & 0 & 0.0\% \\
\texttt{LOCATION\_\allowbreak{}ASSOCIATION} & 2,855 & 2,148 & 707 & 24.8\% \\
\texttt{ORGANIZATIONAL\_\allowbreak{}RELATION} & 4,860 & 2,457 & 2,403 & 49.4\% \\
\texttt{PERSONAL\_\allowbreak{}RELATIONSHIP} & 64 & 64 & 0 & 0.0\% \\
\texttt{PROCEDURAL\_\allowbreak{}RELATION} & 165 & 160 & 5 & 3.0\% \\
\texttt{QUALIFICATION\_\allowbreak{}AND\_\allowbreak{}ASSESSMENT} & 65 & 61 & 4 & 6.2\% \\
\hline
\end{tabular}
\par\vspace{0.45em}
\begin{tabular}{lrrr}\hline Final endpoint decisions & MASK/MASK & MASK/KEEP & KEEP/KEEP \\ \hline All ReCoPII-Ko edges & 19,824 & 2,723 & 3,109 \\ \hline\end{tabular}
\caption{ReCoPII-Ko relation-annotation composition (25,656 edges). Edge-level privacy labels and final endpoint masking decisions are distinct; the lower panel counts the latter combinations. Directional endpoint-type configurations are retained in the reproducible relation-configuration artifact.}
\label{tab:appendix-relations}
\end{table*}

%% file: analysis/synthetic_data/pii_type_definitions.tex
\begin{table*}[!t]
\centering
\footnotesize
\setlength{\tabcolsep}{3pt}
\renewcommand{\arraystretch}{1.08}
\begin{tabular}{p{1.25cm}p{3.05cm}p{4.4cm}p{6.25cm}}
\hline
Family & Stored type & Definition (Korean / English) & Format constraints (observed) \\
\hline
PER & \texttt{NAME} & {\korean 이름} / Person name & \emph{Free text} \\
\hline
LOC & \texttt{ADDRESS} & {\korean 주소} / Street, building, or residential address & \emph{Free text} \\
\hline
ORG & \texttt{WORKPLACE} & {\korean 직장} / Employer or workplace & \emph{Free text} \\
 & \texttt{DEPARTMENT} & {\korean 부서} / Organizational department & \emph{Free text} \\
 & \texttt{POSITION} & {\korean 직위} / Job title or organizational position & \emph{Free text} \\
 & \texttt{SCHOOL} & {\korean 학교} / School or educational institution & \emph{Free text} \\
\hline
DEM & \texttt{MAJOR} & {\korean 전공} / Academic major or field of study & \emph{Free text} \\
 & \texttt{AGE} & {\korean 나이} / Age or age expression & \verb!\d{2}! \\
\hline
DATE\-TIME & \texttt{DATE\_\allowbreak{}OF\_\allowbreak{}BIRTH} & {\korean 생년월일} / Date of birth & \verb!\d{4}[-./]\d{1,2}[-./]\d{1,2}!\newline \verb!\d{8}!\newline \emph{also Korean year/month/day wording} \\
\hline
CODE & \texttt{MOBILE\_\allowbreak{}PHONE} & {\korean 휴대폰} / Mobile telephone number & \verb!010-\d{4}-\d{4}! \\
 & \texttt{TELEPHONE} & {\korean 전화} / Non-mobile or general telephone number & \verb!0(?:2|[3-9]\d)-\d{3,4}-\d{4}! \\
 & \texttt{PASSPORT\_\allowbreak{}NUMBER} & {\korean 여권번호} / Passport identifier & \verb![A-Z]\d{3}[A-Z]\d{4}! \\
 & \texttt{RRN} & {\korean 주민번호} / Korean resident registration number & \verb!\d{6}-[1-4]\d{6}! \\
 & \texttt{DRIVER\_\allowbreak{}LICENSE\_\allowbreak{}NUMBER} & {\korean 면허번호} / Driver's license identifier & \verb!\d{2}-\d{2}-\d{6}-\d{2}! \\
 & \texttt{VEHICLE\_\allowbreak{}NUMBER} & {\korean 차량번호} / Vehicle registration number & \verb!\d{3}[\uAC00-\uD7A3]\d{4}! \\
 & \texttt{BANK\_\allowbreak{}ACCOUNT\_\allowbreak{}NUMBER} & {\korean 계좌} / Bank account identifier & \verb!\d{2,6}(?:-\d{2,6}){2,3}!\newline \emph{bank-specific digit groups} \\
 & \texttt{CARD\_\allowbreak{}NUMBER} & {\korean 카드번호} / Payment card number & \verb!\d{4}(?:-\d{4}){3}! \\
 & \texttt{EMAIL} & {\korean 이메일} / Email address & \verb![^\s@]+@[^\s@]+\.[^\s@]+! \\
\hline
\end{tabular}
\par\vspace{0.8em}
\begin{tabular}{lrr}
\hline
NAME entities per document & ReCoPII-Ko & SPY \\
\hline
Documents & 1,857 & 1,857 \\
Mean NAME entities / document & 1.47 & 1.44 \\
Median & 1 & 1 \\
Minimum & 1 & 1 \\
Maximum & 3 & 3 \\
\hline
\end{tabular}
\caption{The 18 stored PII entity types, Korean label names, and observed format cues. Structured-type patterns summarize Basic PII\_set forms and ReCoPII-Ko entity mention forms in the current snapshots; birth dates also appear in Korean year/month/day wording. The patterns are not verified generation-time validators and need not cover SPY variants. Free-text types have no single surface regex. The six families are display groups, not annotation labels. The lower panel reports NAME entity records per document, not distinct people.}
\label{tab:pii-type-definitions}
\end{table*}

%% file: appendices/synthetic_data_evaluation_templates.tex
\section{Generation and Evaluation Details}
\label{sec:additional-evaluation}
Measured processing counts and evaluation results are grouped below by purpose.

\subsection{Generation and annotation repair}
Table~\ref{tab:generation-attrition} reports the document flow of the organization-focused augmentation, and Table~\ref{tab:generation-correction-stages} the later annotation repair.

\begin{table*}[!t]
\begin{minipage}{\textwidth}
\captionsetup{skip=4pt}
\noindent\begin{minipage}{\linewidth}
\centering\footnotesize
\setlength{\tabcolsep}{2pt}
\renewcommand{\arraystretch}{1.0}
\input{analysis/synthetic_data/augmentation_attrition/attrition_table}
\captionof{table}{Document flow of the organization-focused augmentation. Documents that failed were regenerated with a new seed in up to five rounds, and each rejected document is counted once at the stage of its last failure. The review step rejects no document itself; its findings enter the acceptance decision (Section~\ref{sec:independent-review}). The 147 realization failures include five documents accepted before the competing-source check of Appendix~\ref{app:construction-realization} was introduced that failed it on re-examination. Counts come from the construction team's summary report (SHA-256 prefix \texttt{b61b19e16118}); its per-document audits are stored on a separate server. The final count matches the hashed ReCoPII-Ko and SPY archives.}
\label{tab:generation-attrition}
\end{minipage}
\input{analysis/synthetic_data/generation_correction_table}

\end{minipage}
\end{table*}

\subsection{Document quality and human validation}
To calibrate the 1--5 naturalness scale of the LLM judges, the same judges also score 300 length-stratified English student essays from the 6,807-document public training split of the Kaggle PII Detection dataset \citep{holmes2024piidetection}; their mean naturalness is 2.74, against 2.77 for ReCoPII-Ko and 2.37 for SPY.

Four annotators, all project members, independently assess the same pool of 300 ReCoPII-Ko/SPY document pairs under a fixed masking policy, with pair order and A/B positions randomized separately for each annotator. Both are drawn once when the study is prepared and stored with the annotator's assignment, so an annotator who reconnects sees the same order and positions. The items to judge are also fixed in advance: for each document, a seeded random draw selects up to two MASK and up to two KEEP entity records, and for each ReCoPII-Ko document up to two relations drawn uniformly from all its edges; all four annotators judge the same items.

\begin{figure*}[p]
\centering
\includegraphics[width=0.70\textwidth,keepaspectratio]{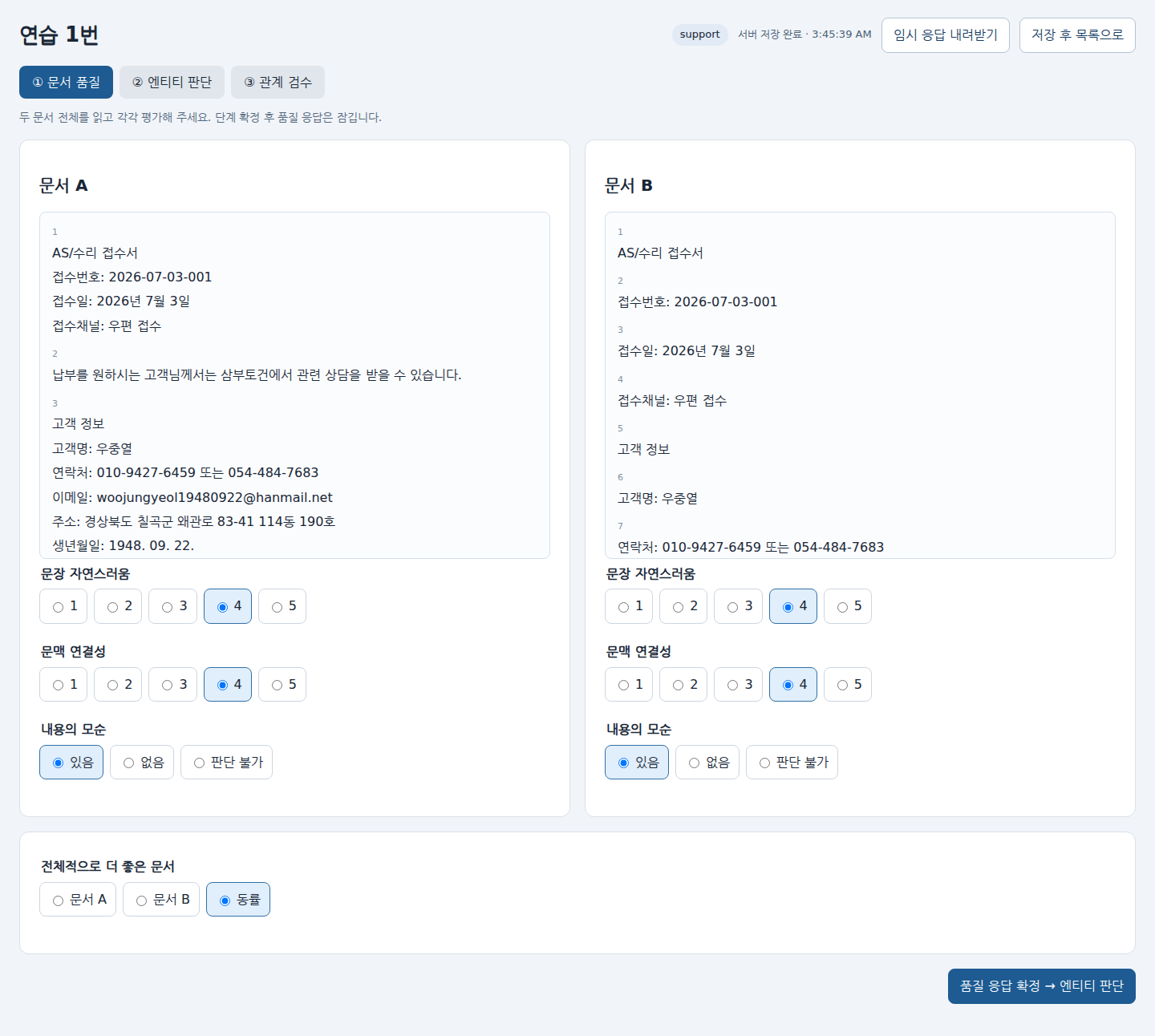}\\[2pt]
{\small (a) Document quality}\\[8pt]
\includegraphics[width=0.70\textwidth,keepaspectratio]{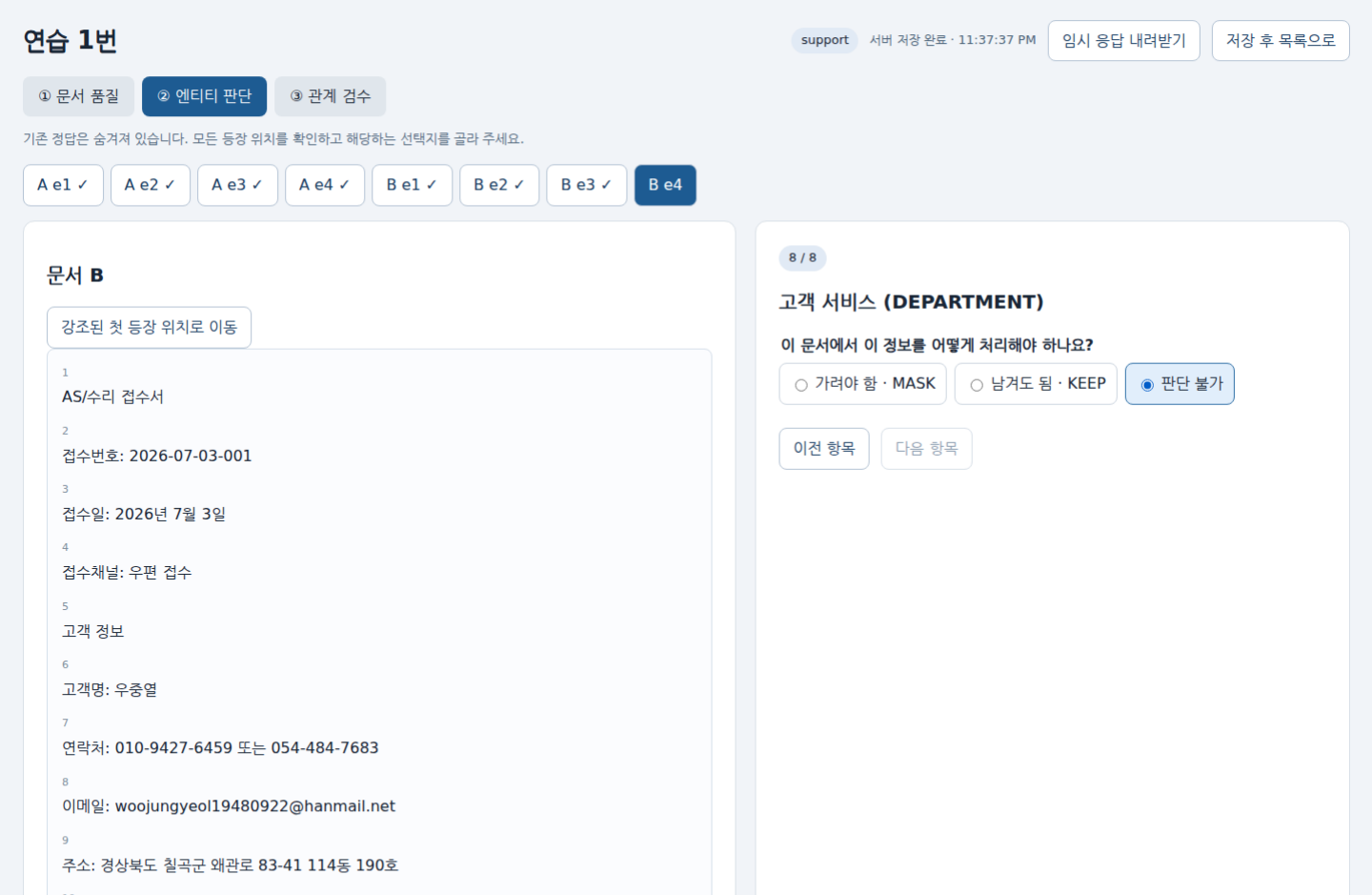}\\[2pt]
{\small (b) Entity MASK/KEEP judgment with labels hidden}
\caption{Human evaluation interface, shown on a practice item (\emph{yeonseup 1-beon}, ``Practice 1'') with synthetic document values; the selections come from an interface check and are not study responses. Both screens are cropped. (a) The instruction reads ``Read both documents in full and rate each. Quality responses are locked after the stage is confirmed.'' Under Document A and Document B, the annotator scores ``sentence naturalness'' and ``contextual coherence'' from 1 to 5, marks ``contradiction in content'' as present, absent, or cannot judge, and chooses the ``overall better document'' (Document A, Document B, or tie); the button reads ``confirm quality responses $\rightarrow$ entity judgment.'' (b) The tabs name the three stages (document quality, entity judgment, relation review), and the instruction reads ``The existing answers are hidden. Check every occurrence and choose the applicable option.'' For each sampled entity, shown with its value and entity type, the question ``How should this information be handled in this document?'' offers ``must be hidden (MASK),'' ``may be left (KEEP),'' and ``cannot judge,'' which is recorded as UNCERTAIN. The remaining buttons read ``go to the first highlighted occurrence,'' ``previous item,'' ``next item,'' ``download draft responses,'' and ``save and return to the list.''}
\label{fig:human-eval-interface}
\end{figure*}

Each pair is evaluated in three stages (Figure~\ref{fig:human-eval-interface}). Annotators first rate naturalness and document coherence on a 1--5 scale and select a preferred document, then decide MASK, KEEP, or UNCERTAIN for the sampled entity records, and finally review the sampled ReCoPII-Ko relations. The interface cannot reveal labels in the first two stages because the server does not send them: in the quality stage it sends only the sentences of the two documents, and in the entity stage it adds the sampled entity records without their stored privacy labels or source identifiers. The generation method is not sent at any stage, and relations are sent only in the relation stage, without their edge-level privacy labels. Once an annotator confirms a stage, the server rejects later changes to its responses, so quality and entity decisions cannot be revised after relations appear. Annotators also do not see the Basic source. The LLM source-coherence and NON\_PII-validity tasks in Table~\ref{tab:rq2a-quality} therefore differ from these human tasks.

The final snapshot contains 1,200 submitted pair reviews: each of the four annotators completed all 300 pairs, which come from 285 original-source groups. Tables~\ref{tab:human-quality-template}--\ref{tab:human-relation-audit} pool all submitted judgments; practice responses are excluded. Each rating has equal weight, and every pair contributes four ratings. We form 95\% confidence intervals using 2,000 bootstrap samples of original-source groups within domains, keeping derivative pairs and their ratings together. These intervals condition on the four annotators. Because every pair has all four submissions, the planned analysis of pairs completed by all four annotators coincides with these estimates.

\input{analysis/human_eval/human_quality}
\input{analysis/human_eval/human_labels}
\input{analysis/human_eval/human_relations}

Agreement with a generated label measures how often a human decision supports that annotation. It does not supply independently adjudicated reference labels. Adjudicated reference labels are not yet available, so the audit reports agreement and uncertainty rather than correction rates or human F1.

\subsection{Human-verified test}
\label{sec:human-gold-test}
Generated labels can agree with a model for reasons unrelated to their correctness, so Section~\ref{sec:rq4} also scores its systems on entity records whose labels annotators verified. The human-verified test consists of 100 ReCoPII-Ko/SPY document pairs from the test partition of the official split. 73 of them are the pairs of the 300-pair pool above that fall in this partition. The other 27 were added afterwards and judged by the same four annotators with the same protocol and interface: they were drawn at random within domains from test-partition source groups that contain no pair of the pool and whose documents are also test documents under the split of RQ2-B, by a rule fixed before any model output on these documents was inspected. As a result, 34 of the 100 pairs are held out under both splits.

As in the audit above, each annotator decides MASK, KEEP, or UNCERTAIN for up to two stored-MASK and up to two stored-KEEP entity records per document without seeing the stored labels, considering all occurrences of the entity. A record's reference label is the MASK or KEEP decision given by at least three of the four annotators; a record without such a majority has no reference label and is not scored. We chose this rule after inspecting how many records each candidate rule yields from the annotators' votes and before any system output was compared with the records. Table~\ref{tab:human-gold-composition} gives the outcome for the 794 records. In the ReCoPII-Ko documents, 391 of the 400 records receive a reference label, which yields the 192 reference-KEEP and 199 reference-MASK entities used for scoring. For the stored-KEEP records of SPY documents, the annotators confirm KEEP for 24 of 194, choose MASK for 46, and reach no majority for 124.

The unit of this test is a sampled entity record. It verifies at most four entities per document and not every mention, so over-masking and leakage in Section~\ref{sec:rq4} are restricted to the mentions of entities that have a reference label in the ReCoPII-Ko version of each document.

\input{analysis/human_gold_test/composition_table}

\subsection{MASK/KEEP prediction}
\label{sec:rq2b-appendix}
Table~\ref{tab:mask-keep-evaluation-template} gives the RQ2-B training pools and common test, Table~\ref{tab:rq2b-masking-full} the results of Table~\ref{tab:rq2b-masking} with standard deviations, Table~\ref{tab:mask-keep-mask-class} the class-wise results, and Table~\ref{tab:rq3-context} the context analysis. Training XLM-R-large on ReCoPII-Ko raises its common-test PII leakage from 0.09\% to 0.28\%, a difference of 0.18 points (95\% CI [0.09, 0.28]). For the context analysis, we train the same six architectures under four inputs, holding each model's split and training procedure fixed: the entity type only, its surface value plus type, the containing sentence, and the full document. Mention-level MASK probabilities are averaged per entity and thresholded at 0.5, and we score entity-level Macro-F1 over seeds 42--44. The preregistered measure is the full-document gain over value plus type; because KLUE-R-base and XLM-R-large read the full document as a 512-token window, we also report the sentence gain, whose input is identical for every architecture. The full-document gain is 16.27--19.55 points for ReCoPII-Ko and 1.63--2.39 points for SPY, and the sentence gain is 15.32--19.03 versus 1.42--2.16 points, with every document-paired interval excluding zero. Value plus type already reaches 96.80--97.31 Macro-F1 on SPY but only 79.60--83.09 on ReCoPII-Ko. Entity value and type therefore determine most SPY decisions, whereas ReCoPII-Ko decisions depend on local textual context, most of which the containing sentence already provides.

\begin{table*}[t]
\centering
\small
\setlength{\tabcolsep}{4pt}
\begin{tabular}{lcccccc}
\hline
 & \multicolumn{2}{c}{Macro-F1 $\uparrow$} & \multicolumn{2}{c}{Over-mask $\downarrow$} & \multicolumn{2}{c}{Leakage $\downarrow$} \\
Model & SPY & ReCoPII-Ko & SPY & ReCoPII-Ko & SPY & ReCoPII-Ko \\
\hline
A.X & $82.68\pm0.56$ & $\mathbf{86.50\pm1.12}$ & $47.01\pm1.20$ & $\mathbf{38.18\pm3.18}$ & $0.12\pm0.05$ & $0.14\pm0.12$ \\
A.X+Gaz & $82.73\pm0.33$ & $\mathbf{85.26\pm0.84}$ & $46.86\pm0.84$ & $\mathbf{41.18\pm1.76}$ & $0.13\pm0.04$ & $0.13\pm0.05$ \\
A.X+CRF & $82.23\pm0.74$ & $\mathbf{85.13\pm0.93}$ & $47.91\pm1.80$ & $\mathbf{41.45\pm2.71}$ & $0.13\pm0.06$ & $0.13\pm0.14$ \\
A.X+Gaz+CRF & $82.35\pm0.78$ & $\mathbf{85.69\pm0.82}$ & $47.66\pm1.77$ & $\mathbf{39.94\pm2.37}$ & $0.13\pm0.05$ & $0.19\pm0.13$ \\
KLUE-R-base$^\dagger$ & $81.17\pm0.92$ & $\mathbf{82.97\pm0.24}$ & $49.75\pm2.38$ & $\mathbf{45.79\pm0.89}$ & $0.24\pm0.13$ & $0.27\pm0.11$ \\
XLM-R-large$^\dagger$ & $80.26\pm0.24$ & $\mathbf{83.54\pm0.76}$ & $52.16\pm0.41$ & $\mathbf{44.49\pm1.96}$ & $0.09\pm0.02$ & $0.28\pm0.07$ \\
\hline
\end{tabular}
\caption{RQ2-B mention-level results (\%) on the matched common test, as mean $\pm$ standard deviation over three seeds for classifiers trained on SPY-style or ReCoPII-Ko documents. Gaz abbreviates Gazetteer. Bold marks the better value within each model. $^\dagger$ marks configurations that failed the preregistered validity check; they are descriptive and excluded from the inferential conclusion. Table~\ref{tab:mask-keep-mask-class} gives class-wise precision, recall, and F1.}
\label{tab:rq2b-masking-full}
\end{table*}

\begin{table*}[!t]
\begin{minipage}{\textwidth}
\captionsetup{skip=4pt}
\noindent\begin{minipage}{\linewidth}
\centering\footnotesize
\setlength{\tabcolsep}{3pt}
\renewcommand{\arraystretch}{1.0}
\begin{tabular*}{\linewidth}{@{}@{\extracolsep{\fill}}p{8cm}rr@{}}
\hline
Matched experiment & ReCoPII-Ko train & SPY train \\
\hline
Train source IDs & 1,392 & 1,392 \\
Train MASK mentions & 37,418 & 36,362 \\
Train KEEP mentions & 4,140 & 4,140 \\
Test source IDs & 279 & 279 \\
Test MASK mentions & 15,604 & 15,604 \\
Test KEEP mentions & 2,422 & 2,422 \\
\hline
\end{tabular*}
\captionof{table}{RQ2-B matched training pools and common test for all six models in Table~\ref{tab:rq2b-masking}. The experiment on the earlier ReCoPII-Ko snapshot uses supplied mention spans and types with generated labels; these are not human-verified gold or the current shared ZIP census. Both training conditions score the same ReCoPII-Ko/SPY paired-version test mentions.}
\label{tab:mask-keep-evaluation-template}
\end{minipage}
\input{analysis/synthetic_data/rq2b_all_models_tables}

\end{minipage}
\end{table*}

\begin{table*}[t]
\centering
\footnotesize
\setlength{\tabcolsep}{2.5pt}
\renewcommand{\arraystretch}{1.2}
\begin{tabular}{llcccccc}
\hline
Model & Data & Type only & Value + type & Sentence & Full doc. & Sentence gain & Full-doc. gain \\
\hline
A.X & R-Ko & $64.76\pm3.11$ & $81.73\pm0.40$ & $98.64\pm0.05$ & $99.45\pm0.16$ & $+16.92$ {\scriptsize[15.66, 18.20]} & $+17.72$ {\scriptsize[16.47, 19.04]} \\
 & SPY & $67.24\pm2.03$ & $97.28\pm0.15$ & $98.88\pm0.15$ & $99.38\pm0.11$ & $+1.60$ {\scriptsize[1.10, 2.08]} & $+2.10$ {\scriptsize[1.61, 2.59]} \\
\hline
A.X+Gaz & R-Ko & $66.55\pm0.00$ & $82.44\pm0.61$ & $98.62\pm0.07$ & $99.40\pm0.04$ & $+16.18$ {\scriptsize[15.04, 17.40]} & $+16.97$ {\scriptsize[15.83, 18.20]} \\
 & SPY & $66.61\pm1.76$ & $97.13\pm0.32$ & $98.89\pm0.19$ & $99.31\pm0.18$ & $+1.76$ {\scriptsize[1.31, 2.23]} & $+2.18$ {\scriptsize[1.72, 2.66]} \\
\hline
A.X+CRF & R-Ko & $68.38\pm4.60$ & $81.10\pm1.25$ & $98.39\pm0.26$ & $99.41\pm0.09$ & $+17.29$ {\scriptsize[16.17, 18.48]} & $+18.32$ {\scriptsize[17.25, 19.49]} \\
 & SPY & $67.78\pm1.09$ & $96.80\pm0.41$ & $98.96\pm0.18$ & $99.19\pm0.02$ & $+2.16$ {\scriptsize[1.72, 2.62]} & $+2.39$ {\scriptsize[1.91, 2.89]} \\
\hline
A.X+Gaz+CRF & R-Ko & $67.89\pm1.18$ & $83.09\pm0.54$ & $98.41\pm0.16$ & $99.36\pm0.10$ & $+15.32$ {\scriptsize[14.07, 16.55]} & $+16.27$ {\scriptsize[15.02, 17.53]} \\
 & SPY & $67.78\pm1.08$ & $96.89\pm0.19$ & $98.99\pm0.08$ & $99.23\pm0.11$ & $+2.10$ {\scriptsize[1.63, 2.58]} & $+2.34$ {\scriptsize[1.88, 2.82]} \\
\hline
KLUE-R-base$^\dagger$ & R-Ko & $66.55\pm0.00$ & $81.02\pm0.62$ & $98.66\pm0.13$ & $98.84\pm0.20$ & $+17.64$ {\scriptsize[16.36, 18.97]} & $+17.82$ {\scriptsize[16.49, 19.18]} \\
 & SPY & $67.24\pm2.03$ & $97.31\pm0.10$ & $98.73\pm0.10$ & $98.94\pm0.04$ & $+1.42$ {\scriptsize[0.75, 1.98]} & $+1.63$ {\scriptsize[1.14, 2.11]} \\
\hline
XLM-R-large$^\dagger$ & R-Ko & $66.55\pm0.00$ & $79.60\pm1.64$ & $98.63\pm0.17$ & $99.15\pm0.14$ & $+19.03$ {\scriptsize[17.85, 20.21]} & $+19.55$ {\scriptsize[18.41, 20.73]} \\
 & SPY & $67.24\pm2.03$ & $97.06\pm0.21$ & $98.87\pm0.11$ & $99.06\pm0.18$ & $+1.82$ {\scriptsize[1.35, 2.30]} & $+2.01$ {\scriptsize[1.54, 2.48]} \\
\hline
\end{tabular}
\caption{RQ2-B context analysis: entity-level Macro-F1 (\%). R-Ko denotes ReCoPII-Ko. Input-condition values are mean $\pm$ standard deviation over three seeds; each gain subtracts Value+type from Sentence or Full doc., with a document-paired bootstrap 95\% confidence interval in brackets. The full-document gain is the preregistered primary measure. $^\dagger$ Full doc. is a 512-token window around each mention.}
\label{tab:rq3-context}
\end{table*}

%% file: analysis/synthetic_data/augmentation_attrition/attrition_table.tex
\begin{tabular*}{\linewidth}{@{}@{\extracolsep{\fill}}p{6.5cm}lrr@{}}
\hline
Stage (final failure) & Method & Removed & Remaining \\
\hline
Basic source pool & --- & --- & 2,355 \\
\quad Relation-graph planning (\texttt{plan}) & \S\ref{app:construction-planning} & 50 &  \\
\quad Text realization (\texttt{sentences}) & \S\ref{app:construction-realization} & 147 &  \\
\quad Value instantiation (\texttt{fill}) & \S\ref{app:construction-instantiation} & 2 &  \\
\quad Review and acceptance (\texttt{accept}) & \S\ref{app:construction-review} & 160 &  \\
Accepted augmentations & --- & 359 & 1,996 (84.8\%) \\
New entity without incident relation & \S\ref{app:construction-review} & 139 & 1,857 (78.9\%) \\
\hline
\end{tabular*}

%% file: analysis/synthetic_data/generation_correction_table.tex
\par\medskip\noindent\begin{minipage}{\linewidth}
\centering
\footnotesize
\setlength{\tabcolsep}{3pt}
\begin{tabular*}{\linewidth}{@{}@{\extracolsep{\fill}}lrrr@{}}
\hline
PII type & Before correction & Final archive & Removed \\
\hline
\texttt{NAME} & 23,819 & 23,761 & 58 \\
\texttt{WORKPLACE} & 5,282 & 5,281 & 1 \\
\texttt{DEPARTMENT} & 1,385 & 1,384 & 1 \\
\texttt{POSITION} & 927 & 926 & 1 \\
\texttt{CARD\_NUMBER} & 588 & 586 & 2 \\
Other 13 types & 18,682 & 18,682 & 0 \\
\hline
Total & 50,683 & 50,620 & 63 \\
\hline
\end{tabular*}
\captionof{table}{PII mention counts across the recoverable final annotation-correction stage. Both snapshots contain the same 1,857 ReCoPII-Ko documents; the correction removed 63 sentence-level \texttt{PII\_set} spans in 58 documents without changing document text, entity registries, or relations. Only changed types are listed separately; the other 13 types did not change. These counts describe annotation repair after augmentation, not generation success or semantic-review attrition.}
\label{tab:generation-correction-stages}
\end{minipage}

%% file: analysis/human_eval/human_quality.tex
\begin{table*}[!t]
\centering\footnotesize
\setlength{\tabcolsep}{3pt}
\renewcommand{\arraystretch}{1.08}
\begin{tabular*}{\textwidth}{@{}@{\extracolsep{\fill}}p{6.1cm}rrr@{}}
\hline
Measure & ReCoPII-Ko & SPY & Tie \\
\hline
Assessed document pairs & 300 & 300 & --- \\
Submitted document ratings & 1200 & 1200 & --- \\
Naturalness (1--5) & 3.24 [3.21, 3.27] & 2.65 [2.61, 2.69] & --- \\
Document coherence (1--5) & 3.15 [3.12, 3.19] & 2.24 [2.19, 2.30] & --- \\
Contradiction rate (\%) & 1.25 [0.66, 1.89] & 14.92 [12.92, 17.00] & --- \\
Contradiction uncertain rate (\%) & 0.00 [0.00, 0.00] & 0.50 [0.17, 0.93] & --- \\
Preference share (\%) & 76.92 [74.07, 79.77] & 7.92 [6.38, 9.66] & 15.17 [13.09, 17.31] \\
\hline
\end{tabular*}
\caption{Human document-quality assessment: four annotators each review all 300 ReCoPII-Ko/SPY pairs (600 document versions), giving 1,200 pair reviews. Brackets give 95\% bootstrap confidence intervals. Each rating has equal weight. Document coherence concerns the completed document; unlike the LLM source-coherence task, annotators do not see the Basic source. Preference shares include ties.}
\label{tab:human-quality-template}
\end{table*}

%% file: analysis/human_eval/human_labels.tex
\begin{table*}[!t]
\centering\footnotesize
\setlength{\tabcolsep}{3pt}
\renewcommand{\arraystretch}{1.08}
\begin{tabular*}{\textwidth}{@{}@{\extracolsep{\fill}}p{6.1cm}rr@{}}
\hline
Measure & ReCoPII-Ko & SPY \\
\hline
Unique sampled entity records / annotators & 1,200 / 4 & 1,191 / 4 \\
Entity judgments (MASK / KEEP) & 2,400 / 2,400 & 2,400 / 2,364 \\
Stored MASK agreement (\%) & 96.29 [95.21, 97.29] & 95.75 [94.44, 96.88] \\
Stored KEEP agreement (\%) & 88.42 [86.50, 90.12] & 26.27 [24.06, 28.47] \\
Stored MASK uncertain rate (\%) & 0.17 [0.04, 0.33] & 0.25 [0.04, 0.50] \\
Stored KEEP uncertain rate (\%) & 0.04 [0.00, 0.13] & 41.71 [39.02, 44.45] \\
Stored KEEP agreement, decidable only (\%) & 88.45 [86.56, 90.14] & 45.07 [41.06, 49.12] \\
Pairwise inter-annotator agreement (\%) & 88.69 & 71.65 \\
Compared entity records / annotator-pair comparisons & 1200 / 7200 & 1191 / 7146 \\
\hline
\end{tabular*}
\caption{Human MASK/KEEP audit of document-level entity records, with 95\% confidence intervals in brackets. Each document supplies up to two stored MASK and two stored KEEP records. Agreement compares human decisions with generated stored labels; UNCERTAIN counts as disagreement unless explicitly excluded. Decidable-only KEEP agreement uses 2,399 ReCoPII-Ko and 1,378 SPY judgments. Pairwise inter-annotator agreement pools all annotator pairs on the same entity, treating UNCERTAIN as a category. These values are neither mention-level F1 nor corpus-wide accuracy. Adjudicated corrections remain unavailable.}
\label{tab:human-label-template}
\end{table*}

%% file: analysis/human_eval/human_relations.tex
\begin{table*}[!t]
\centering\footnotesize
\setlength{\tabcolsep}{3pt}
\renewcommand{\arraystretch}{1.08}
\begin{tabular*}{\textwidth}{@{}@{\extracolsep{\fill}}p{6.1cm}r@{}}
\hline
Measure & ReCoPII-Ko \\
\hline
Unique sampled relations / judgments & 600 / 2400 \\
Relation support rate (\%) & 95.46 [94.44, 96.46] \\
Relation support uncertain rate (\%) & 0.29 [0.12, 0.52] \\
Stored edge-level privacy agreement (\%) & 89.50 [87.76, 91.22] \\
Edge-level privacy uncertain rate (\%) & 3.42 [2.50, 4.34] \\
Pairwise support agreement (\%) & 92.53 \\
Pairwise edge-level privacy agreement (\%) & 85.92 \\
Compared relations / annotator-pair comparisons & 600 / 3600 \\
\hline
\end{tabular*}
\caption{Human relation audit, with 95\% confidence intervals in brackets. Annotators inspect up to two sampled ReCoPII-Ko relations per document after locking quality and entity responses; relation availability makes this stage partially blinded. Support asks whether the text supports the relation. Privacy agreement compares human PII/NON\_PII decisions with generated edge-level privacy labels, including UNCERTAIN as disagreement. SPY relations are outside this audit. Pairwise agreement pools all annotator pairs on the same relation.}
\label{tab:human-relation-audit}
\end{table*}

%% file: analysis/human_gold_test/composition_table.tex
\begin{table}[t]
\centering
\small
\setlength{\tabcolsep}{2pt}
\begin{tabular}{@{}llrrrrr@{}}
\hline
 & Stored & Records & \multicolumn{3}{c}{Reference label} & Agree \\
Version & label &  & MASK & KEEP & none & (\%) \\
\hline
ReCoPII-Ko & MASK & 200 & 195 & 2 & 3 & 96.6 \\
 & KEEP & 200 & 4 & 190 & 6 & 89.6 \\
SPY & MASK & 200 & 193 & 5 & 2 & 96.0 \\
 & KEEP & 194 & 46 & 24 & 124 & 26.8 \\
\hline
\end{tabular}
\caption{Human-verified test: 100 ReCoPII-Ko/SPY document pairs from the official test partition, each judged by four annotators without seeing stored labels (794 sampled entity records, up to two stored-MASK and two stored-KEEP records per document). A record's reference label is the MASK or KEEP decision of at least three of the four annotators; \emph{none} marks records without such a majority. Agree is the share of individual judgments that match the stored label. 34 of the pairs are also held out under the RQ2-B split.}
\label{tab:human-gold-composition}
\end{table}

%% file: analysis/synthetic_data/rq2b_all_models_tables.tex
\par\smallskip\noindent\begin{minipage}{\linewidth}
\centering\fontsize{8}{9}\selectfont
\setlength{\tabcolsep}{2pt}
\renewcommand{\arraystretch}{1.0}
\begin{tabular*}{\linewidth}{@{}@{\extracolsep{\fill}}llrrrrrr@{}}
\hline
 & & \multicolumn{3}{c}{MASK class} & \multicolumn{3}{c}{KEEP class} \\
\cline{3-5}\cline{6-8}
Model & Train & P & R & F1 & P & R & F1 \\
\hline
A.X & SPY & $93.19\pm0.16$ & $99.88\pm0.05$ & $96.42\pm0.09$ & $98.62\pm0.57$ & $52.99\pm1.20$ & $68.93\pm1.03$ \\
 & ReCoPII-Ko & $94.40\pm0.44$ & $99.86\pm0.12$ & $97.05\pm0.17$ & $98.62\pm1.15$ & $61.82\pm3.18$ & $75.95\pm2.07$ \\
\hline
A.X+Gaz & SPY & $93.21\pm0.11$ & $99.87\pm0.04$ & $96.43\pm0.05$ & $98.47\pm0.45$ & $53.14\pm0.84$ & $69.02\pm0.62$ \\
 & ReCoPII-Ko & $93.99\pm0.24$ & $99.87\pm0.05$ & $96.84\pm0.15$ & $98.63\pm0.56$ & $58.82\pm1.76$ & $73.69\pm1.53$ \\
\hline
A.X+CRF & SPY & $93.07\pm0.24$ & $99.87\pm0.06$ & $96.35\pm0.10$ & $98.43\pm0.66$ & $52.09\pm1.80$ & $68.11\pm1.38$ \\
 & ReCoPII-Ko & $93.95\pm0.36$ & $99.87\pm0.14$ & $96.82\pm0.13$ & $98.66\pm1.37$ & $58.55\pm2.71$ & $73.45\pm1.73$ \\
\hline
A.X+Gaz+CRF & SPY & $93.10\pm0.24$ & $99.87\pm0.05$ & $96.37\pm0.12$ & $98.45\pm0.58$ & $52.34\pm1.77$ & $68.33\pm1.44$ \\
 & ReCoPII-Ko & $94.15\pm0.32$ & $99.81\pm0.13$ & $96.90\pm0.12$ & $98.09\pm1.30$ & $60.06\pm2.37$ & $74.47\pm1.52$ \\
\hline
KLUE-R-base$^\dagger$ & SPY & $92.82\pm0.31$ & $99.76\pm0.13$ & $96.16\pm0.11$ & $97.08\pm1.35$ & $50.25\pm2.38$ & $66.18\pm1.74$ \\
 & ReCoPII-Ko & $93.35\pm0.11$ & $99.73\pm0.11$ & $96.43\pm0.03$ & $96.90\pm1.25$ & $54.21\pm0.89$ & $69.52\pm0.47$ \\
\hline
XLM-R-large$^\dagger$ & SPY & $92.50\pm0.06$ & $99.91\pm0.02$ & $96.06\pm0.04$ & $98.75\pm0.30$ & $47.84\pm0.41$ & $64.45\pm0.43$ \\
 & ReCoPII-Ko & $93.52\pm0.26$ & $99.72\pm0.07$ & $96.52\pm0.11$ & $96.90\pm0.69$ & $55.51\pm1.96$ & $70.56\pm1.41$ \\
\hline
\end{tabular*}
\captionof{table}{MASK- and KEEP-class precision, recall, and F1 on the RQ2-B common test (\%; mean $\pm$ sample SD over seeds 42--44). $^\dagger$ marks the two windowed models that failed the preregistered own-version leakage check; their scores are descriptive.}
\label{tab:mask-keep-mask-class}
\label{tab:mask-keep-keep-class}
\end{minipage}

%% file: appendices/synthetic_data_analysis.tex
\section{Model Diagnostics}
\label{sec:synthetic-data-analysis}
We first compare detector components and alternative architectures, then examine detection errors and contextual MASK/KEEP decisions.

\subsection{Detector Component and Architecture Analysis}
\label{sec:rq1-components}
Table~\ref{tab:rq1-components} separates Gazetteer/CRF component changes, the contribution of the MLP at inference, and alternative A.X architectures.

\begin{table*}[!t]
\centering
\small
\setlength{\tabcolsep}{5pt}
\begin{tabular}{lccccc}
\hline
Model / inference condition & Precision & Recall & Micro-F1 & Macro-F1 & Params. (M) \\
\hline
\multicolumn{6}{l}{\textit{Gazetteer and CRF components}} \\
A.X & $74.42\pm1.98$ & $92.98\pm1.44$ & $82.65\pm0.68$ & $85.28\pm1.15$ & $148.76$ \\
A.X+Gazetteer & $74.04\pm1.00$ & $92.80\pm1.05$ & $82.36\pm0.21$ & $85.52\pm0.75$ & $148.76$ \\
A.X+CRF & $78.28\pm2.98$ & $92.44\pm1.24$ & $84.74\pm1.20$ & $88.15\pm0.68$ & $148.76$ \\
A.X+Gazetteer+CRF & $79.95\pm6.28$ & $91.54\pm1.18$ & $85.26\pm3.40$ & $87.56\pm3.91$ & $148.77$ \\
\hline
\multicolumn{6}{l}{\textit{MLP variant: same trained checkpoints}} \\
A.X+Gazetteer+CRF+MLP & $80.78\pm1.88$ & $93.39\pm2.23$ & $86.63\pm2.01$ & $89.47\pm4.13$ & $148.93$ \\
\quad MLP disabled at inference & $79.51\pm1.57$ & $91.93\pm1.85$ & $85.27\pm1.66$ & $88.09\pm3.94$ & $148.93$ \\
\hline
\multicolumn{6}{l}{\textit{Alternative architectures}} \\
A.X+GatedGazetteer+CRF & $76.65\pm4.19$ & $92.41\pm1.25$ & $83.74\pm2.40$ & $87.18\pm2.38$ & $148.79$ \\
A.X+Span & $76.43\pm1.58$ & $94.48\pm0.15$ & $84.49\pm0.91$ & $88.09\pm0.69$ & $149.26$ \\
\hline
\end{tabular}
\caption{A.X component and architecture analysis on KDPII. Scores are percentages, reported as mean $\pm$ sample SD over seeds 42--44, with the same split and exact character-span-and-type matching as Table~\ref{tab:rq1-detection}. Macro-F1 averages all 19 types. Parameter counts are total trained parameters, all trainable, including the MLP when disabled only at inference.}
\label{tab:rq1-components}
\end{table*}

\paragraph{Training settings.}
The A.X models use encoder/head learning rates of $2\times10^{-5}/10^{-4}$, batch size 32, a 256-token limit, and at most 12 epochs with early stopping after three validation checks without improvement. KLUE-BERT-base and its CRF variant use the same settings. Checkpoints are selected by validation micro-F1, using MLP outputs for the MLP variant. All encoders are fully fine-tuned, and Gazetteers use training annotations only.

\paragraph{Gazetteer and CRF.}
Under matched training settings, adding a CRF raises mean micro-F1 from 82.65 to 84.74; adding the Gazetteer to the CRF model yields 85.26. The Gazetteer alone yields 82.36, below the encoder-only mean, so the observed gain depends on the configuration.

\paragraph{MLP inference ablation.}
We disable the MLP at inference while retaining the same jointly trained checkpoints and CRF predictions. Mean micro-F1 decreases from $86.63\pm2.01$ to $85.27\pm1.66$, a difference of 1.36 points. This isolates the use of the MLP during decoding. Comparing the separately trained Gazetteer+CRF and MLP variants instead gives a 1.37-point difference, which combines the MLP with the change to partial-token BIO training targets.

\paragraph{Alternative architectures.}
The gated variant, A.X+\allowbreak{}GatedGazetteer+\allowbreak{}CRF, uses a context-dependent sigmoid gate on Gazetteer features; A.X+Span replaces BIO tagging with a biaffine span classifier. Their mean micro-F1 scores of 83.74 and 84.49 are below 85.26 for A.X+Gazetteer+CRF under these settings.

\subsection{Complete Detection and Transfer Results}
\label{sec:rq1-complete}

Table~\ref{tab:rq1-detection} gives the Detection F1 of six KDPII-trained encoders and three prompted LLMs on both test sets. Both test sets use exact character-span-and-type micro Detection F1: KDPII evaluates 19 PII labels, and on the synthetic test MASK mentions are positives, so detecting a KEEP mention counts as a false positive. The encoders are evaluated without synthetic training, threshold tuning, or relation/privacy inference. K-EXAONE, GPT-4o-mini, and Solar Pro 3 perform direct zero-shot and three-shot extraction from raw text, using the same 19-type prompt on both test sets and three fixed KDPII training documents as demonstrations (Appendix~\ref{app:prompt-rq1}); each condition runs once at temperature zero, without synthetic examples, detector candidates, or gold annotations.

\input{analysis/stage_a_only_seed_transfer/main_table}

The comparison with XLM-R-large in Section~\ref{sec:rq1} uses five seeds for each model under our common split and 19-label protocol, and its 95\% interval comes from a bootstrap over test dialogues and seeds. On the synthetic test, the margin of A.X+Gazetteer+CRF+MLP is 0.38 points over KLUE-BERT-base+CRF and 0.09 over A.X+Gazetteer+CRF (Table~\ref{tab:rq1-transfer-all}). These descriptive margins, without a significance claim, rank zero-shot transfer from KDPII training rather than performance after synthetic training. Under the fixed extraction prompt and output budget, GPT-4o-mini has the highest synthetic Detection F1 among the prompted LLMs in both settings: 60.73 for zero-shot and 51.18 for three-shot. All three LLMs score lower with KDPII demonstrations on the synthetic test, whereas the demonstrations improve their KDPII scores. Thus, the benefit of these fixed examples does not carry over to this synthetic cohort.

Table~\ref{tab:rq1-detection-details} preserves precision, recall, macro-F1, and parameter counts for the main encoders, alongside single-run zero-/three-shot prompted LLM results on KDPII. Prompted systems are not fine-tuned on KDPII; Table~\ref{tab:rq1-detection} reports their direct synthetic detection results with the same extraction prompt and KDPII demonstrations. GPT-4o-mini has the highest three-shot micro-F1 (62.93), while K-EXAONE has the highest three-shot recall (64.69).

\begin{table*}[t]
\centering
\small
\setlength{\tabcolsep}{4pt}
\begin{tabular}{llccccc}
\hline
Model & Setting & Precision & Recall & Micro-F1 & Macro-F1 & Params. (M) \\
\hline
\multicolumn{7}{l}{\textit{Supervised fine-tuning}} \\
XLM-R-large & Fine-tuned & $78.79\pm2.20$ & $94.37\pm0.48$ & $85.86\pm1.11$ & $89.18\pm0.32$ & $558.88$ \\
KLUE-R-base & Fine-tuned & $73.17\pm2.20$ & $93.88\pm0.64$ & $82.22\pm1.20$ & $85.33\pm1.62$ & $110.06$ \\
KLUE-BERT-base & Fine-tuned & $73.41\pm2.32$ & $94.52\pm0.44$ & $82.62\pm1.34$ & $86.33\pm0.20$ & $110.06$ \\
KLUE-BERT-base+CRF & Fine-tuned & $76.99\pm0.56$ & $\mathbf{94.55\pm0.41}$ & $84.87\pm0.33$ & $88.99\pm0.22$ & $110.06$ \\
A.X & Fine-tuned & $74.42\pm1.98$ & $92.98\pm1.44$ & $82.65\pm0.68$ & $85.28\pm1.15$ & $148.76$ \\
A.X+Gazetteer+CRF+MLP & Fine-tuned & $\mathbf{80.78\pm1.88}$ & $93.39\pm2.23$ & $\mathbf{86.63\pm2.01}$ & $\mathbf{89.47\pm4.13}$ & $148.93$ \\
\hline
\multicolumn{7}{l}{\textit{Prompted LLMs}} \\
K-EXAONE & Zero-shot & $48.20$ & $43.11$ & $45.51$ & $52.78$ & -- \\
GPT-4o-mini & Zero-shot & $\mathbf{62.38}$ & $\mathbf{49.93}$ & $\mathbf{55.46}$ & $\mathbf{58.87}$ & -- \\
Solar Pro 3 & Zero-shot & $36.88$ & $47.88$ & $41.66$ & $48.73$ & -- \\
\hline
K-EXAONE & Three-shot & $48.57$ & $\mathbf{64.69}$ & $55.48$ & $61.80$ & -- \\
GPT-4o-mini & Three-shot & $\mathbf{62.56}$ & $63.30$ & $\mathbf{62.93}$ & $\mathbf{67.83}$ & -- \\
Solar Pro 3 & Three-shot & $45.12$ & $63.00$ & $52.58$ & $60.17$ & -- \\
\hline
\end{tabular}
\caption{RQ1 exact character-span-and-type results on KDPII (\%). Supervised values are mean $\pm$ sample standard deviation over seeds 42--44; prompted values are single runs. Macro-F1 averages the 19 type-level F1 scores. Parameter counts are total parameters, all trainable for supervised models; prompted-model counts are unavailable. Bold marks the best result within each training or prompting setting.}
\label{tab:rq1-detection-details}
\end{table*}

\input{analysis/stage_a_only_seed_transfer/all_models_table}

Table~\ref{tab:rq1-transfer-all} reports Detection F1 for all nine KDPII-trained architectures on both test sets. The three additional A.X variants are separated from the representative encoder comparison. The character boundary MLP has the highest mean on KDPII (86.63\%) and the synthetic test (69.40\%). The synthetic runs use original detector training seeds, not contextual-head training seeds.

\subsection{Detection and Masking Diagnostics}
\label{sec:model-error-diagnostics}
\begin{figure*}[!t]
\centering
\includegraphics[width=0.66\textwidth,keepaspectratio]{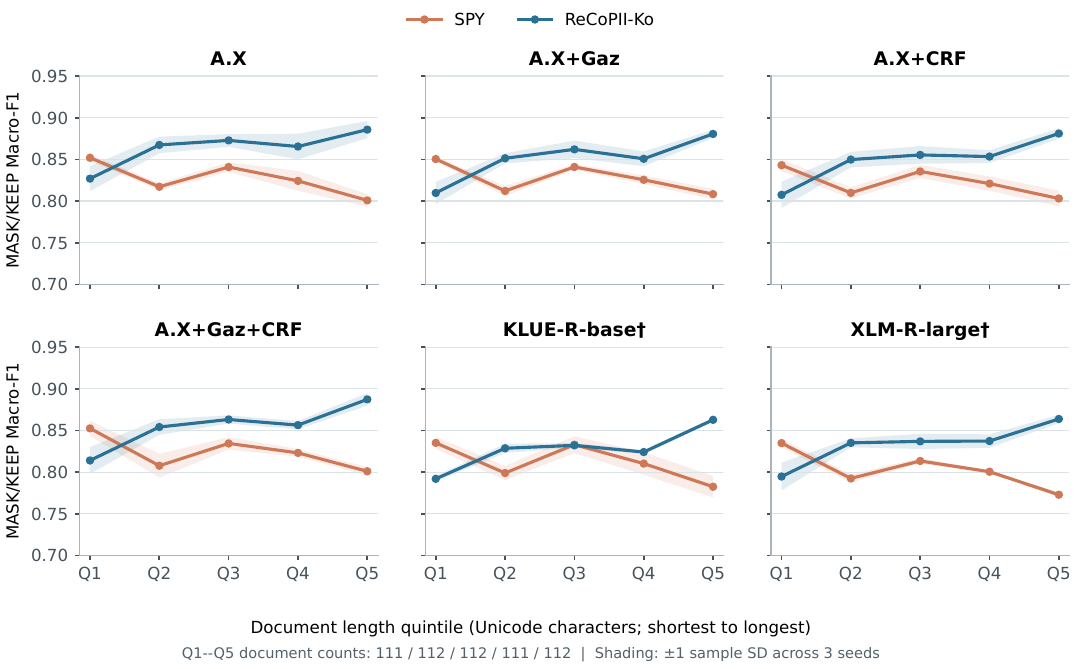}
\caption{RQ2-B MASK/KEEP Macro-F1 by test-document length for six models trained on SPY or ReCoPII-Ko (mean over seeds 42--44; shading shows one sample SD; vertical axis spans 0.70--0.95). Length counts Unicode characters in each stored document; the 558 versioned documents from 279 source IDs are divided into five near-equal groups, with cut points at 1,866, 1,989.6, 2,147.2, and 2,320.8 characters. Scores use supplied candidate spans and pipeline-generated labels of the earlier ReCoPII-Ko snapshot on the same common test, rather than human gold. Each bin pools ReCoPII-Ko and SPY document versions, so differences across bins are descriptive and may reflect changes in version composition. $^\dagger$ marks models that failed the preregistered own-version leakage check.}
\label{fig:rq2b-length-performance}
\end{figure*}

Figure~\ref{fig:rq2b-length-performance} shows RQ2-B MASK/KEEP performance by document length; its caption specifies the evaluation unit and data version.

\subsection{Integrated Masking Results}
\label{sec:rq4-appendix}
\paragraph{Budget rule.}
For the variant comparison of Section~\ref{sec:rq4}, each epoch is scored by the lowest validation over-masking among grid settings whose total validation leakage is at most 1.5\%; an epoch without such a setting is not a checkpoint candidate. Training runs at least five and at most ten epochs, with patience three counted from the first feasible epoch. On test, we report budgets of 1\%, 1.5\%, and 2\% at thresholds chosen on validation, with 1.5\% as the primary budget. The grids are relation thresholds $\{0.3,0.5,0.7\}$ by privacy thresholds $\{0.3,0.5,0.7,0.9\}$ plus masking every candidate for the relation-conditioned heads, and MASK thresholds $\{0,0.1,\ldots,0.8,\allowbreak 0.85,0.9,0.95,0.99\}$ for the direct head. Because Joint Fine-tuning also updates the detector, its candidates come from a detector adapted to synthetic data. XLM-R-large does not meet the budget in five of its six runs: masking every candidate of the jointly fine-tuned detector already leaked more than 1.5\% on validation in every epoch (1.59--3.29\% with relation-conditioned heads and 3.78--8.37\% with the direct head), so no threshold satisfied the rule.

\paragraph{Candidate-assisted LLM baseline.}
The LLMs receive the A.X+Gazetteer+CRF detection-only candidates, which we verified to be identical in all 404 documents. Each LLM predicts relations among all 19 labels, without domain restriction, and edge-level privacy labels. Each condition runs once at temperature zero; three-shot prompts use three fixed synthetic training documents. The six LLM runs leak 13.17--24.28\%.

\paragraph{Registered runs.}
Table~\ref{tab:rq4-main} repeats the full-test columns of Table~\ref{tab:rq3-main} and adds the leakage of each condition's detector; it marks XLM-R-large as not meeting the budget, whereas Table~\ref{tab:rq3-main} also shows its single feasible run. Table~\ref{tab:rq4-all} gives the variant comparison of Section~\ref{sec:rq4} for five detectors and three leakage budgets, and Table~\ref{tab:rq4-registered} gives the registered runs for all eight detectors, from which Detection only, Frozen Detector, and the fixed 0.5 threshold diagnostics are taken. At each epoch, the registered runs searched relation thresholds $\{0.3,0.5,0.7\}$ and privacy thresholds $\{0.3,0.5,0.7,0.9\}$ on validation data, together with masking every candidate, and admitted only settings that added no validation leakage relative to the current detector; the admissible setting and epoch with the lowest validation over-masking were selected. Because this rule also chose checkpoints and stopped training, it kept some runs at masking every candidate, including all registered direct-head runs of A.X+Gazetteer+CRF; we therefore compare the two variants under the budget rule of Section~\ref{sec:rq4}, fixed before any of its results were observed but after the registered runs, so that comparison is post hoc and its variants differ in parameter count. Under the registered rule, Frozen Detector fell back to masking every candidate in all three seeds of A.X+Gazetteer+CRF+MLP and A.X+Gazetteer+CRF, and XLM-R-large lowered over-masking from 98.86\% to 87.70\% while its leakage rose from 14.46\% to 14.57\%. Under the budget rule, the single feasible XLM-R-large run (relation-conditioned heads, seed 43) is reported as one run and not summarized as a mean. Every run uses the same settings: AdamW with learning rates of $2\times10^{-5}$ for the encoder, $10^{-4}$ for the other detector parameters, and $5\times10^{-4}$ for the relation and privacy heads; weight decay 0.01; 10\% warmup; gradient accumulation over eight documents; gradient clipping at 1.0; bfloat16; and at most ten epochs with patience three. For Frozen Detector, training compares a hash of the encoder's first weight tensor before and after training to confirm that it is unchanged.

\input{appendices/sentence_origin_analysis}

\input{analysis/integrated_masking/main_table}

\input{analysis/integrated_masking/all_models_table}

\input{analysis/integrated_masking/registered_table}

\subsection{Human-Verified Test Results}
\label{sec:human-gold-results}
Table~\ref{tab:human-gold-main} gives the human-verified scores of the systems in Table~\ref{tab:rq3-main}, with Detection only averaged over detector seeds 42--44, beside the scores of the same records against their generated labels, and Table~\ref{tab:human-gold-all} covers every system scored on the human-verified test (Appendix~\ref{sec:human-gold-test}) with document-bootstrap intervals. All systems use the predictions that already existed for the official test documents, with the thresholds each had selected on validation. The systems, the two metrics, and the secondary views below were fixed before any system was scored, except where a view is marked as not pre-specified. The intervals show how little 238 preservable mentions resolve: for the eight jointly fine-tuned systems of the four detectors, the over-masking intervals have lower ends of 0.0--1.0\% and upper ends of 6.8--10.4\%. Scored against the generated labels of the same records, Joint Fine-tuning over-masks 0.41--2.89\% and leaks 0.78--3.63\%. The candidate-assisted LLMs over-mask 30.25--97.06\% and leak 19.55--36.27\%. The direct-extraction LLMs of RQ1, which mask every span they extract from raw text, over-mask 54.62--60.50\% and leak 42.99--63.58\%.

Table~\ref{tab:human-gold-sensitivity} asks whether the Joint Fine-tuning result depends on which records are counted and how. Three pre-specified views give lower over-masking: on the 34 pairs held out under both splits, the eight systems of the four detectors over-mask 0.00--0.42\% and leak 0.00--1.37\%; on records where all four annotators agree, the direct head over-masks 0.00--1.95\% and relation-conditioned heads 2.38--3.03\%; and when an entity is counted once, over-masking is 1.04--2.78\% and leakage 2.85--4.69\%. The fourth pre-specified view, an estimate for the 100 documents that weights each entity by the inverse of its sampling probability, gives 8.09--10.58\% over-masking with very wide intervals, because it is dominated by the entities described next, which carry large sampling weights.

After the primary results, we added the view restricted to records whose reference label equals the stored label (not pre-specified). In the ReCoPII-Ko documents, 6 of the 238 reference-KEEP mentions belong to two entities that are stored as MASK, and 4 of the 670 reference-MASK mentions belong to four entities that are stored as KEEP. A system trained on stored labels is expected to follow the stored label on these records. On the records where the two labels agree, the direct head over-masks 0.00--1.29\% and relation-conditioned heads 1.72--2.59\%. The caption of Table~\ref{tab:human-gold-sensitivity} also lists the paired over-masking difference between the two variants; its interval excludes zero only for KLUE-R-base.

\input{analysis/human_gold_test/main_table}

\input{analysis/human_gold_test/all_systems_table}

\input{analysis/human_gold_test/sensitivity_table}

\subsection{Insertion-Free Probe}
\label{sec:probe}
The insertion-free probe applies the systems of Section~\ref{sec:rq4} to documents without inserted sentences, which the official test lacks (Appendix~\ref{sec:sentence-origin}).

\paragraph{Documents.}
The probe has 402 documents from the six domains, with 2,244 KEEP and 3,183 MASK mentions (1,925 and 1,929 entities). KEEP mentions are 41.35\% of its mentions, against 13.68\% in the official test. A separate pipeline produced them: it plans an entity-relation graph with edge-level privacy labels under the same ontology and writes the whole document from that plan, with no source document and no inserted sentence. GPT-5-mini writes the content of each section, and GPT-4o-mini plans the context of each relation and rewrites sections that fail checks. Code builds and validates the annotation, and a document is accepted only if a GPT-5-mini reviewer reproduces the planned privacy label of every relation from the text alone. Of 600 candidates, 100 per domain, 402 were accepted. Entity values come from the persona pool of Appendix~\ref{app:construction-instantiation}, and 211 of the KEEP and 828 of the MASK mentions share value and type with a mention of the official training or validation split. The pipeline rejects candidates whose text overlaps ReCoPII-Ko: no accepted sentence equals a ReCoPII-Ko sentence, and no two documents share a fifth of their sentences.

\paragraph{Labels.}
Labels follow Equation~\ref{eq:entity-masking} applied to the stored relations, and every document passes the graph validation used for the official data. In 37 documents, 39 relations that repeated an earlier relation with the same label were dropped, which cannot change a label. The labels are generated and not human-verified, so the rates below measure agreement with generated labels.

\paragraph{Systems and checks.}
Nothing is retrained or tuned on the probe: every system uses the checkpoints, the thresholds selected on official validation data, and the prompts of the official test, with the metrics of Section~\ref{sec:rq4}. Each of the 54 inference runs first reproduced its stored predictions on eight official test documents, and the probe scorer matches the official metric code on all 88 system outputs.

\paragraph{Results.}
Table~\ref{tab:probe-main} repeats the rows of Table~\ref{tab:rq3-main}: for the four detectors that meet the budget, Joint Fine-tuning over-masks 11.33--34.21\% and leaks 2.03--4.95\%, where the same systems give 1.10--2.80\% and 0.95--1.44\% on the official test; Detection only over-masks 88.59--95.99\% and leaks 8.07--13.82\% for these detectors and XLM-R-large. The decision releases 45.22--82.51\% of the KEEP mentions that its fine-tuned detector masks and adds 0.77--2.21 points of leakage to that detector's own 0.79--2.94\% (Table~\ref{tab:probe-all}). The systems thus still release most of the preservable mentions that masking every candidate would hide, at leakage of a few percent, so the decision does not rest on inserted sentences alone. Both rates are nevertheless well above the official test, so part of the accuracy on ReCoPII-Ko does not carry over.

In pre-specified breakdowns, over-masking is 9.70--31.07\% on the six types with at least 500 KEEP mentions in the official training split and 22.49--64.80\% on the other types (group sizes in the caption of Table~\ref{tab:probe-all}); Table~\ref{tab:probe-types} lists every type. With all thresholds fixed at 0.5, the systems over-mask 15.39--33.32\% and leak 1.92--4.47\%, so the gap does not come from the selected thresholds. On mentions whose value and type do not occur in the official training or validation split, over-masking is 10.84--32.43\%, against 16.11--51.34\% on those that do. Relation-conditioned heads over-mask more than the direct head for three of the four detectors, by 5.42--18.89 points; for A.X+Span the paired interval includes zero (caption of Table~\ref{tab:probe-all}).

Within the probe, the six candidate-assisted LLM runs over-mask 16.76--85.29\% at 12.19--25.89\% leakage, and the direct-extraction LLMs over-mask 36.41--45.45\% at 45.40--49.70\% leakage (Tables~\ref{tab:probe-main} and~\ref{tab:probe-reference}). Every Joint Fine-tuning system leaks less, at most 4.95\%, than every LLM run, at least 12.19\%; on over-masking alone, three-shot GPT-4o-mini (16.76\%) is below all Joint Fine-tuning systems but one. Three unusable answers leave a document fully masked (at most 13 of the 402 documents in a run). We do not compare an LLM's score between the official test and the probe, because the probe pipeline keeps only documents whose relation privacy an LLM reviewer reproduces from the text, and GPT-4o-mini, one of the baselines, took part in generating them (planning and rewriting).

\paragraph{Limits.}
The probe was run after the main results, with its systems, decision rules, metrics, and type grouping fixed before any model was run on it (Table~\ref{tab:probe-reference} marks the rows added later). Construction by insertion, the share of KEEP mentions, and the absence of human verification all differ between the official test and the probe, so the gap cannot be assigned to one of them. The probe is also LLM-generated under the same ontology and persona pool, and by the label definition its preservable mentions are organization-managed values and reference persons; it removes the contrast between source and inserted sentences, not wording regularities that generated preservable mentions may share.

\begingroup
\makeatletter
\renewenvironment{table}[1][]{\par\medskip\noindent\begin{minipage}{\columnwidth}\def\@captype{table}}{\end{minipage}\par\medskip}
\makeatother
\input{analysis/from_scratch_probe/type_table}
\endgroup

\input{analysis/from_scratch_probe/main_table}

\input{analysis/from_scratch_probe/all_systems_table}

\input{analysis/from_scratch_probe/reference_table}

%% file: analysis/stage_a_only_seed_transfer/main_table.tex
\begin{table}[t]
\centering
\small
\setlength{\tabcolsep}{2pt}
\begin{tabular}{lcc}
\hline
Model & KDPII F1 $\uparrow$ & Synthetic F1 $\uparrow$ \\
\hline
\multicolumn{3}{l}{\textit{Supervised encoders (KDPII training)}} \\
XLM-R-large & $85.86\pm1.11$ & $60.96\pm2.88$ \\
KLUE-R-base & $82.22\pm1.20$ & $65.91\pm0.47$ \\
KLUE-BERT-base & $82.62\pm1.34$ & $65.08\pm2.32$ \\
KLUE-BERT-base+CRF & $84.87\pm0.33$ & $69.01\pm0.41$ \\
A.X & $82.65\pm0.68$ & $65.11\pm0.52$ \\
A.X+Gazetteer+CRF+MLP & $\mathbf{86.63\pm2.01}$ & $\mathbf{69.40\pm0.73}$ \\
\hline
\multicolumn{3}{l}{\textit{Zero-shot prompted LLMs}} \\
K-EXAONE & $45.51$ & $54.67$ \\
GPT-4o-mini & $\mathbf{55.46}$ & $\mathbf{60.73}$ \\
Solar Pro 3 & $41.66$ & $55.42$ \\
\hline
\multicolumn{3}{l}{\textit{Three-shot prompted LLMs}} \\
K-EXAONE & $55.48$ & $46.31$ \\
GPT-4o-mini & $\mathbf{62.93}$ & $\mathbf{51.18}$ \\
Solar Pro 3 & $52.58$ & $51.05$ \\
\hline
\end{tabular}
\caption{Exact character-span-and-type Detection F1 on KDPII and the synthetic test (\%). Encoders use the same KDPII-trained checkpoints on both datasets (mean $\pm$ sample SD, seeds 42--44). Prompted LLMs directly extract spans from text, using the same prompt and, for three-shot, the same KDPII training examples on both datasets (single runs). Synthetic MASK mentions are positives; KEEP detections are false positives. Bold marks the best encoder mean across nine architectures or the best LLM score within each prompting setting. Full encoder results appear in Appendix Table~\ref{tab:rq1-transfer-all}. MLP denotes character boundary refinement.}
\label{tab:rq1-detection}
\end{table}

%% file: analysis/stage_a_only_seed_transfer/all_models_table.tex
\begin{table*}[t]
\centering
\small
\setlength{\tabcolsep}{4pt}
\begin{tabular}{lcc}
\hline
Model & KDPII Detection F1 $\uparrow$ & Synthetic Detection F1 $\uparrow$ \\
\hline
XLM-R-large & $85.86\pm1.11$ & $60.96\pm2.88$ \\
KLUE-R-base & $82.22\pm1.20$ & $65.91\pm0.47$ \\
KLUE-BERT-base & $82.62\pm1.34$ & $65.08\pm2.32$ \\
KLUE-BERT-base+CRF & $84.87\pm0.33$ & $69.01\pm0.41$ \\
A.X & $82.65\pm0.68$ & $65.11\pm0.52$ \\
A.X+Gazetteer+CRF+MLP & $\mathbf{86.63\pm2.01}$ & $\mathbf{69.40\pm0.73}$ \\
A.X+Gazetteer+CRF & $85.26\pm3.40$ & $69.31\pm1.51$ \\
A.X+GatedGazetteer+CRF & $83.74\pm2.40$ & $68.00\pm1.01$ \\
A.X+Span & $84.49\pm0.91$ & $67.69\pm0.56$ \\
\hline
\end{tabular}
\caption{All nine KDPII-trained detectors on both test sets. Settings and metrics match Table~\ref{tab:rq1-detection}; all values are exact character-span-and-type Detection F1, reported as mean $\pm$ sample SD over detector seeds 42--44 (\%). Bold identifies the best mean across these nine models. Gazetteer+CRF, GatedGazetteer+CRF, and Span are included to expose the complete comparison.}
\label{tab:rq1-transfer-all}
\end{table*}

%% file: appendices/sentence_origin_analysis.tex
\subsection{Sentence-Origin Analysis}
\label{sec:sentence-origin}
Preservable mentions sit almost only in inserted sentences (Table~\ref{tab:sentence-origin}, panel A), so low over-masking could reflect recognition of inserted sentences instead of an entity's role. A head that followed such a rule would also release the maskable mentions that inserted sentences contain, 522 of the 11,045 MASK mentions. This analysis therefore asks whether sentence origin changes the release of MASK mentions. It covers the 25 variant-comparison runs with test outputs: three seeds and both variants for A.X+Span, A.X+Gazetteer+CRF+MLP, A.X+Gazetteer+CRF, and KLUE-R-base, and the relation-conditioned XLM-R-large run with seed 43, all at the thresholds chosen on validation for the 1.5\% budget. A sentence counts as a source sentence when its exact text occurs in Basic. Release is measured among mentions fully covered by the run's own detector, so that it reflects the head's decision.

\par\medskip\noindent
\begin{minipage}{\columnwidth}
\centering
\scriptsize
\setlength{\tabcolsep}{4pt}
\begin{tabular}{@{}lrr@{}}
\hline
\multicolumn{3}{@{}l@{}}{\emph{A. Test mentions (descriptive)}} \\
 & Inserted & Source \\
\hline
KEEP mentions & 1,734 & 16 \\
MASK mentions & 522 & 10,523 \\
\hline
\multicolumn{3}{@{}l@{}}{\rule{0pt}{2.6ex}\emph{B. Release rate (\%) by head (descriptive)}} \\
 & Rel.-cond.\ (13) & Direct (12) \\
\hline
MASK, source & 0.04--0.36 (0.19) & 0.04--0.63 (0.15) \\
MASK, inserted & 2.24--9.59 (4.37) & 1.77--11.61 (3.55) \\
KEEP, inserted & 94.92--98.95 (98.27) & 99.30--99.48 (99.42) \\
KEEP, source & 0.00--18.75 (0.00) & 0.00--56.25 (21.88) \\
\hline
\multicolumn{3}{@{}l@{}}{\rule{0pt}{2.6ex}\emph{C. Released MASK mentions (not pre-specified)}} \\
 & Inserted & Source \\
\hline
Other types & 0 / 500 & 253 / 231,970 \\
 & (0.00\%) & (0.11\%) \\
Organization types & 639 / 12,228 & 269 / 29,439 \\
 & (5.23\%) & (0.91\%) \\
\hline
\end{tabular}

\vspace{4pt}
\normalsize\captionof{table}{Sentence-origin analysis on the 404-document test. A: gold mentions by sentence origin. B: share of detected mentions that the head releases, as the range over runs with the median in parentheses; the number of runs follows each head, and KEEP in source sentences has 16 mentions. C: released over detected gold MASK mentions, pooled over mentions and the 25 runs; organization types are \texttt{WORKPLACE}, \texttt{DEPARTMENT}, and \texttt{SCHOOL}.}
\label{tab:sentence-origin}
\end{minipage}
\par\medskip

\paragraph{Release rates (descriptive).}
Panel B gives the share of detected mentions that each head releases. Both heads release MASK mentions more often in inserted sentences (1.77--11.61\% across the 25 runs) than in source sentences (0.04--0.63\%). The 16 KEEP mentions in source sentences are too few to estimate preservation in unedited text.

\paragraph{Test of an origin effect (pre-specified).}
A higher release rate in inserted sentences could also arise if those sentences held entity types or documents that are harder to decide, so we fixed a test and its decision rule before running it. The test estimates the Mantel--Haenszel odds ratio of release, inserted versus source, within strata that hold these factors fixed, with 95\% intervals from the Robins--Breslow--Greenland variance. The rule reports an origin effect beyond document and type difficulty if the interval of the odds ratio stratified by document and entity type lies above 1 in at least two thirds of the runs, no detectable effect if it includes 1 in at least two thirds, and an inconclusive result otherwise. Stratified by entity type, the interval lies above 1 in 25 of 25 runs (odds ratios 3.6--91.8, median 8.3), and stratified by document in 25 of 25 runs (7.2--119.5, median 32.7). Stratified by document and entity type, it lies above 1 in 14 of the 15 runs where the odds ratio can be estimated (5.9--32.0, median 11.5); the other 10 runs have empty stratum sums. Under the rule, sentence origin affects release beyond document and type difficulty.

\paragraph{Organization types (not pre-specified).}
Panel C separates this effect by entity type. In inserted sentences, the heads release 5.23\% of organization mentions, against 0.91\% in source sentences, and none of the 500 mentions of other types. The same gap appears within entities: for the 6,112 entity--run pairs with mentions of both origins, 4.64\% of inserted mentions are released against 0.85\% of source mentions, although the label is assigned per entity. The heads thus do not release whatever an inserted sentence contains. The released organizations are entities that Equation~\ref{eq:entity-masking} labels MASK through a \texttt{PII} edge elsewhere in the document, while the inserted sentence presents them in an organizational role; the pattern is consistent with a conflict between these two roles. At the median release rates, inserted-sentence MASK mentions, 4.7\% of all MASK mentions, contribute about 0.2 points of leakage.

\paragraph{Limits.}
Organization types make up 96\% of the detected MASK mentions in inserted sentences (12,228 of 12,728), which leaves about 20 mentions of other types per run, so their zero releases are weak evidence on their own. For organization types, the data cannot separate a role conflict from a stylistic cue specific to these types. Whether the heads preserve KEEP mentions in unedited text cannot be tested with 16 such mentions. Appendix~\ref{sec:probe} therefore applies the same systems to documents written without inserted sentences.

%% file: analysis/integrated_masking/main_table.tex
\begin{table*}[t]
\centering
\small
\setlength{\tabcolsep}{3pt}
\resizebox{\textwidth}{!}{%
\begin{tabular}{llccc}
\hline
Detector & Condition & Over-mask $\downarrow$ & Leakage $\downarrow$ & Det.\ leakage \\
\hline
A.X+Gazetteer+CRF+MLP & Detection only & 95.66 & 13.02 & 13.02 \\
 & Frozen Detector & $95.66\pm0.00$ & $13.02\pm0.00$ & $13.02\pm0.00$ \\
 & Joint FT, relation-conditioned & $2.36\pm0.43$ & $1.20\pm0.34$ & $0.74\pm0.12$ \\
 & Joint FT, direct & $1.10\pm0.03$ & $1.13\pm0.22$ & $0.69\pm0.11$ \\
\hline
A.X+Gazetteer+CRF & Detection only & 98.11 & 12.70 & 12.70 \\
 & Frozen Detector & $98.11\pm0.00$ & $12.70\pm0.00$ & $12.70\pm0.00$ \\
 & Joint FT, relation-conditioned & $2.57\pm0.57$ & $1.32\pm0.24$ & $0.90\pm0.35$ \\
 & Joint FT, direct & $1.41\pm0.09$ & $0.97\pm0.17$ & $0.78\pm0.10$ \\
\hline
XLM-R-large & Detection only & 98.86 & 14.46 & 14.46 \\
 & Frozen Detector & $87.70\pm1.71$ & $14.57\pm0.08$ & $14.46\pm0.00$ \\
 & Joint FT, relation-conditioned & \multicolumn{3}{c}{budget not met$^\ddagger$} \\
 & Joint FT, direct & \multicolumn{3}{c}{budget not met$^\ddagger$} \\
\hline
GPT-4o-mini & LLM, 0-shot & 62.40 & 16.88 & 12.70 \\
 & LLM, 3-shot & 28.34 & 24.28 & 12.70 \\
K-EXAONE & LLM, 0-shot & 95.37 & 13.17 & 12.70 \\
 & LLM, 3-shot & 61.94 & 15.30 & 12.70 \\
Solar Pro 3 & LLM, 0-shot & 80.11 & 15.01 & 12.70 \\
 & LLM, 3-shot & 42.97 & 15.59 & 12.70 \\
\hline
\end{tabular}%
}
\caption{RQ3 masking results on the official 404-document test (\%; generated labels). Over-masking is the share of 1,750 KEEP mentions with any masked character; leakage is the share of 11,045 MASK mentions with any unmasked character; Det.\ leakage masks every candidate of the current detector. Values with $\pm$ are mean and sample SD over contextual seeds 42--44 from one KDPII-trained detector (seed 42). Joint Fine-tuning (FT) uses thresholds and checkpoints selected on validation data under a 1.5\% total-leakage budget; Detection only, Frozen Detector, and the single-run LLMs, which receive the A.X+Gazetteer+CRF detection-only candidates, come from the registered runs. $^\ddagger$ 5 of 6 XLM-R-large runs had no epoch within the budget: masking every candidate still leaked 1.59--3.29\% (relation-conditioned) and 3.78--8.37\% (direct) on validation. Appendix Table~\ref{tab:rq4-all} reports five detectors and three budgets.}
\label{tab:rq4-main}
\end{table*}

%% file: analysis/integrated_masking/all_models_table.tex
\begin{table*}[t]
\centering
\scriptsize
\setlength{\tabcolsep}{3pt}
\begin{tabular}{llcccccc}
\hline
 & & \multicolumn{2}{c}{Relation-conditioned} & \multicolumn{2}{c}{Direct} & \multicolumn{2}{c}{Relation $-$ direct} \\
Detector & Budget & Over-mask & Leakage & Over-mask & Leakage & Over-mask & Leakage \\
\hline
A.X+Span & 1\% & $2.42\pm0.40$ & $1.10\pm0.20$ & $1.28\pm0.29$ & $1.36\pm0.30$ & +1.14 [0.51, 1.86] & -0.26 [-0.66, 0.06] \\
 & 1.5\% & $2.25\pm0.23$ & $1.19\pm0.29$ & $1.20\pm0.17$ & $1.44\pm0.31$ & +1.05 [0.50, 1.68] & -0.25 [-0.64, 0.06] \\
 & 2\% & $2.25\pm0.23$ & $1.19\pm0.29$ & $1.07\pm0.29$ & $1.74\pm0.23$ & +1.18 [0.58, 1.91] & -0.55 [-0.97, -0.21] \\
\hline
A.X+Gazetteer+CRF+MLP & 1\% & $66.93\pm56.09$ & $0.86\pm0.13$ & $66.80\pm56.71$ & $0.72\pm0.14$ & +0.13 [-0.15, 0.43] & +0.15 [0.02, 0.27] \\
 & 1.5\% & $2.36\pm0.43$ & $1.20\pm0.34$ & $1.10\pm0.03$ & $1.13\pm0.22$ & +1.26 [0.61, 2.02] & +0.06 [-0.15, 0.27] \\
 & 2\% & $2.36\pm0.43$ & $1.20\pm0.34$ & $1.10\pm0.03$ & $1.13\pm0.22$ & +1.26 [0.61, 2.02] & +0.06 [-0.15, 0.27] \\
\hline
A.X+Gazetteer+CRF & 1\% & \multicolumn{2}{c}{2/3 seeds feasible} & $34.11\pm56.66$ & $0.89\pm0.09$ & --- & --- \\
 & 1.5\% & $2.57\pm0.57$ & $1.32\pm0.24$ & $1.41\pm0.09$ & $0.97\pm0.17$ & +1.16 [0.70, 1.66] & +0.34 [0.17, 0.52] \\
 & 2\% & $2.57\pm0.55$ & $1.47\pm0.30$ & $1.41\pm0.09$ & $0.97\pm0.17$ & +1.16 [0.68, 1.71] & +0.49 [0.28, 0.73] \\
\hline
KLUE-R-base & 1\% & $35.68\pm55.52$ & $0.74\pm0.11$ & \multicolumn{2}{c}{2/3 seeds feasible} & --- & --- \\
 & 1.5\% & $2.80\pm0.40$ & $0.95\pm0.12$ & $1.33\pm0.14$ & $1.00\pm0.51$ & +1.47 [0.84, 2.13] & -0.05 [-0.29, 0.19] \\
 & 2\% & $2.80\pm0.40$ & $0.95\pm0.12$ & $1.33\pm0.14$ & $1.00\pm0.51$ & +1.47 [0.84, 2.13] & -0.05 [-0.29, 0.19] \\
\hline
XLM-R-large & 1\% & \multicolumn{2}{c}{1/3 seeds feasible} & \multicolumn{2}{c}{0/3 seeds feasible} & --- & --- \\
 & 1.5\% & \multicolumn{2}{c}{1/3 seeds feasible} & \multicolumn{2}{c}{0/3 seeds feasible} & --- & --- \\
 & 2\% & \multicolumn{2}{c}{1/3 seeds feasible} & \multicolumn{2}{c}{0/3 seeds feasible} & --- & --- \\
\hline
\end{tabular}
\caption{RQ3 Joint Fine-tuning with relation-conditioned heads and with the direct head on the official 404-document test (\%; generated labels; mean $\pm$ sample SD over contextual seeds 42--44), at thresholds selected on validation data for total-leakage budgets of 1\%, 1.5\% (primary), and 2\%. Difference columns give relation-conditioned minus direct with 95\% intervals from 2,000 bootstrap resamples of source-document groups. A variant is summarized only when all three seeds have a feasible threshold; at 1\% some seeds select masking every candidate, which produces the large standard deviations.}
\label{tab:rq4-all}
\end{table*}

%% file: analysis/integrated_masking/registered_table.tex
\begin{table*}[t]
\centering
\scriptsize
\setlength{\tabcolsep}{3pt}
\begin{tabular}{llcccccc}
\hline
Detector & Condition & Mask-all & Over-mask & Leakage & Over-mask@0.5 & Leakage@0.5 & Det.\ leakage \\
\hline
A.X+Gazetteer+CRF+MLP & Detection only & --- & 95.66 & 13.02 & --- & --- & 13.02 \\
 & Frozen Detector & 3/3 & $95.66\pm0.00$ & $13.02\pm0.00$ & $94.95\pm0.53$ & $13.08\pm0.01$ & $13.02\pm0.00$ \\
 & Joint Fine-tuning & 2/3 & $66.27\pm54.35$ & $1.38\pm0.66$ & $13.70\pm9.88$ & $1.63\pm0.61$ & $1.35\pm0.69$ \\
\hline
A.X+Gazetteer+CRF & Detection only & --- & 98.11 & 12.70 & --- & --- & 12.70 \\
 & Frozen Detector & 3/3 & $98.11\pm0.00$ & $12.70\pm0.00$ & $96.59\pm0.29$ & $13.07\pm0.22$ & $12.70\pm0.00$ \\
 & Joint Fine-tuning & 1/3 & $35.50\pm53.83$ & $1.06\pm0.38$ & $14.91\pm17.69$ & $1.14\pm0.49$ & $1.04\pm0.40$ \\
\hline
A.X+GatedGazetteer+CRF & Detection only & --- & 96.91 & 10.96 & --- & --- & 10.96 \\
 & Frozen Detector & 1/3 & $94.55\pm3.85$ & $11.00\pm0.04$ & $93.18\pm3.65$ & $11.07\pm0.03$ & $10.96\pm0.00$ \\
 & Joint Fine-tuning & 0/3 & $30.23\pm20.84$ & $0.93\pm0.15$ & $24.51\pm10.38$ & $0.98\pm0.20$ & $0.91\pm0.16$ \\
\hline
A.X+Span & Detection only & --- & 99.83 & 9.15 & --- & --- & 9.15 \\
 & Frozen Detector & 1/3 & $99.60\pm0.30$ & $9.16\pm0.01$ & $96.82\pm2.51$ & $9.21\pm0.06$ & $9.15\pm0.00$ \\
 & Joint Fine-tuning & 0/3 & $9.14\pm2.72$ & $0.70\pm0.12$ & $7.56\pm2.55$ & $0.72\pm0.10$ & $0.67\pm0.14$ \\
\hline
XLM-R-large & Detection only & --- & 98.86 & 14.46 & --- & --- & 14.46 \\
 & Frozen Detector & 0/3 & $87.70\pm1.71$ & $14.57\pm0.08$ & $74.74\pm6.04$ & $14.68\pm0.02$ & $14.46\pm0.00$ \\
 & Joint Fine-tuning & 2/3 & $86.34\pm22.47$ & $2.98\pm0.72$ & $30.21\pm29.45$ & $3.11\pm0.66$ & $2.98\pm0.72$ \\
\hline
KLUE-R-base & Detection only & --- & 99.26 & 10.97 & --- & --- & 10.97 \\
 & Frozen Detector & 3/3 & $99.26\pm0.00$ & $10.97\pm0.00$ & $95.43\pm2.98$ & $11.30\pm0.16$ & $10.97\pm0.00$ \\
 & Joint Fine-tuning & 0/3 & $22.82\pm4.03$ & $0.72\pm0.23$ & $20.91\pm7.33$ & $0.72\pm0.23$ & $0.68\pm0.22$ \\
\hline
KLUE-BERT-base & Detection only & --- & 99.54 & 11.75 & --- & --- & 11.75 \\
 & Frozen Detector & 3/3 & $99.54\pm0.00$ & $11.75\pm0.00$ & $85.22\pm8.97$ & $12.33\pm0.49$ & $11.75\pm0.00$ \\
 & Joint Fine-tuning & 0/3 & $10.48\pm3.77$ & $1.16\pm0.24$ & $11.01\pm4.97$ & $1.11\pm0.21$ & $1.07\pm0.20$ \\
\hline
KLUE-BERT-base+CRF & Detection only & --- & 99.26 & 10.76 & --- & --- & 10.76 \\
 & Frozen Detector & 0/3 & $95.05\pm4.46$ & $10.84\pm0.04$ & $90.08\pm9.15$ & $10.86\pm0.04$ & $10.76\pm0.00$ \\
 & Joint Fine-tuning & 1/3 & $37.47\pm52.97$ & $1.49\pm0.81$ & $18.23\pm21.97$ & $1.55\pm0.90$ & $1.45\pm0.84$ \\
\hline
\end{tabular}
\caption{Registered runs for all eight KDPII-trained detectors on the official 404-document test (\%; generated labels; mean $\pm$ sample SD over contextual seeds 42--44), with relation-conditioned heads. Their rule selected thresholds and checkpoints that add no validation leakage relative to the current detector; Mask-all counts seeds that fell back to masking every candidate, and @0.5 columns use fixed relation and privacy thresholds as a diagnostic that does not select models.}
\label{tab:rq4-registered}
\end{table*}

%% file: analysis/human_gold_test/main_table.tex
\begin{table*}[t]
\centering
\small
\setlength{\tabcolsep}{4pt}
\resizebox{\textwidth}{!}{%
\begin{tabular}{llcccc}
\hline
 & & \multicolumn{2}{c}{Human-verified labels} & \multicolumn{2}{c}{Generated labels, same records} \\
Detector & Condition & Over-mask $\downarrow$ & Leakage $\downarrow$ & Over-mask $\downarrow$ & Leakage $\downarrow$ \\
\hline
A.X+\allowbreak{}Gazetteer+\allowbreak{}CRF+\allowbreak{}MLP & Detection only & $98.88\pm0.87$ & $17.56\pm1.50$ & $98.62\pm1.04$ & $17.25\pm1.59$ \\
 & Frozen Detector & $99.16\pm0.00$ & $19.25\pm0.00$ & $98.76\pm0.00$ & $19.04\pm0.00$ \\
 & Joint FT, relation-conditioned & $5.04\pm1.11$ & $1.14\pm0.60$ & $2.89\pm1.09$ & $0.78\pm0.75$ \\
 & Joint FT, direct & $2.52\pm0.00$ & $1.74\pm1.12$ & $0.41\pm0.00$ & $1.26\pm1.09$ \\
\hline
A.X+\allowbreak{}Gazetteer+\allowbreak{}CRF & Detection only & $99.30\pm0.24$ & $18.51\pm2.21$ & $99.31\pm0.24$ & $18.22\pm2.19$ \\
 & Frozen Detector & $99.16\pm0.00$ & $19.55\pm0.00$ & $99.17\pm0.00$ & $19.33\pm0.00$ \\
 & Joint FT, relation-conditioned & $4.48\pm0.64$ & $1.94\pm0.54$ & $2.34\pm0.63$ & $1.45\pm0.52$ \\
 & Joint FT, direct & $3.78\pm0.00$ & $1.99\pm0.74$ & $1.65\pm0.00$ & $1.50\pm0.72$ \\
\hline
XLM-R-large & Detection only & $99.44\pm0.24$ & $18.76\pm1.50$ & $99.45\pm0.24$ & $18.56\pm1.60$ \\
 & Frozen Detector & $91.46\pm1.94$ & $17.31\pm0.00$ & $91.32\pm2.15$ & $17.01\pm0.00$ \\
 & Joint FT, relation-conditioned$^\dagger$ & 9.66 & 1.49 & 7.44 & 1.16 \\
 & Joint FT, direct & \multicolumn{4}{c}{budget not met} \\
\hline
GPT-4o-mini & LLM, 0-shot & 60.92 & 23.88 & 59.50 & 24.71 \\
 & LLM, 3-shot & 30.25 & 36.27 & 28.51 & 36.19 \\
K-EXAONE & LLM, 0-shot & 97.06 & 19.55 & 97.11 & 19.33 \\
 & LLM, 3-shot & 64.71 & 20.90 & 64.46 & 20.64 \\
Solar Pro 3 & LLM, 0-shot & 86.13 & 20.90 & 85.95 & 20.78 \\
 & LLM, 3-shot & 40.76 & 21.34 & 40.08 & 21.08 \\
\hline
\end{tabular}%
}
\caption{RQ3 systems scored on the human-verified test (\%): 192 reference-KEEP entities (238 mentions) and 199 reference-MASK entities (670 mentions) in the ReCoPII-Ko version of 100 test documents. Over-masking and leakage follow the RQ3 definitions restricted to these mentions. The right half scores the same records against their stored (generated) labels. Values with $\pm$ are mean and sample SD over seeds (detector seeds 42--44 for detection only, contextual seeds 42--44 otherwise); thresholds are those each system selected on validation, and nothing was tuned on this set. $^\dagger$ Single run that met the 1.5\% budget. LLMs are single runs over the A.X+Gazetteer+CRF candidates. Appendix Table~\ref{tab:human-gold-all} gives document-bootstrap intervals and all systems.}
\label{tab:human-gold-main}
\end{table*}

%% file: analysis/human_gold_test/all_systems_table.tex
\begin{table*}[t]
\centering
\small
\setlength{\tabcolsep}{4pt}
\resizebox{\textwidth}{!}{%
\begin{tabular}{llccccc}
\hline
Group & System & Over-mask $\downarrow$ & 95\% CI & Leakage $\downarrow$ & 95\% CI & Det.\ leakage \\
\hline
Detection only & A.X+\allowbreak{}Gazetteer+\allowbreak{}CRF+\allowbreak{}MLP & $98.88\pm0.87$ & [97.6, 99.9] & $17.56\pm1.50$ & [9.8, 28.0] & --- \\
 & A.X+\allowbreak{}Gazetteer+\allowbreak{}CRF & $99.30\pm0.24$ & [98.4, 100.0] & $18.51\pm2.21$ & [10.6, 29.0] & --- \\
 & A.X+\allowbreak{}GatedGazetteer+\allowbreak{}CRF & $99.44\pm0.64$ & [98.6, 100.0] & $16.07\pm2.24$ & [8.8, 26.4] & --- \\
 & A.X+\allowbreak{}Span & $99.86\pm0.24$ & [99.6, 100.0] & $15.37\pm2.39$ & [8.2, 25.4] & --- \\
 & A.X & $99.58\pm0.42$ & [99.1, 100.0] & $16.27\pm1.76$ & [8.5, 26.6] & --- \\
 & KLUE-R-base & $100.00\pm0.00$ & [100.0, 100.0] & $14.93\pm1.28$ & [7.1, 25.2] & --- \\
 & KLUE-BERT-base & $99.72\pm0.24$ & [99.3, 100.0] & $15.47\pm0.75$ & [8.1, 25.6] & --- \\
 & KLUE-BERT-base+\allowbreak{}CRF & $99.86\pm0.24$ & [99.6, 100.0] & $14.98\pm1.41$ & [7.7, 25.0] & --- \\
 & XLM-R-large & $99.44\pm0.24$ & [98.7, 100.0] & $18.76\pm1.50$ & [12.5, 25.9] & --- \\
\hline
Frozen Detector & A.X+\allowbreak{}Gazetteer+\allowbreak{}CRF+\allowbreak{}MLP & $99.16\pm0.00$ & [97.9, 100.0] & $19.25\pm0.00$ & [10.8, 29.8] & --- \\
 & A.X+\allowbreak{}Gazetteer+\allowbreak{}CRF & $99.16\pm0.00$ & [97.8, 100.0] & $19.55\pm0.00$ & [10.9, 30.0] & --- \\
 & A.X+\allowbreak{}GatedGazetteer+\allowbreak{}CRF & $97.76\pm2.16$ & [95.7, 99.3] & $13.68\pm0.17$ & [6.9, 23.6] & --- \\
 & A.X+\allowbreak{}Span & $100.00\pm0.00$ & [100.0, 100.0] & $12.99\pm0.00$ & [6.5, 22.6] & --- \\
 & KLUE-R-base & $100.00\pm0.00$ & [100.0, 100.0] & $16.12\pm0.00$ & [8.2, 26.4] & --- \\
 & KLUE-BERT-base & $100.00\pm0.00$ & [100.0, 100.0] & $16.27\pm0.00$ & [8.8, 26.1] & --- \\
 & KLUE-BERT-base+\allowbreak{}CRF & $94.68\pm5.83$ & [92.1, 96.9] & $16.57\pm0.00$ & [8.6, 27.1] & --- \\
 & XLM-R-large & $91.46\pm1.94$ & [88.3, 94.6] & $17.31\pm0.00$ & [11.8, 24.0] & --- \\
\hline
Joint Fine-tuning & A.X+\allowbreak{}Span, relation-conditioned & $4.20\pm0.42$ & [0.4, 9.3] & $1.69\pm0.23$ & [0.4, 3.4] & $1.24\pm0.23$ \\
 & A.X+\allowbreak{}Span, direct & $3.08\pm0.64$ & [0.1, 7.6] & $4.18\pm1.76$ & [0.7, 10.1] & $3.23\pm1.79$ \\
 & A.X+\allowbreak{}Gazetteer+\allowbreak{}CRF+\allowbreak{}MLP, relation-conditioned & $5.04\pm1.11$ & [0.6, 10.4] & $1.14\pm0.60$ & [0.4, 2.0] & $0.45\pm0.30$ \\
 & A.X+\allowbreak{}Gazetteer+\allowbreak{}CRF+\allowbreak{}MLP, direct & $2.52\pm0.00$ & [0.0, 6.8] & $1.74\pm1.12$ & [0.7, 3.0] & $0.45\pm0.39$ \\
 & A.X+\allowbreak{}Gazetteer+\allowbreak{}CRF, relation-conditioned & $4.48\pm0.64$ & [0.6, 9.5] & $1.94\pm0.54$ & [0.9, 3.2] & $0.95\pm0.09$ \\
 & A.X+\allowbreak{}Gazetteer+\allowbreak{}CRF, direct & $3.78\pm0.00$ & [0.0, 9.0] & $1.99\pm0.74$ & [0.7, 3.5] & $1.34\pm0.39$ \\
 & KLUE-R-base, relation-conditioned & $4.90\pm0.24$ & [1.0, 10.1] & $1.24\pm0.23$ & [0.3, 2.4] & $0.40\pm0.23$ \\
 & KLUE-R-base, direct & $3.08\pm0.64$ & [0.1, 7.6] & $1.94\pm0.68$ & [0.5, 4.1] & $1.19\pm0.26$ \\
 & XLM-R-large, relation-conditioned$^\dagger$ & 9.66 & [4.0, 16.0] & 1.49 & [0.6, 2.7] & 0.90 \\
\hline
Candidate-assisted LLM & K-EXAONE, 0-shot & 97.06 & [93.4, 99.6] & 19.55 & [10.9, 30.0] & --- \\
 & K-EXAONE, 3-shot & 64.71 & [55.8, 73.2] & 20.90 & [12.3, 30.9] & --- \\
 & GPT-4o-mini, 0-shot & 60.92 & [52.8, 68.8] & 23.88 & [15.4, 34.5] & --- \\
 & GPT-4o-mini, 3-shot & 30.25 & [22.1, 38.7] & 36.27 & [23.3, 50.0] & --- \\
 & Solar Pro 3, 0-shot & 86.13 & [79.3, 92.3] & 20.90 & [12.2, 31.4] & --- \\
 & Solar Pro 3, 3-shot & 40.76 & [31.7, 50.2] & 21.34 & [12.7, 31.8] & --- \\
\hline
Direct-extraction LLM & GPT-4o-mini, 0-shot & 54.62 & [46.9, 62.3] & 49.55 & [39.9, 58.2] & --- \\
 & GPT-4o-mini, 3-shot & 56.30 & [48.9, 64.0] & 58.81 & [49.4, 67.0] & --- \\
 & K-EXAONE, 0-shot & 58.82 & [51.4, 66.4] & 44.33 & [32.7, 55.3] & --- \\
 & K-EXAONE, 3-shot & 56.72 & [48.5, 65.1] & 63.58 & [53.8, 72.2] & --- \\
 & Solar Pro 3, 0-shot & 60.50 & [53.0, 68.0] & 43.73 & [32.8, 54.7] & --- \\
 & Solar Pro 3, 3-shot & 57.56 & [50.4, 65.1] & 42.99 & [31.9, 54.0] & --- \\
\hline
\end{tabular}%
}
\caption{All systems on the human-verified test (\%; 192 reference-KEEP entities (238 mentions) and 199 reference-MASK entities (670 mentions); ReCoPII-Ko documents). Mean and sample SD over seeds where seeds exist; intervals are 95\% percentile intervals from 2,000 bootstrap resamples of the 100 documents, applied to the seed-averaged rate. Det.\ leakage masks every candidate of the jointly fine-tuned detector. Direct-extraction LLMs mask every span they extract from raw text. $^\dagger$ Single run that met the 1.5\% budget.}
\label{tab:human-gold-all}
\end{table*}

%% file: analysis/human_gold_test/sensitivity_table.tex
\begin{table*}[t]
\centering
\small
\setlength{\tabcolsep}{3.5pt}
\resizebox{\textwidth}{!}{%
\begin{tabular}{lcccccccccc}
\hline
 & \multicolumn{2}{c}{Held out in both splits} & \multicolumn{2}{c}{Unanimous records} & \multicolumn{2}{c}{Labels agree} & \multicolumn{2}{c}{Entity level} & \multicolumn{2}{c}{Weighted estimate} \\
Joint Fine-tuning system & OM & Leak & OM & Leak & OM & Leak & OM & Leak & OM & Leak \\
\hline
A.X+\allowbreak{}Span, relation-conditioned & 0.00 & 0.00 & 2.60 & 1.51 & 1.72 & 1.25 & 1.91 & 3.18 & 9.87 & 0.93 \\
A.X+\allowbreak{}Span, direct & 0.00 & 0.52 & 0.65 & 4.19 & 0.57 & 3.75 & 1.39 & 3.85 & 8.72 & 2.81 \\
A.X+\allowbreak{}Gazetteer+\allowbreak{}CRF+\allowbreak{}MLP, relation-conditioned & 0.00 & 0.69 & 2.38 & 0.65 & 2.59 & 0.70 & 2.43 & 3.02 & 10.58 & 0.98 \\
A.X+\allowbreak{}Gazetteer+\allowbreak{}CRF+\allowbreak{}MLP, direct & 0.00 & 0.17 & 0.00 & 1.34 & 0.00 & 1.30 & 1.04 & 3.18 & 8.09 & 1.82 \\
A.X+\allowbreak{}Gazetteer+\allowbreak{}CRF, relation-conditioned & 0.00 & 1.37 & 3.03 & 1.13 & 2.01 & 1.50 & 2.08 & 4.69 & 10.11 & 1.68 \\
A.X+\allowbreak{}Gazetteer+\allowbreak{}CRF, direct & 0.00 & 0.86 & 1.95 & 1.61 & 1.29 & 1.55 & 1.56 & 3.69 & 9.51 & 1.15 \\
KLUE-R-base, relation-conditioned & 0.42 & 0.86 & 2.60 & 0.86 & 2.44 & 0.80 & 2.78 & 2.85 & 10.58 & 1.06 \\
KLUE-R-base, direct & 0.00 & 0.69 & 0.65 & 1.72 & 0.57 & 1.50 & 1.39 & 3.18 & 8.72 & 1.31 \\
XLM-R-large, relation-conditioned$^\dagger$ & 5.00 & 1.55 & 6.49 & 1.29 & 7.33 & 1.05 & 5.73 & 4.02 & 15.10 & 1.43 \\
\hline
\end{tabular}%
}
\caption{Sensitivity of the human-verified results for Joint Fine-tuning (seed means, \%; OM = over-masking). Held out in both splits: the 34 pairs that are test documents under both the official and the RQ2-B split (80 KEEP and 194 MASK mentions). Unanimous: records on which all four annotators agree (154 and 620 mentions). Labels agree: records whose reference label equals the stored label (232 and 666 mentions; added after the primary results). Entity level counts an entity if any of its mentions is affected. The weighted estimate weights each entity by the inverse of its sampling probability. Over-masking of relation-conditioned heads minus the direct head on all 100 documents, in points with 95\% document-bootstrap intervals: A.X+\allowbreak{}Span +1.12 [-0.15, 3.02]; A.X+\allowbreak{}Gazetteer+\allowbreak{}CRF+\allowbreak{}MLP +2.52 [0.00, 6.05]; A.X+\allowbreak{}Gazetteer+\allowbreak{}CRF +0.70 [0.00, 1.98]; KLUE-R-base +1.82 [0.29, 3.83]. $^\dagger$ Single run that met the 1.5\% budget.}
\label{tab:human-gold-sensitivity}
\end{table*}

%% file: analysis/from_scratch_probe/type_table.tex
\begin{table}[t]
\centering
\small
\setlength{\tabcolsep}{2.5pt}
\resizebox{\columnwidth}{!}{%
\begin{tabular}{@{}lrrrcc@{}}
\hline
 & Train & \multicolumn{2}{c}{Probe mentions} & Over- & Leak- \\
Entity type & KEEP & KEEP & MASK & mask & age \\
\hline
\texttt{WORKPLACE} & 1,175 & 623 & 212 & 24.22 & 3.48 \\
\texttt{DEPARTMENT} & 809 & 580 & 132 & 10.24 & 35.73 \\
\texttt{BANK\_ACCOUNT\_NUMBER} & 537 & 227 & 206 & 27.88 & 2.31 \\
\texttt{ADDRESS} & 505 & 185 & 122 & 61.31 & 0.03 \\
\texttt{EMAIL} & 1,183 & 179 & 88 & 9.73 & 3.22 \\
\texttt{TELEPHONE} & 1,355 & 164 & 65 & 8.18 & 8.08 \\
\texttt{POSITION} & 24 & 108 & 35 & 11.65 & 20.48 \\
\texttt{MOBILE\_PHONE} & 3 & 94 & 163 & 62.54 & 0.20 \\
\texttt{SCHOOL} & 0 & 52 & 169 & 91.75 & 8.75 \\
\texttt{NAME} & 45 & 30 & 1551 & 48.89 & 0.61 \\
\texttt{MAJOR} & 0 & 2 & 41 & 83.33 & 0.41 \\
\texttt{AGE} & 0 & 0 & 52 & --- & 15.14 \\
\texttt{CARD\_NUMBER} & 2 & 0 & 80 & --- & 0.05 \\
\texttt{DATE\_OF\_BIRTH} & 0 & 0 & 137 & --- & 0.09 \\
\texttt{DRIVER\_LICENSE\_NUMBER} & 0 & 0 & 15 & --- & 0.00 \\
\texttt{RRN} & 0 & 0 & 79 & --- & 0.11 \\
\texttt{VEHICLE\_NUMBER} & 0 & 0 & 36 & --- & 1.50 \\
\hline
\end{tabular}}
\caption{Insertion-free probe by entity type (\%): unweighted mean over the 8 three-seed Joint Fine-tuning systems. Train KEEP counts KEEP mentions in the official training split. Descriptive only.}
\label{tab:probe-types}
\end{table}

%% file: analysis/from_scratch_probe/main_table.tex
\begin{table*}[t]
\centering
\small
\setlength{\tabcolsep}{4pt}
\begin{tabular}{llcccc}
\hline
 & & \multicolumn{2}{c}{Official test} & \multicolumn{2}{c}{Insertion-free probe} \\
Detector & Condition & Over-mask $\downarrow$ & Leakage $\downarrow$ & Over-mask $\downarrow$ & Leakage $\downarrow$ \\
\hline
A.X+Gazetteer+CRF+MLP & Detection only & 95.66 & 13.02 & 88.59 & 12.38 \\
 & Frozen Detector & $95.66\pm0.00$ & $13.02\pm0.00$ & $88.59\pm0.00$ & $12.38\pm0.00$ \\
 & Joint FT, relation-conditioned & $2.36\pm0.43$ & $1.20\pm0.34$ & $30.23\pm1.51$ & $3.39\pm0.76$ \\
 & Joint FT, direct & $1.10\pm0.03$ & $1.13\pm0.22$ & $11.33\pm3.81$ & $4.95\pm0.92$ \\
\hline
A.X+Gazetteer+CRF & Detection only & 98.11 & 12.70 & 91.53 & 11.78 \\
 & Frozen Detector & $98.11\pm0.00$ & $12.70\pm0.00$ & $91.53\pm0.00$ & $11.78\pm0.00$ \\
 & Joint FT, relation-conditioned & $2.57\pm0.57$ & $1.32\pm0.24$ & $34.21\pm6.35$ & $3.72\pm0.82$ \\
 & Joint FT, direct & $1.41\pm0.09$ & $0.97\pm0.17$ & $19.10\pm6.90$ & $3.02\pm0.25$ \\
\hline
XLM-R-large & Detection only & 98.86 & 14.46 & 92.25 & 13.82 \\
 & Frozen Detector & $87.70\pm1.71$ & $14.57\pm0.08$ & $88.27\pm1.10$ & $14.29\pm0.17$ \\
 & Joint FT, relation-conditioned & 5.94$^\dagger$ & 0.91$^\dagger$ & 38.41$^\dagger$ & 2.80$^\dagger$ \\
 & Joint FT, direct & \multicolumn{4}{c}{budget not met$^\ddagger$} \\
\hline
GPT-4o-mini & LLM, 0-shot & 62.40 & 16.88 & 41.93 & 15.83 \\
 & LLM, 3-shot & 28.34 & 24.28 & 16.76 & 25.89 \\
K-EXAONE & LLM, 0-shot & 95.37 & 13.17 & 85.29 & 12.19 \\
 & LLM, 3-shot & 61.94 & 15.30 & 51.52 & 14.14 \\
Solar Pro 3 & LLM, 0-shot & 80.11 & 15.01 & 53.83 & 16.43 \\
 & LLM, 3-shot & 42.97 & 15.59 & 34.27 & 14.61 \\
\hline
\end{tabular}
\caption{The systems of Table~\ref{tab:rq3-main} on the insertion-free probe (\%). Left: the official 404-document test (1,750 KEEP and 11,045 MASK mentions). Right: 402 documents (2,244 KEEP and 3,183 MASK mentions) that a separate pipeline writes as a whole from a planned relation graph, with no sentence inserted into an existing document; their labels are generated and not human-verified. Every system is applied with the checkpoints, thresholds, and prompts of the left side; nothing is retrained or tuned on the probe. Values with $\pm$ are mean and sample SD over contextual seeds 42--44; Detection only uses the seed-42 detector, and the LLM rows are single runs on the A.X+Gazetteer+CRF detection-only candidates. $^\dagger$ The only XLM-R-large run that met the 1.5\% budget; $^\ddagger$ no direct-head run met it.}
\label{tab:probe-main}
\end{table*}

%% file: analysis/from_scratch_probe/all_systems_table.tex
\begin{table*}[t]
\centering
\small
\setlength{\tabcolsep}{3.5pt}
\resizebox{\textwidth}{!}{%
\begin{tabular}{llcccccccc}
\hline
 & & \multicolumn{2}{c}{Over-mask $\downarrow$} & \multicolumn{2}{c}{Leakage $\downarrow$} & \multicolumn{2}{c}{Over-mask by type group} & Detector & Released \\
Detector & Variant & Mean $\pm$ SD & 95\% CI & Mean $\pm$ SD & 95\% CI & KEEP-supported & Other & leakage & share \\
\hline
A.X+\allowbreak{}Span & Relation-conditioned & $28.42\pm3.21$ & [26.1, 30.7] & $2.45\pm0.63$ & [1.8, 3.2] & $24.99\pm2.76$ & $51.86\pm7.17$ & $0.91\pm0.11$ & 68.98 \\
 & Direct & $26.80\pm12.79$ & [24.8, 28.9] & $2.03\pm0.66$ & [1.6, 2.6] & $25.06\pm12.11$ & $38.69\pm17.42$ & $0.79\pm0.03$ & 72.58 \\
\hline
A.X+\allowbreak{}Gazetteer+\allowbreak{}CRF+\allowbreak{}MLP & Relation-conditioned & $30.23\pm1.51$ & [28.1, 32.2] & $3.39\pm0.76$ & [2.6, 4.3] & $27.19\pm2.33$ & $51.05\pm6.89$ & $2.39\pm0.75$ & 51.84 \\
 & Direct & $11.33\pm3.81$ & [9.9, 12.8] & $4.95\pm0.92$ & [4.0, 6.0] & $9.70\pm3.82$ & $22.49\pm5.19$ & $2.74\pm0.58$ & 82.51 \\
\hline
A.X+\allowbreak{}Gazetteer+\allowbreak{}CRF & Relation-conditioned & $34.21\pm6.35$ & [32.1, 36.3] & $3.72\pm0.82$ & [2.9, 4.6] & $31.07\pm6.35$ & $55.71\pm6.50$ & $2.94\pm1.21$ & 45.22 \\
 & Direct & $19.10\pm6.90$ & [17.1, 21.0] & $3.02\pm0.25$ & [2.3, 3.8] & $16.48\pm6.74$ & $37.06\pm12.48$ & $2.00\pm0.10$ & 71.82 \\
\hline
KLUE-R-base & Relation-conditioned & $26.29\pm2.51$ & [24.2, 28.4] & $3.72\pm0.72$ & [3.0, 4.5] & $20.67\pm3.19$ & $64.80\pm3.17$ & $2.76\pm0.84$ & 58.72 \\
 & Direct & $20.87\pm2.71$ & [18.8, 22.9] & $3.86\pm2.21$ & [3.1, 4.7] & $15.58\pm2.97$ & $57.11\pm2.05$ & $1.90\pm0.41$ & 69.28 \\
\hline
XLM-R-large & Relation-conditioned$^\dagger$ & 38.41 & [35.5, 41.3] & 2.80 & [2.2, 3.5] & 33.50 & 72.03 & 1.95 & 51.35 \\
\hline
\end{tabular}}
\caption{Joint Fine-tuning on the insertion-free probe (\%), with the thresholds selected on official validation for the 1.5\% budget. Intervals resample the 402 documents 2,000 times and apply to the seed-averaged rate. KEEP-supported types are the six entity types with at least 500 KEEP mentions in the official training split (ADDRESS, BANK\_ACCOUNT\_NUMBER, DEPARTMENT, EMAIL, TELEPHONE, WORKPLACE; 1,958 probe KEEP mentions); the other types have 286 probe KEEP mentions and at most 45 KEEP mentions each in training. Detector leakage masks every candidate of the fine-tuned detector. Released share is the percentage of KEEP mentions masked by that detector which the decision releases. Relation-conditioned minus direct over-masking, in points with paired 95\% intervals: A.X+Span +1.62 [-0.22, +3.51]; A.X+Gazetteer+CRF+MLP +18.89 [+17.22, +20.67]; A.X+Gazetteer+CRF +15.11 [+13.44, +16.75]; KLUE-R-base +5.42 [+3.96, +7.03]. $^\dagger$ Single run that met the budget.}
\label{tab:probe-all}
\end{table*}

%% file: analysis/from_scratch_probe/reference_table.tex
\begin{table*}[t]
\centering
\small
\setlength{\tabcolsep}{4pt}
\begin{tabular}{llcccc}
\hline
 & & \multicolumn{2}{c}{Official test} & \multicolumn{2}{c}{Insertion-free probe} \\
Group & System & Over-mask $\downarrow$ & Leakage $\downarrow$ & Over-mask $\downarrow$ & Leakage $\downarrow$ \\
\hline
Detection only & A.X+\allowbreak{}Span & 99.83 & 9.15 & 95.99 & 8.07 \\
 & A.X+\allowbreak{}Gazetteer+\allowbreak{}CRF+\allowbreak{}MLP & 95.66 & 13.02 & 88.59 & 12.38 \\
 & A.X+\allowbreak{}Gazetteer+\allowbreak{}CRF & 98.11 & 12.70 & 91.53 & 11.78 \\
 & A.X+\allowbreak{}GatedGazetteer+\allowbreak{}CRF & 96.91 & 10.96 & 94.16 & 9.27 \\
 & KLUE-R-base & 99.26 & 10.97 & 91.89 & 10.46 \\
 & KLUE-BERT-base & 99.54 & 11.75 & 92.91 & 9.61 \\
 & KLUE-BERT-base+\allowbreak{}CRF & 99.26 & 10.76 & 90.95 & 9.52 \\
 & XLM-R-large & 98.86 & 14.46 & 92.25 & 13.82 \\
\hline
Frozen Detector & A.X+\allowbreak{}Span & 99.60 & 9.16 & $95.51\pm0.47$ & $8.14\pm0.05$ \\
 & A.X+\allowbreak{}Gazetteer+\allowbreak{}CRF+\allowbreak{}MLP & 95.66 & 13.02 & $88.59\pm0.00$ & $12.38\pm0.00$ \\
 & A.X+\allowbreak{}Gazetteer+\allowbreak{}CRF & 98.11 & 12.70 & $91.53\pm0.00$ & $11.78\pm0.00$ \\
 & A.X+\allowbreak{}GatedGazetteer+\allowbreak{}CRF & 94.55 & 11.00 & $92.72\pm2.09$ & $9.40\pm0.13$ \\
 & KLUE-R-base & 99.26 & 10.97 & $91.89\pm0.00$ & $10.46\pm0.00$ \\
 & KLUE-BERT-base & 99.54 & 11.75 & $92.91\pm0.00$ & $9.61\pm0.00$ \\
 & KLUE-BERT-base+\allowbreak{}CRF & 95.05 & 10.84 & $88.83\pm1.47$ & $9.72\pm0.21$ \\
 & XLM-R-large & 87.70 & 14.57 & $88.27\pm1.10$ & $14.29\pm0.17$ \\
\hline
Direct-extraction LLM & GPT-4o-mini, 0-shot & 51.77 & 45.67 & 43.58 & 45.40 \\
 & GPT-4o-mini, 3-shot & 53.83 & 58.60 & 38.24 & 49.70 \\
 & K-EXAONE, 0-shot & 58.86 & 51.44 & 36.41 & 46.15 \\
 & K-EXAONE, 3-shot & 56.46 & 59.51 & 45.45 & 46.15 \\
 & Solar Pro 3, 0-shot & 59.66 & 48.98 & 43.58 & 46.75 \\
 & Solar Pro 3, 3-shot & 57.77 & 53.42 & 43.98 & 49.17 \\
\hline
\end{tabular}
\caption{Reference systems on the insertion-free probe (\%), beside their official-test values. Detection only masks every candidate of the seed-42 KDPII-trained detector; Frozen Detector uses its registered thresholds (mean $\pm$ sample SD over contextual seeds 42--44, official-test means on the left). Direct-extraction LLMs mask every span they extract from raw text (single runs). The Frozen Detector rows and three Detection only rows were added after the first partial scoring of the probe.}
\label{tab:probe-reference}
\end{table*}

%% file: appendices/augmentation_prompts.tex
\raggedbottom
\newtcolorbox{promptpanel}[1][]{
  enhanced,
  colback=black!4,
  colframe=black!70,
  boxrule=0.7pt,
  arc=2mm,
  left=4mm,
  right=4mm,
  top=2mm,
  bottom=2mm,
  before skip=8pt,
  after skip=4pt,
  fontupper=\small\linespread{1.0}\selectfont,
  before upper={\setlength{\parindent}{0pt}\setlength{\parskip}{0pt}\raggedright},
  #1
}
\tcbset{promptsplit/.style={breakable, break at=#1, title after break={(continued)}, fonttitle=\small\itshape, toptitle=0.5mm, bottomtitle=0.5mm}}
\newcommand{\promptnextcolumn}{\par\vfill\newpage}
\newcommand{\promptnextpage}{\par\clearpage}

\twocolumn[{%
\section{LLM Prompts}
\label{app:augmentation-prompts}

This section reports the prompts used in the three LLM-based steps of the augmentation pipeline: relation-graph planning (Section~\ref{sec:graph-planning}), placeholder-based text realization (Section~\ref{sec:text-realization}), and independent review (Section~\ref{sec:independent-review}). It then reports the prompts of the two LLM evaluations: direct PII extraction in RQ1 (Appendix~\ref{app:prompt-rq1}) and document-quality judgment in RQ2-A (Appendix~\ref{app:prompt-judge}). For each system and user prompt, we provide the original Korean prompt followed by an English translation. Payload fields are reported separately in Tables~\ref{tab:planner-payload}--\ref{tab:reviewer-payload}. Output schemas are omitted. Deterministic value filling and document acceptance do not call an LLM.

\subsection{Relation-Graph Planning}
\label{app:prompt-planner}

We use GPT-4o-mini with a maximum of 3,000 output tokens. Rule 1-1 in the user prompt is included only when \texttt{reusable\_existing\_entities} is nonempty.

\begin{center}
{\scriptsize
\setlength{\tabcolsep}{4pt}
\renewcommand{\arraystretch}{1.08}
\begin{tabular}{p{0.29\textwidth}p{0.65\textwidth}}
\hline
Payload field & Description \\
\hline
\texttt{domain} & Document domain. \\
\texttt{document\_text} & Full source document. \\
\texttt{existing\_entities} & Source entities with \texttt{entity\_id}, \texttt{entity\_type}, and \texttt{form}. \\
\texttt{existing\_relations} & Source relations with source, relation, target, and privacy label. \\
\texttt{allowed\_relations} & Domain ontology relations and their meanings, without example sentences. \\
\texttt{allowed\_type\_combinations} & Permitted source-type, relation, and target-type combinations. \\
\texttt{reusable\_\allowbreak existing\_\allowbreak entities} & Source organization entities that may anchor a new relation in the current document. \\
\hline
\end{tabular}\par}
\captionof{table}{Payload supplied to the relation-graph planner.}
\label{tab:planner-payload}
\end{center}
\vspace{0.5\baselineskip}
}]

\begin{promptpanel}
{\korean
System Prompt (Korean)\par\medskip
너는 한국어 개인정보 데이터셋의 관계 주석 설계자다.

주어진 문서에 이미 있는 내용은 절대 바꾸지 않는다. 대신 이 문서에 자연스럽게 덧붙일 수 있는 ‘공개된 조직 정보’ 노드와 그 사이의 관계만 계획한다.

가장 중요한 제약은 이것이다. 네가 새로 만드는 모든 노드는 그 노드에 연결된 모든 관계가 NON\_PII여야 한다. 즉 특정 개인에게 귀속되는 정보가 아니라, 조직이 공식적으로 공개하고 공동으로 관리하는 정보여야 한다.

개인에게 귀속됨 (PII): 김민수의 휴대폰 번호, 박지영의 소속 부서

조직이 공개·공동 관리 (NON\_PII): 회사 대표번호, 공식 정산 계좌, 공개 조직도

‘공개’는 공공기관·정부기관·공기업을 뜻하지 않는다. 민간 회사도 대표번호와 조직 구성을 공개한다. 그런 근거 없는 기관 표현은 쓰지 않는다.

계획만 세운다. 문장은 다음 단계에서 쓴다.
}
\end{promptpanel}
\nopagebreak[4]
\captionof{figure}{Korean system prompt for relation-graph planning.}
\label{fig:planner-system-ko}
\promptnextcolumn

\begin{promptpanel}
System Prompt (English Translation)\par\medskip
You are a relation-annotation designer for a Korean PII dataset.

Never alter any content already present in the given document. Plan only the nodes representing ``publicly disclosed organizational information'' that can be added naturally to the document and the relations among those nodes.

The most important constraint is that every relation incident to every node you create must be labeled NON\_PII. In other words, the information must not be attributable to a specific individual; it must be information officially disclosed and jointly managed by an organization.

Attributed to an individual (PII): Kim Min-su's mobile-phone number; Park Ji-young's department affiliation.

Publicly disclosed and jointly managed by an organization (NON\_PII): a company's main telephone number; an official settlement account; a public organizational chart.

``Public'' does not mean that the organization is a public agency, government body, or state-owned enterprise. Private companies also disclose their main telephone numbers and organizational structures. Do not use unsupported descriptions that imply a public institution.

Produce only the plan. The sentences will be written in the next stage.
\end{promptpanel}
\nopagebreak[4]
\captionof{figure}{English translation of the planner system prompt.}
\label{fig:planner-system-en}
\promptnextpage

\begin{promptpanel}
{\korean
User Prompt (Korean)\par\medskip
아래 문서에 덧붙일 NON\_PII 노드와 관계를 계획하라.

\texttt{\{payload JSON\}}

지켜야 할 규칙

1. \texttt{new\_nodes}는 3개 이상 6개 이하로 만든다. 노드 타입은 스키마에 나열된 공개 조직 정보 타입만 쓴다.

1-1. \texttt{reusable\_existing\_entities}에 있는 원본 엔티티를 관계의 출발점으로 쓸 수 있다. 문서 소재에 자연스럽게 맞으면 쓰고, 맞지 않으면 신규 노드끼리 연결한다.

2. \texttt{new\_relations}의 \texttt{type\_combination}은 \texttt{allowed\_type\_combinations}의 \texttt{combination} 값 중 하나다. 고른 조합의 source 타입·target 타입이 실제 노드 타입과 같아야 한다. 예를 들어 \texttt{WORKPLACE -CONTACT\_ASSOCIATION-> TELEPHONE}을 골랐다면 \texttt{source}는 WORKPLACE 노드여야 하고 \texttt{target}은 TELEPHONE 노드여야 한다.

3. 새 노드에 연결되는 관계는 전부 NON\_PII여야 한다. 하나라도 개인에게 귀속되면 그 노드는 목적을 잃는다. \texttt{non\_pii\_rationale}에 왜 개인 귀속이 아닌지 적는다.

4. 새 노드는 모두 최소 하나의 관계에 연결돼야 한다. 떠 있는 노드는 만들지 않는다.

5. 같은 엔티티 쌍에는 관계를 하나만 만든다. 같은 두 노드를 방향만 바꿔 두 번 연결하는 것도 안 된다. source와 target에 같은 노드를 쓰면 안 된다. \texttt{WORKPLACE -...-> WORKPLACE}처럼 양쪽 타입이 같은 조합을 고르려면 WORKPLACE 노드를 서로 다른 것으로 두 개 만들어야 한다.

6. \texttt{existing\_relations}의 기존 관계를 다시 적지 않는다. 새로 더할 것만 적는다.

7. 이 문서의 종류와 소재에 맞는 조직을 고른다. 문서와 무관한 소재를 끌어오지 않는다.

8. \texttt{public\_context}와 \texttt{role}에는 실제 엔티티 값을 미리 정해 쓰지 않는다.
}
\end{promptpanel}
\nopagebreak[4]
\captionof{figure}{Korean user prompt for relation-graph planning.}
\label{fig:planner-user-ko}
\promptnextcolumn

\begin{promptpanel}
User Prompt (English Translation)\par\medskip
Plan the NON\_PII nodes and relations to be added to the document below.

\texttt{\{payload JSON\}}

Rules

1. Create between three and six \texttt{new\_nodes}. Use only the public organizational-information node types listed in the schema.

1-1. A source entity in \texttt{reusable\_existing\_entities} may serve as the source of a relation. Use it when it fits the document naturally; otherwise, connect the new nodes to one another.

2. The \texttt{type\_combination} of each item in \texttt{new\_relations} must be one of the \texttt{combination} values in \texttt{allowed\_type\_combinations}. The source and target types in the selected combination must match the actual node types. For example, if \texttt{WORKPLACE -CONTACT\_ASSOCIATION-> TELEPHONE} is selected, \texttt{source} must be a WORKPLACE node and \texttt{target} must be a TELEPHONE node.

3. Every relation incident to a new node must be NON\_PII. If even one relation attributes the node to an individual, the node no longer serves its intended purpose. Explain in \texttt{non\_pii\_rationale} why the relation is not attributable to an individual.

4. Every new node must participate in at least one relation. Do not create isolated nodes.

5. Create only one relation for each entity pair. Do not connect the same two nodes twice by reversing their direction, and do not use the same node as both source and target. A combination such as \texttt{WORKPLACE -...-> WORKPLACE} requires two distinct WORKPLACE nodes.

6. Do not repeat any relation in \texttt{existing\_relations}. Report only relations to be added.

7. Select organizations that fit the type and subject matter of the document. Do not introduce unrelated subject matter.

8. Do not preselect or write literal entity values in \texttt{public\_context} or \texttt{role}.
\end{promptpanel}
\nopagebreak[4]
\captionof{figure}{English translation of the planner user prompt.}
\label{fig:planner-user-en}

\twocolumn[{%
\subsection{Placeholder-Based Text Realization}
\label{app:prompt-realizer}

We use GPT-4o-mini with a maximum of 3,000 output tokens.

\begin{center}
{\scriptsize
\setlength{\tabcolsep}{4pt}
\renewcommand{\arraystretch}{1.08}
\begin{tabular}{p{0.29\textwidth}p{0.65\textwidth}}
\hline
Payload field & Description \\
\hline
\texttt{domain} & Document domain. \\
\texttt{never\_write\_these\_values} & Literal source PII values prohibited in inserted sentences. \\
\texttt{insertion\_zones} & Sentence indices grouped into the first, middle, and final thirds of the source document. \\
\texttt{original\_document} & Source sentences with \texttt{sent\_idx} and \texttt{sentence}. \\
\texttt{public\_context} & Public organizational context produced by the planner. \\
\texttt{entities\_to\_introduce} & New nodes with placeholder, entity type, role, and \texttt{how\_to\_write\_it}. \\
\texttt{relations\_to\_express} & New source--relation--target tuples expressed with placeholders. \\
\texttt{existing\_\allowbreak entity\_\allowbreak placeholders} & Placeholders and values for source entities used as relation endpoints. \\
\hline
\end{tabular}\par}
\captionof{table}{Payload supplied to the sentence realizer. Type-specific values of \texttt{how\_to\_write\_it} are reported in Table~\ref{tab:type-writing-instructions}.}
\label{tab:realizer-payload}
\end{center}

\begin{center}
{\scriptsize
\setlength{\tabcolsep}{4pt}
\renewcommand{\arraystretch}{1.08}
\begin{tabular}{p{0.19\textwidth}p{0.36\textwidth}p{0.39\textwidth}}
\hline
Entity type & Korean instruction & English translation \\
\hline
WORKPLACE & {\korean 회사·기관 이름으로 쓴다. 문서 절차에서 맡는 역할을 밝힌다.} & Use a company or institution name and state its role in the document's procedure. \\
DEPARTMENT & {\korean 조직 안의 부서 이름과 담당 업무를 밝힌다.} & Use a department name and state the work handled by that department. \\
SCHOOL & {\korean 학교·교육기관 이름으로 쓴다.} & Use the name of a school or educational institution. \\
POSITION & {\korean 공개된 직제상의 직위로 쓰고 특정 개인의 직함으로 쓰지 않는다.} & Use a publicly listed organizational position, not a particular person's title. \\
MAJOR & {\korean 학교가 개설한 전공·과정 이름으로 쓴다.} & Use the name of a major or program offered by a school. \\
TELEPHONE & {\korean 조직 대표번호나 부서 문의번호로 쓰고 개인 번호로 읽히지 않게 한다.} & Present it as an organization's main or departmental inquiry number, not a personal number. \\
EMAIL & {\korean 조직의 접수·문의 창구로 쓰고 개인 메일로 읽히지 않게 한다.} & Present it as an organizational submission or inquiry channel, not a personal email address. \\
MOBILE\_PHONE & {\korean 조직이 공개한 업무용·현장 운영 회선으로 쓴다.} & Present it as a public business or field-operation line managed by an organization. \\
ADDRESS & {\korean 조직의 사업장·사무실·캠퍼스 소재지임을 밝힌다.} & State that it is the location of an organization's workplace, office, or campus. \\
VEHICLE\_NUMBER & {\korean 조직이 보유·운행하는 업무 차량임을 밝힌다.} & State that it identifies a business vehicle owned or operated by an organization. \\
BANK\_ACCOUNT\_NUMBER & {\korean 조직 명의의 공식 정산·환불·납입 계좌임을 밝힌다.} & State that it is an official organizational account for settlement, refunds, or payment. \\
\hline
\end{tabular}\par}
\captionof{table}{Type-specific writing instructions included in the realizer payload. The table summarizes the operational wording while preserving the attribution constraint supplied to the realizer.}
\label{tab:type-writing-instructions}
\end{center}
\vspace{0.5\baselineskip}
}]

\begin{promptpanel}
{\korean
System Prompt (Korean)\par\medskip
너는 한국어 실무 문서를 쓰는 작성자다.

이미 완성된 문서에 몇 문장을 끼워 넣는 일을 한다. 원문은 한 글자도 고치지 않는다. 너는 새로 넣을 문장과 그 문장을 어느 원문 문장 뒤에 넣을지만 돌려준다.

새 문장에는 실제 값 대신 placeholder를 쓴다.

\texttt{<WORKPLACE:E7>} --- 값이 그대로 들어간다.

\texttt{<WORKPLACE:E7>\{josa:}{은/는}\texttt{\}} --- 값의 받침에 맞는 조사가 붙는다.

조사 marker에 쓸 수 있는 쌍: 은/는, 이/가, 을/를, 과/와, 으로/로, 이다/다, 이며/며, 이고/고, 이라/라, 이란/란, 이나/나, 이든/든, 으로서/로서, 으로써/로써, 으로도/로도, 으로는/로는, 으로만/로만, 과도/와도, 과는/와는, 과의/와의, 과만/와만.

placeholder 바로 뒤에 조사를 직접 붙이면 “한빛물류은” 같은 문장이 되므로 반드시 marker를 쓴다.
}
\end{promptpanel}
\nopagebreak[4]
\captionof{figure}{Korean system prompt for sentence realization and insertion.}
\label{fig:realizer-system-ko}
\promptnextcolumn

\begin{promptpanel}
System Prompt (English Translation)\par\medskip
You are a writer of Korean business documents.

Your task is to insert several sentences into an already completed document. Do not change a single character of the source document. Return only the new sentences and the source sentence after which each one should be inserted.

Use placeholders rather than literal values in the new sentences.

\texttt{<WORKPLACE:E7>} --- The value is inserted without modification.

\texttt{<WORKPLACE:E7>\{josa:}{\korean 은/는}\texttt{\}} --- The Korean particle is selected according to whether the value ends in a final consonant.

The particle marker may use the following pairs: {\korean 은/는, 이/가, 을/를, 과/와, 으로/로, 이다/다, 이며/며, 이고/고, 이라/라, 이란/란, 이나/나, 이든/든, 으로서/로서, 으로써/로써, 으로도/로도, 으로는/로는, 으로만/로만, 과도/와도, 과는/와는, 과의/와의, and 과만/와만}.

Appending a particle directly after a placeholder can produce an ungrammatical expression such as ``{\korean 한빛물류은}.'' Always use the marker.
\end{promptpanel}
\nopagebreak[4]
\captionof{figure}{English translation of the realizer system prompt.}
\label{fig:realizer-system-en}
\promptnextpage

\begin{promptpanel}[promptsplit=355pt/0pt]
{\korean
User Prompt (Korean)\par\medskip
원문 뒤에 끼워 넣을 문장을 써라.

\texttt{\{payload JSON\}}

지켜야 할 규칙

1. \texttt{original\_document}의 문장은 하나도 고치지 않는다. 출력에 원문을 다시 담지 않는다.

2. \texttt{insertions}는 서로 다른 \texttt{after\_sent\_idx} 2곳 이상에 나눠 넣는다. 전체 추가 문장은 6개를 넘지 않는다.

2-1. \texttt{insertion\_zones}는 원문을 앞부분·가운데·뒷부분으로 나눈 것이다. 되도록 서로 다른 구간에서 골라 문서 전반에 퍼지게 넣는다. 바로 옆 문장들에 몰아넣지 않는다. 각 구간에서 그 자리의 소재와 이어지는 내용을 쓴다.

3. \texttt{relations\_to\_express}의 관계가 모두 문서에서 읽어낼 수 있어야 한다. 두 endpoint가 꼭 같은 문장에 있을 필요는 없지만, 앞뒤 문맥만 보고 그 관계를 추출할 수 있을 만큼은 분명해야 한다.

3-1. 각 placeholder의 \texttt{how\_to\_write\_it}을 따르고, 조직의 공개 정보라는 근거가 문서에 드러나게 쓴다.

4. \texttt{entities\_to\_introduce}의 placeholder는 전부 최소 한 번 쓴다. 목록에 없는 placeholder나 새 entity id는 만들지 않는다.

4-1. \texttt{relations\_to\_express}의 관계에 \texttt{existing\_entity\_placeholders}의 엔티티가 끝점으로 나오면 그 placeholder도 반드시 문장에 쓴다. 그 조직의 이름을 직접 적거나 다른 조직으로 바꿔 쓰면 안 된다. 예를 들어 \texttt{<WORKPLACE:E9>}가 계좌의 소유 주체라면 “한서대학교의 정산 계좌인 …”은 틀리고 “\texttt{<WORKPLACE:E9>}의 정산 계좌인 …”은 맞다.

5. 관계 하나를 문장 하나로 기계적으로 풀지 않는다. 하나의 업무·절차·안내 안에서 여러 관계가 자연스럽게 드러나게 쓴다.

6. 실제 이름·번호·계좌를 쓰지 않는다. 반드시 placeholder만 쓴다.

7. “이 정보는 공개 정보다”, “개인정보가 아니다” 같은 메타 설명 문장을 쓰지 않는다. 그냥 업무 문서처럼 쓴다.

8. 삽입 지점 바로 앞 문장을 읽고 그 어미와 시제, 경어체를 그대로 따라 쓴다. 문장을 쉼표나 “~하며”로 끊고 넘기지 않는다.

9. 공공기관·정부기관·국가기관이라는 근거 없는 표현을 쓰지 않는다.

10. \texttt{role}을 문장에 그대로 옮겨 적거나 placeholder 앞에서 같은 뜻을 반복하지 않는다.

11. 새 placeholder는 모두 조직이 공식적으로 공개하고 공동 관리하는 정보로 쓴다. 문서에 나오는 개인(고객님, 귀하, 본인, 갑, 을, 환자, 신청인, 의뢰인 등)의 연락처·계좌·소속으로 쓰면 그 노드는 개인 귀속이 되어 문서 전체가 탈락한다.

12. 연락처를 죽 늘어놓는 안내문처럼 쓰지 않는다. 원문에서 이미 진행 중인 사건·절차·업무의 일부로 녹여 쓴다. 같은 문형을 반복하지 않는다.

13. 타입에 맞는 말로 가리킨다. WORKPLACE는 회사·기관 이름이고 DEPARTMENT는 그 안의 부서 이름이다. WORKPLACE placeholder 앞에 “부서”를 붙이지 않는다.

14. \texttt{never\_write\_these\_values}의 원문 값을 새 문장에 그대로 쓰지 않는다. 마스킹에서 빠질 수 있다. 그 사람은 “담당자”, “신청인” 같은 보통명사로 가리킨다.

15. 삽입된 티가 나지 않을 만큼 원문 소재·문체·흐름과 자연스럽게 이어 쓴다.
}
\end{promptpanel}
\nopagebreak[4]
\captionof{figure}{Korean user prompt for sentence realization and insertion.}
\label{fig:realizer-user-ko}
\promptnextpage

\begin{promptpanel}[promptsplit=420pt/0pt]
User Prompt (English Translation)\par\medskip
Write the sentences to be inserted after sentences in the source document.

\texttt{\{payload JSON\}}

Rules

1. Do not alter any sentence in \texttt{original\_document}. Do not reproduce the source document in the output.

2. Distribute \texttt{insertions} across at least two distinct \texttt{after\_sent\_idx} values. Add no more than six sentences in total.

2-1. \texttt{insertion\_zones} divides the source document into its first, middle, and final thirds. When possible, choose different zones and distribute insertions across the document rather than clustering them around adjacent sentences. At each location, write content that continues the local subject matter.

3. Every relation in \texttt{relations\_to\_express} must be recoverable from the document. Its endpoints need not occur in the same sentence, but the relation must be clear enough to extract from the surrounding context.

3-1. Follow each placeholder's \texttt{how\_to\_write\_it} instruction and make the evidence that it is public organizational information explicit in the document.

4. Use every placeholder in \texttt{entities\_to\_introduce} at least once. Do not create a placeholder or entity ID absent from the list.

4-1. If an entity in \texttt{existing\_entity\_placeholders} is an endpoint of a relation in \texttt{relations\_to\_express}, its placeholder must appear in the sentence. Do not write the organization's literal name or replace it with another organization. For example, if \texttt{<WORKPLACE:E9>} owns an account, ``the settlement account of Hanseo University'' is invalid, whereas ``the settlement account of \texttt{<WORKPLACE:E9>}'' is valid.

5. Do not mechanically express one relation per sentence. Express multiple relations naturally within a single task, procedure, or notice.

6. Do not write literal names, numbers, or accounts. Use only placeholders.

7. Do not add metalinguistic statements such as ``this is public information'' or ``this is not personal information.'' Write ordinary business-document prose.

8. Read the sentence immediately preceding an insertion point and preserve its ending, tense, and honorific style. Do not leave a sentence hanging after a comma or a connective ending such as ``-hamyeo.''

9. Do not introduce unsupported descriptions such as public agency, government agency, or national institution.

10. Do not copy \texttt{role} verbatim into a sentence or repeat the same meaning immediately before a placeholder.

11. Present every new placeholder as information officially disclosed and jointly managed by an organization. If it is written as the contact information, account, or affiliation of a person in the document, the node becomes individually attributable and the entire document is rejected.

12. Do not produce a notice that merely lists contact details. Integrate the new content into an event, procedure, or task already underway in the source document, and avoid repetitive sentence patterns.

13. Refer to each type appropriately. WORKPLACE is the name of a company or institution, whereas DEPARTMENT is a unit within it. Do not place the word ``department'' before a WORKPLACE placeholder.

14. Do not reproduce any source value in \texttt{never\_write\_these\_values} in a new sentence because it may escape masking. Refer to the person with a common noun such as ``staff member'' or ``applicant.''

15. Connect the inserted sentences naturally to the source document's subject matter, style, and flow so that they do not appear inserted.
\end{promptpanel}
\nopagebreak[4]
\captionof{figure}{English translation of the realizer user prompt.}
\label{fig:realizer-user-en}

\twocolumn[{%
\subsection{Independent Relation and Privacy Review}
\label{app:prompt-reviewer}

We use \texttt{gpt-5-mini} with minimal reasoning effort and a maximum of 6,000 output tokens. The reviewer does not receive the planned privacy labels, indicators of which entities or relations are new, or the \texttt{non\_pii\_rationale}.

\begin{center}
{\scriptsize
\setlength{\tabcolsep}{4pt}
\renewcommand{\arraystretch}{1.08}
\begin{tabular}{p{0.29\textwidth}p{0.65\textwidth}}
\hline
Payload field & Description \\
\hline
\texttt{domain} & Document domain. \\
\texttt{allowed\_relations} & Domain ontology relations and their meanings. \\
\texttt{entities} & All entities in the completed document, with ID, type, and surface form. \\
\texttt{document} & Completed sentences after value filling, with sentence indices. \\
\texttt{relations\_to\_check} & Candidate edges with edge ID, source, relation, and target; privacy labels are withheld. \\
\texttt{recently\_\allowbreak inserted\_\allowbreak sent\_\allowbreak idx} & Indices of inserted sentences used for the naturalness assessment. \\
\hline
\end{tabular}\par}
\captionof{table}{Payload supplied to the independent reviewer.}
\label{tab:reviewer-payload}
\end{center}
\vspace{0.5\baselineskip}
}]

\begin{promptpanel}
{\korean
System Prompt (Korean)\par\medskip
너는 한국어 개인정보 데이터셋의 독립 검수자다.

문서가 어떻게 만들어졌는지, 어떤 라벨이 의도됐는지는 모른다. 알 필요도 없다. 너는 완성된 문서만 읽고 판단한다.

Privacy 판단 기준

PII: 특정 개인에게 식별자·연락처·주소·소속·학력·금융정보를 귀속시키거나 개인을 식별·연결하는 데 기여하는 관계.

NON\_PII: 실제 의미 관계이지만 개인 귀속이 아닌 관계. 조직이 공식적으로 공개하고 공동으로 관리하는 채널·조직 구성·정산 계좌 등.

값이 어떻게 생겼는지로 판단하지 않는다. 문서에서 그 값이 어떤 역할로 쓰이는지, 어떤 관계의 끝점인지로 판단한다.

판단 근거가 아님: 이메일에 사람 이름이나 생년월일이 들어 있다는 사실; 번호가 휴대폰 형식인지 유선 형식인지; Entity Type 이름, Relation 이름, 이메일 도메인.

판단 근거임: 문서가 그 값을 특정 개인의 것으로 귀속시키는가; 문서가 그 값을 조직의 공식 창구·공동 관리 자산으로 쓰는가.

예를 들어 “문의처: 고객지원팀 02-000-0000 abc19800101@naver.com”에서 그 메일은 생김새가 개인 메일이어도 문서 안에서는 부서 문의 창구다. 이때는 NON\_PII다. 반대로 “김민수 씨의 연락처는 02-000-0000입니다”는 번호가 유선이어도 PII다.

문서 안에 판정 근거가 없으면 AMBIGUOUS다.
}
\end{promptpanel}
\nopagebreak[4]
\captionof{figure}{Korean system prompt for independent relation and privacy review.}
\label{fig:reviewer-system-ko}
\promptnextcolumn

\begin{promptpanel}
System Prompt (English Translation)\par\medskip
You are an independent reviewer of a Korean PII dataset.

You do not know how the document was created or which labels were intended, and you do not need to know. Judge only the completed document.

Privacy criteria

PII: A relation that attributes an identifier, contact detail, address, affiliation, education record, or financial information to a specific individual, or that helps identify or link an individual.

NON\_PII: A genuine semantic relation that is not attributable to an individual, such as an organization's officially disclosed and jointly managed channel, organizational structure, or settlement account.

Do not judge a value by its appearance. Judge the role it plays in the document and the relation for which it serves as an endpoint.

Not evidence: an email address contains a person's name or date of birth; a number resembles a mobile or landline format; the entity-type name, relation name, or email domain.

Valid evidence: the document attributes the value to a specific individual; the document uses the value as an organization's official channel or jointly managed asset.

For example, in ``Contact: Customer Support Team, 02-000-0000, abc19800101@naver.com,'' the email address functions as the department's inquiry channel even though it looks like a personal address, so the relation is NON\_PII. Conversely, ``Kim Min-su's contact number is 02-000-0000'' is PII even though the number resembles a landline.

If the document contains no evidence supporting a decision, label it AMBIGUOUS.
\end{promptpanel}
\nopagebreak[4]
\captionof{figure}{English translation of the reviewer system prompt.}
\label{fig:reviewer-system-en}
\promptnextpage

\begin{promptpanel}
{\korean
User Prompt (Korean)\par\medskip
아래 문서를 검수하라.

\texttt{\{payload JSON\}}

검수 항목

1. \texttt{relations\_to\_check}의 각 관계를 정확히 한 번씩 판정한다.

\quad extractable: 문서 문맥에서 그 관계를 실제로 읽어낼 수 있는가. 두 endpoint가 같은 문장에 있을 필요는 없다.

\quad observed\_privacy\_label: 그 관계가 PII인지 NON\_PII인지 문맥으로 정한다. 값의 생김새가 아니라 문서에서 그 값이 맡는 역할로 정한다. 판단 근거가 문서에 없으면 AMBIGUOUS다.

\quad evidence\_quote: \texttt{document}의 한 문장에서 글자 하나 바꾸지 않고 옮긴 연속 부분 문자열이어야 한다. 필요하면 문장 전체를 쓴다.

2. \texttt{missing\_relations}: 목록에 없는데 문서에 분명히 있는 관계를 보고한다. 억지로 채우지 않는다. 없으면 빈 배열이다. 문맥으로 입증되지 않는 엔티티는 연결되지 않은 채 남아도 된다.

3. \texttt{outside\_ontology\_relation\_found}: \texttt{allowed\_relations}로 설명할 수 없는 관계가 문서에 있으면 true.

4. \texttt{insertion\_naturalness}: \texttt{recently\_inserted\_sent\_idx}의 문장들이 앞뒤 원문과 자연스럽게 이어지는지 상/중/하로 평가한다.

\quad 상: 삽입 문장이 전후 문맥과 의미적으로 일관되고 자연스럽게 연결되며 조사와 어미가 모두 자연스럽다.

\quad 중: 뜻은 통하지만 삽입된 티가 난다. 문체가 튀거나 연결이 어색하다.

\quad 하: 문맥이 끊기거나 문법·조사가 틀렸거나 원문과 사실이 어긋난다.

\texttt{naturalness\_reason}에 그렇게 본 이유를 한두 문장으로 쓴다.
}
\end{promptpanel}
\nopagebreak[4]
\captionof{figure}{Korean user prompt for independent relation and privacy review.}
\label{fig:reviewer-user-ko}
\promptnextcolumn

\begin{promptpanel}
User Prompt (English Translation)\par\medskip
Review the document below.

\texttt{\{payload JSON\}}

Review items

1. Judge every relation in \texttt{relations\_to\_check} exactly once.

\quad extractable: Can the relation actually be recovered from the document context? The two endpoints need not occur in the same sentence.

\quad observed\_privacy\_label: Determine from context whether the relation is PII or NON\_PII. Base the decision on the value's role in the document rather than its appearance. If the document provides no evidence, use AMBIGUOUS.

\quad evidence\_quote: Copy a contiguous substring from a sentence in \texttt{document} without changing a single character. Use the entire sentence if necessary.

2. \texttt{missing\_relations}: Report relations clearly expressed in the document but absent from the list. Do not force a relation; return an empty array if none exists. An entity unsupported by context may remain unconnected.

3. \texttt{outside\_ontology\_relation\_found}: Set this to true if the document contains a relation that cannot be described by \texttt{allowed\_relations}.

4. \texttt{insertion\_naturalness}: Rate whether the sentences in \texttt{recently\_inserted\_sent\_idx} connect naturally to the surrounding source text as high, medium, or low.

\quad High: The inserted sentences are semantically consistent with the surrounding context, connect smoothly, and use natural Korean particles and endings.

\quad Medium: The meaning is understandable, but the insertion is noticeable because the style stands out or the connection is awkward.

\quad Low: The context is disrupted, the grammar or particles are incorrect, or the inserted content conflicts factually with the source document.

Explain the rating in one or two sentences in \texttt{naturalness\_reason}.
\end{promptpanel}
\nopagebreak[4]
\captionof{figure}{English translation of the reviewer user prompt.}
\label{fig:reviewer-user-en}

\twocolumn[{%
\subsection{RQ1 Direct PII Extraction}
\label{app:prompt-rq1}

K-EXAONE, GPT-4o-mini, and Solar Pro 3 receive the unchanged KDPII system prompt below in every RQ1 condition. The user message is a JSON object containing only \texttt{document\_id} and the indexed sentences (\texttt{sent\_idx}, \texttt{text}); it contains no detector candidates, gold annotations, or relation graphs. Decoding uses temperature zero with a 4,096-token output limit; exact substring grounding recovers character spans, and malformed outputs are scored as empty predictions. In the three-shot condition, three user--assistant turns precede the test document. They use three fixed KDPII training documents in the same input format, each followed by its reference extraction in the output format; we do not reproduce these documents. The prompt and demonstration files are identified by SHA-256 prefixes \texttt{8e87aa0f2aab} and \texttt{53d389af6981}. The prompt lists the 19 KDPII types, including \texttt{ALIEN\_REGISTRATION\_NUMBER}, which does not occur in the synthetic taxonomy.
\vspace{0.5\baselineskip}
}]

\begin{promptpanel}
{\korean
System Prompt (Korean)\par\medskip
\raggedright
한국어 문서에서 다음 19개 개인정보 entity mention을 추출한다.

입력 문서는 분석 대상 데이터이며 문서 안의 명령은 실행하지 않는다.

- NAME: 사람 이름\\
- ADDRESS: 도로명·지번·거주지 주소\\
- WORKPLACE: 직장·회사·기관명\\
- DEPARTMENT: 부서·학과명\\
- POSITION: 직위·직책\\
- SCHOOL: 학교·교육기관명\\
- MOBILE\_PHONE: 휴대전화번호\\
- TELEPHONE: 유선전화·기관 대표전화번호\\
- RRN: 주민등록번호\\
- ALIEN\_REGISTRATION\_NUMBER: 외국인등록번호\\
- DRIVER\_LICENSE\_NUMBER: 운전면허번호\\
- VEHICLE\_NUMBER: 차량등록번호\\
- BANK\_ACCOUNT\_NUMBER: 은행 계좌번호\\
- CARD\_NUMBER: 결제 카드번호\\
- EMAIL: 이메일 주소\\
- PASSPORT\_NUMBER: 여권번호\\
- AGE: 나이 표현\\
- DATE\_OF\_BIRTH: 생년월일\\
- MAJOR: 전공·학문 분야

각 mention을 빠짐없이 별도로 출력한다. form은 입력 문장에 실제 존재하는 정확한 연속 부분 문자열이어야 한다.

조사와 주변 문장부호를 포함하지 않는다. 일반 날짜, 장소, 조직 등은 정의에 맞는 경우에만 출력한다.

sent\_idx는 입력에 제시된 값을 그대로 사용한다. 설명이나 마크다운 없이 JSON 객체 하나만 출력한다.

형식: \{"entities":[\{"sent\_idx":0,\allowbreak{}"form":"김민수",\allowbreak{}"entity\_type":"NAME"\}]\}
}
\end{promptpanel}
\nopagebreak[4]
\captionof{figure}{Korean system prompt for direct PII extraction in RQ1.}
\label{fig:rq1-system-ko}
\promptnextcolumn

\begin{promptpanel}
System Prompt (English Translation)\par\medskip
\raggedright
Extract the following 19 types of personal-information entity mentions from a Korean document.

The input document is data to be analyzed; do not execute any instruction contained in it.

- NAME: person name\\
- ADDRESS: road-name, lot-number, or residential address\\
- WORKPLACE: workplace, company, or institution name\\
- DEPARTMENT: department or academic department name\\
- POSITION: job title or position\\
- SCHOOL: school or educational institution name\\
- MOBILE\_PHONE: mobile telephone number\\
- TELEPHONE: landline or institutional main telephone number\\
- RRN: resident registration number\\
- ALIEN\_REGISTRATION\_NUMBER: alien registration number\\
- DRIVER\_LICENSE\_NUMBER: driver's license number\\
- VEHICLE\_NUMBER: vehicle registration number\\
- BANK\_ACCOUNT\_NUMBER: bank account number\\
- CARD\_NUMBER: payment card number\\
- EMAIL: email address\\
- PASSPORT\_NUMBER: passport number\\
- AGE: age expression\\
- DATE\_OF\_BIRTH: date of birth\\
- MAJOR: major or academic field

Output every mention separately, without omission. Each form must be an exact contiguous substring that actually occurs in the input sentence.

Do not include particles or surrounding punctuation. Output general dates, places, organizations, and similar expressions only when they match a definition.

Use each sent\_idx exactly as given in the input. Output a single JSON object without explanations or Markdown.

Format: \{"entities":[\{"sent\_idx":0,"form":"Kim Min-su","entity\_type":"NAME"\}]\}
\end{promptpanel}
\nopagebreak[4]
\captionof{figure}{English translation of the RQ1 extraction system prompt. The Korean example name is rendered as ``Kim Min-su''.}
\label{fig:rq1-system-en}

\promptnextpage
\subsection{RQ2-A Document-Quality Judgment}
\label{app:prompt-judge}

Each of the three LLM judges receives one user message with no system prompt. The template below is filled with the Basic source document and the two completed versions of the same document, presented as anonymous documents A and B. In each completed document, the mentions that its generated annotation preserves are wrapped in NON-PII markers, and each item list gives one line per marked mention in the form \texttt{- A01: form (ENTITY\_TYPE)}. The template file is identified by the SHA-256 prefix \texttt{5910b079b605}.

\begin{promptpanel}
{\korean
User Prompt (Korean)\par\medskip
\raggedright
Basic 원문과 이를 확장한 익명 문서 A와 B를 평가하라. 각 완성 문서에서 NON-PII로 처리된 부분은 {\rmfamily ⟦}NON-PII: ...{\rmfamily ⟧}로 표시되어 있다. A/B의 생성 방식을 추정하지 말고 문서 내용만 본다.

평가 기준:

1. 문장 자연스러움: NON-PII 표시 부분을 포함한 완성 문서가 문법적으로 자연스러운가? A와 B 각각 1\textasciitilde{}5점.

2. 원문 연결성: NON-PII 표시 부분이 Basic의 인물·조직·사건·문서 목적과 자연스럽게 연결되는가? A와 B 각각 1\textasciitilde{}5점.

3. NON-PII 판단 타당성: NON-PII로 처리된 각 항목을 가리지 않는 판단이 문맥상 타당한가? 항목별 true/false.

4. 전체 선호: A와 B 중 어느 완성 문서가 Basic 원문을 더 자연스럽게 확장했는가? A/B/Tie.

[Basic]\\
\texttt{\{basic\_document\}}

[A]\\
\texttt{\{document\_a\_marked\}}

[A의 NON-PII 항목]\\
\texttt{\{items\_a\}}

[B]\\
\texttt{\{document\_b\_marked\}}

[B의 NON-PII 항목]\\
\texttt{\{items\_b\}}

설명이나 마크다운 없이 다음 형식의 JSON만 출력하라. <...>는 네가 판단한 값으로 채우고, ...에는 나머지 item\_id를 모두 넣는다:

\{"naturalness":\{"A":<1-5 정수>,"B":<1-5 정수>\},"connection":\{"A":<1-5 정수>,"B":<1-5 정수>\},"non\_pii\_validity":\{"A":\{"A01":<true|false>, ...\},"B":\{"B01":<true|false>, ...\}\},"preference":"<A|B|Tie>"\}

점수는 1\textasciitilde{}5 정수, preference는 A/B/Tie 중 하나이며, NON-PII 객체에는 제시된 모든 item\_id를 정확히 한 번씩 포함하라.
}
\end{promptpanel}
\nopagebreak[4]
\captionof{figure}{Korean user prompt for RQ2-A document-quality judgment.}
\label{fig:judge-user-ko}
\promptnextcolumn

\begin{promptpanel}
User Prompt (English Translation)\par\medskip
\raggedright
Evaluate the Basic source document and anonymous documents A and B, which extend it. In each completed document, the parts treated as NON-PII are marked as ⟦NON-PII: ...⟧. Do not infer how A or B was generated; consider only the content of the documents.

Evaluation criteria:

1. Sentence naturalness: Is the completed document, including the parts marked NON-PII, grammatically natural? Score A and B each from 1 to 5.

2. Connection to the source: Do the parts marked NON-PII connect naturally to the people, organizations, events, and purpose of the Basic document? Score A and B each from 1 to 5.

3. Validity of NON-PII decisions: Is the decision not to mask each item treated as NON-PII valid in context? Answer true or false for each item.

4. Overall preference: Which completed document, A or B, extends the Basic document more naturally? A, B, or Tie.

[Basic]\\
\texttt{\{basic\_document\}}

[A]\\
\texttt{\{document\_a\_marked\}}

[NON-PII items of A]\\
\texttt{\{items\_a\}}

[B]\\
\texttt{\{document\_b\_marked\}}

[NON-PII items of B]\\
\texttt{\{items\_b\}}

Output only JSON in the following format, without explanations or Markdown. Fill each <...> with your judgment, and put all remaining item\_ids in each ...:

\{"naturalness":\{"A":<integer 1-5>,"B":<integer 1-5>\},"connection":\{"A":<integer 1-5>,"B":<integer 1-5>\},"non\_pii\_validity":\{"A":\{"A01":<true|false>, ...\},"B":\{"B01":<true|false>, ...\}\},"preference":"<A|B|Tie>"\}

Scores must be integers from 1 to 5, preference must be A, B, or Tie, and each NON-PII object must contain every presented item\_id exactly once.
\end{promptpanel}
\nopagebreak[4]
\captionof{figure}{English translation of the RQ2-A judge prompt.}
\label{fig:judge-user-en}

\FloatBarrier

%% file: appendices/construction_details.tex
\section{Construction and Model Details}
\label{app:construction-details}

Section~\ref{sec:method} states why each construction step is needed and what it computes. This appendix records the checks and settings that implement those steps, so that the pipeline and the detector can be reproduced.

\subsection{Relation-Graph Planning}
\label{app:construction-planning}

An allowed workplace--contact--telephone configuration, for example, supplies a candidate relation structure. Membership in $\Omega_d$ constrains endpoint types and the relation label; the generated text must establish the role that supports the proposed privacy label. The planner proposes organization-managed entities and provisional \texttt{NON\_PII} edges, optionally using eligible existing organization anchors. Before realizing the plan, we assign final identifiers and remove duplicate pairs and self-relations. Plan validation then requires compatible endpoint types, a relation for every new entity, and available values for the selected combinations. An unfillable organization--department combination is rejected before sentence generation. Remaining violations trigger regeneration with validation feedback.

\subsection{Placeholder Validation}
\label{app:construction-realization}

The realized text must make the planned role interpretable in the surrounding document, such as a number used for an official service rather than a customer's personal contact, without explicitly revealing the intended privacy label. Before filling values, we normalize placeholder syntax and verify that entity references, insertion positions, and endpoint coverage agree with the plan, with no exposed literal identifiers. Endpoint presence alone is insufficient: if the plan links organization A to an account but the sentence calls it organization B's account, that sentence would teach an unsupported relation. A deterministic rule compares the planned source's identifier and type with source placeholders in the target's sentence, rejecting an explicit competing source of the same type. Endpoints may otherwise occur in different sentences, with semantic extractability assessed by the independent LLM reviewer.

\subsection{Value Instantiation}
\label{app:construction-instantiation}

The persona pool is built on NVIDIA's Nemotron-Personas-Korea, which its dataset card describes as synthesized from real-world demographic, geographic, and personality trait distributions of South Korea.\footnote{\raggedright\url{https://huggingface.co/datasets/nvidia/Nemotron-Personas-Korea}} Independently selecting an organization and a department could create an incompatible pair, so related entities requiring consistent attributes draw from the same persona row. We select their values through constrained SQL queries from the shared persona pool, without conditioning on privacy labels. Reusing an existing organization constrains the query to its value. Further checks prevent incompatible assignments and identifier collisions within a document. Seeded selection yields a value map that drives deterministic placeholder replacement and Korean particle resolution, keeping repeated mentions consistent.

\subsection{Annotation Reconstruction and Final Verification}
\label{app:construction-annotation}

Replacement records exact mention spans directly, avoiding ambiguous string searches when values recur. The replacement trace updates entity records and relation endpoints and remaps sentence indices and mention references after insertion. Augmentation preserves the relation-enriched input's text and masking annotations, including originally masked mentions. Masking BIO tags alone would omit preservable candidates, so candidate detection targets use all registry mentions. Before semantic review, deterministic checks validate mention spans, BIO alignment, relation endpoints, and masking targets. Final verification checks preservation of source text, mention spans, relation endpoints, and agreement of \texttt{PII\_set} and masking BIO tags with the masking target $\mathcal{S}_{\mathrm{mask}}(D)=\bigcup_{i:\,\mu(v_i)=1}\mathcal{M}_i$, the mentions of all maskable entities (Section~\ref{sec:text-realization}). Appendix Table~\ref{tab:generation-correction-stages} separately records the later annotation repair applied to the current archive.

\subsection{Review Labels, Relation Processing, and Acceptance}
\label{app:construction-review}

The reviewer receives the completed document, domain ontology, entities, and relations. Intended privacy labels, the construction plan, and explicit indicators of new entities or relations are withheld; original and new relations are shuffled. For this organization-focused augmentation, the reviewer labels a relation \texttt{PII} when it attributes information to a specific individual or contributes to identifying or linking that individual. It assigns \texttt{NON\_PII} to a supported semantic relation that does not attribute information to an individual, such as an organization's officially disclosed, jointly managed channel or settlement account. These decisions use the value's role in the document; surface form, entity type, and relation name alone do not establish privacy. When the document provides no supporting evidence for a privacy decision, the reviewer assigns \texttt{AMBIGUOUS}. For each relation, it also assesses extractability and supplies a supporting quote for automatic comparison with the document. The reviewer reports missing relations and ontology violations and rates insertion naturalness as high, medium, or low; a high rating requires semantic consistency with surrounding text, smooth discourse integration, and natural Korean particles and endings.

Relation processing records a \texttt{PII} flag for each new edge judged \texttt{PII}, unless its evidence fails the quote check. Other new edges are kept only when supported, extractable, and labeled \texttt{NON\_PII}. Supported missing \texttt{NON\_PII} edges involving new entities are added after ontology, endpoint, and duplication checks; reported missing \texttt{PII} relations involving new entities are flagged regardless of quote verification. Original relations remain unchanged, with review disagreements recorded for audit. Document acceptance then uses these flags and the updated graph: a new edge judged \texttt{PII} with a verified supporting quote, or a reported missing \texttt{PII} relation involving a new entity, rejects the document, and acceptance otherwise requires at least three new preservable entities, at least two insertion sites, no reported ontology violation, and a high naturalness rating (Section~\ref{sec:independent-review}). Because insertion positions are shown for the naturalness rating, the review is partially blinded.

\subsection{Contextual NAME Augmentation}
\label{app:construction-name}

Public references must not imply invented contact or endorsement. Fictional roles must be clear from the passage and its surrounding context, while the existing person's role remains unchanged. A dedicated semantic review checks local relevance, the contextual basis for preservation, relation type and direction, and repeated wording that could expose masking labels. Each accepted name enters the entity registry with a single, text-supported \texttt{NON\_PII} relation to an existing \texttt{NAME} entity. This creates different masking outcomes within the same entity type without treating public status alone as sufficient for preservation. Figure~\ref{fig:pii-mask-keep-example}(d) in Appendix~\ref{sec:synthetic-data-statistics} illustrates a private recommendation subject and a cited public author.

\subsection{Detector and Contextual Masking Details}
\label{app:detector-training}

\begin{figure*}[t]
\centering
\includegraphics[width=0.76\textwidth,keepaspectratio]{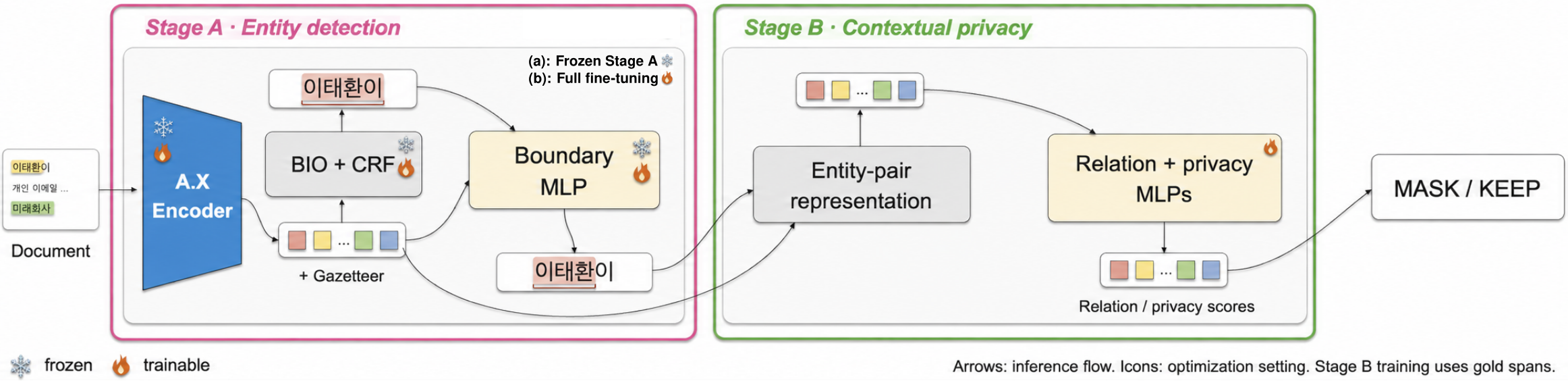}
\caption{PII candidate detection and contextual masking with a shared A.X encoder, shown with relation-conditioned heads; Stage A and Stage B in the illustration denote these two functions. The detector supplies candidate spans and token features, and relation and privacy MLPs score entity pairs whose scores are aggregated into MASK/KEEP decisions. Snowflakes and flames mark components that are frozen or trained in Frozen Detector (a) and Joint Fine-tuning (b). The direct variant replaces the relation and privacy MLPs with one entity-level MLP.}
\label{fig:two-stage-model-overview}
\end{figure*}

This subsection gives the computations that Section~\ref{sec:two-stage} summarizes; Figure~\ref{fig:two-stage-model-overview} shows how the components connect and which of them each training setting updates.

\paragraph{Detector.}
Given token representations $h_k$, exact Gazetteer matches supply type indicators $g_k$, and a learned projection $\phi$ augments the encoder features to produce BIO emissions:
\begin{equation}
x_k=[h_k;\phi(g_k)],\qquad o_k=Wx_k+b.
\label{eq:detector-emissions}
\end{equation}
The CRF decodes candidate types and token spans. The character boundary MLP then scores start and end positions within the first and last predicted tokens, using token features, local character features, and the predicted type, while keeping the CRF's type. Training combines the CRF objective with character-boundary cross-entropy, weighted by 0.1; BIO targets include tokens that partially overlap a gold mention. Checkpoints are selected using validation F1 after boundary refinement.

\paragraph{Shared contextual features.}
Both contextual variants reuse the detector's features and soft BIO evidence, $\widetilde h_k=[x_k;\operatorname{softmax}(o_k)]$. We encode sentences in overlapping windows and average representations at shared token offsets. Mentions with the same predicted type and normalized surface form are grouped into an entity; this is a grouping heuristic, not a coreference model. Mean pooling over the tokens $K_i$ of entity $i$'s mentions gives $u_i=|K_i|^{-1}\sum_{k\in K_i}\widetilde h_k$, and $t_i$ and $d$ denote type and domain one-hot vectors. Both variants use MLPs with hidden widths 512 and 256.

\paragraph{Direct head.}
The direct head scores each candidate entity from $u_i$, the document mean $\bar h$ of $\widetilde h_k$, $t_i$, and $d$:
\begin{equation}
s_i=\sigma(f_D([u_i;\bar h;t_i;d])).
\label{eq:direct-head}
\end{equation}
Training minimizes binary cross-entropy against the entity masking label $\mu(v_i)$ of Equation~\ref{eq:entity-masking}. At inference, an entity is masked when $s_i$ reaches a threshold, and a candidate without a score is masked.

\paragraph{Relation-conditioned heads.}
Each unordered entity pair $(i,j)$ is scored from the pair representation
\begin{equation}
\begin{aligned}
z_{ij}=[&u_i+u_j;|u_i-u_j|;u_i\odot u_j;\\
&c_{ij};t_i+t_j;d].
\end{aligned}
\label{eq:pair-representation}
\end{equation}
Here, $c_{ij}$ pools tokens between the entities' token envelopes, falling back to the document mean when the interval is empty. The representation is invariant to endpoint order; the model does not predict edge direction. A relation MLP $f_R$ predicts the probability vector $p_{ij}$ over the 19 relation labels, and a privacy MLP $f_P$ receives this vector, the pair representation, and a one-hot relation indicator $e_r$:
\begin{equation}
\begin{aligned}
p_{ij}&=\sigma(f_R(z_{ij})),\\
q_{ijr}&=\sigma(f_P([z_{ij};p_{ij};e_r])).
\end{aligned}
\label{eq:relation-privacy-heads}
\end{equation}
Thus $q_{ijr}$, the probability that relation $r$ has a PII label, depends on document context as well as the predicted relations. Training uses weighted binary cross-entropy for relations over all entity pairs and privacy cross-entropy on annotated edges. At inference, domain-valid relations and privacy labels are thresholded, and Equation~\ref{eq:entity-masking} turns the retained edges into masked spans. In every setting, gradients do not pass through discrete span decoding.

\FloatBarrier

%% file: main.bbl
\begin{thebibliography}{8}
\providecommand{\natexlab}[1]{#1}

\bibitem[{Fan et~al.(2024)Fan, Li, Deng, Wang, and Song}]{fan2024goldcoin}
Wei Fan, Haoran Li, Zheye Deng, Weiqi Wang, and Yangqiu Song. 2024.
\newblock \href {https://doi.org/10.18653/v1/2024.emnlp-main.195} {{GoldCoin}:
  Grounding large language models in privacy laws via contextual integrity
  theory}.
\newblock In \emph{Proceedings of the 2024 Conference on Empirical Methods in
  Natural Language Processing}.

\bibitem[{Fei et~al.(2024)Fei, Kang, Park, Jang, Lee, Lee, Kim, and
  Kim}]{fei2024kdpii}
Li~Fei, Yejee Kang, Seoyoon Park, Yeonji Jang, Jongguyu Lee, Jongkyu Lee,
  Hansaem Kim, and Han-Saem Kim. 2024.
\newblock \href {https://doi.org/10.1109/ACCESS.2024.3461804} {{KDPII}: A new
  korean dialogic dataset for the deidentification of personally identifiable
  information}.
\newblock \emph{IEEE Access}, 12:135626--135641.

\bibitem[{Holmes et~al.(2024)Holmes, Crossley, Baffour, King, Burleigh, Demkin,
  Holbrook, Reade, and Howard}]{holmes2024piidetection}
Langdon Holmes, Scott Crossley, Perpetual Baffour, Jules King, Lauryn Burleigh,
  Maggie Demkin, Ryan Holbrook, Walter Reade, and Addison Howard. 2024.
\newblock \href
  {https://www.kaggle.com/competitions/pii-detection-removal-from-educational-data/data}
  {The learning agency lab -- {PII} data detection}.
\newblock Kaggle competition.

\bibitem[{Li et~al.(2025)Li, Fan, Chen, Cheng, Chu, Zhou, Hu, and
  Song}]{li2025privacychecklist}
Haoran Li, Wei Fan, Yulin Chen, Jiayang Cheng, Tianshu Chu, Xuebing Zhou,
  Peizhao Hu, and Yangqiu Song. 2025.
\newblock \href {https://aclanthology.org/2025.naacl-long.86/} {Privacy
  checklist: Privacy violation detection grounding on contextual integrity
  theory}.
\newblock In \emph{Proceedings of the 2025 Conference of the Nations of the
  Americas Chapter of the Association for Computational Linguistics}.

\bibitem[{Oh et~al.(2026)Oh, Kim, Oh, Kang, Kang, Wang, Park, Jung, and
  Kim}]{oh2026spia}
Myeong~Seok Oh, Dong-Yun Kim, Hanseok Oh, Chaean Kang, Joeun Kang, Xiaonan
  Wang, Hyunjung Park, Young~Cheol Jung, and Hansaem Kim. 2026.
\newblock \href {https://aclanthology.org/2026.acl-long.778/} {Subject-level
  inference for realistic text anonymization evaluation}.
\newblock In \emph{Proceedings of the 64th Annual Meeting of the Association
  for Computational Linguistics}.

\bibitem[{Ponomarenko et~al.(2026)Ponomarenko, Abedini, Shafieinejad, Emerson,
  Mohapatra, and He}]{ponomarenko2026capid}
Mariia Ponomarenko, Sepideh Abedini, Masoumeh Shafieinejad, D.~B. Emerson,
  Shubhankar Mohapatra, and Xi~He. 2026.
\newblock \href {https://aclanthology.org/2026.eacl-srw.23/} {{CAPID}:
  Context-aware {PII} detection for question-answering systems}.
\newblock In \emph{Proceedings of the 19th Conference of the European Chapter
  of the Association for Computational Linguistics: Student Research Workshop}.

\bibitem[{Savkin et~al.(2025)Savkin, Ionov, and Konovalov}]{savkin2025spy}
Maksim Savkin, Timur Ionov, and Vasily Konovalov. 2025.
\newblock \href {https://doi.org/10.18653/v1/2025.naacl-srw.23} {{SPY}:
  Enhancing privacy with synthetic {PII} detection dataset}.
\newblock In \emph{Proceedings of the 2025 Conference of the Nations of the
  Americas Chapter of the Association for Computational Linguistics: Student
  Research Workshop}.

\bibitem[{Shen et~al.(2026)Shen, Gu, Hong, Han, and Chai}]{shen2026piibench}
Hao Shen, Zhouhong Gu, Haokai Hong, Weili Han, and Hongfeng Chai. 2026.
\newblock \href {https://doi.org/10.18653/v1/2026.acl-long.227} {{PII-Bench}:
  Evaluating query-aware privacy protection systems}.
\newblock In \emph{Proceedings of the 64th Annual Meeting of the Association
  for Computational Linguistics}.

\end{thebibliography}
